\documentclass[dvipsnames,format=sigconf]{acmart}

\usepackage[export]{adjustbox}
\usepackage{subcaption}
\usepackage{hyperref}
\usepackage{longtable}
\usepackage{xcolor}

\usepackage{tcolorbox}
\tcbuselibrary{listings,skins}

\tcbset{
  promptlisting/.style={
    listing only,
    enhanced,
    width=\linewidth, 
    colback=black!1,
    colframe=black!25,
    boxrule=0.4pt,
    arc=2pt,
    left=1mm,right=1mm,top=1mm,bottom=1mm,
    colbacktitle=black!10,
    coltitle=black,
    fonttitle=\bfseries\small,
    listing options={
      basicstyle=\ttfamily\small,
      columns=fullflexible,
      keepspaces=true,
      upquote=true,
      breaklines=true,
      breakatwhitespace=true,
      prebreak={},
      postbreak={},
      breakindent=0pt,
      breakautoindent=false,
      showspaces=false,
      showtabs=false,
      showstringspaces=false,
    },
  },
}

\usepackage{multirow}
\usepackage[table]{xcolor}
\usepackage[normalem]{ulem} 

\AtBeginDocument{%
  }

\copyrightyear{2026}
\acmYear{2026}
\setcopyright{cc}
\setcctype{by}
\acmConference[GECCO '26]{Genetic and Evolutionary Computation Conference}{July 13--17, 2026}{San Jose, Costa Rica}
\acmBooktitle{Genetic and Evolutionary Computation Conference (GECCO '26), July 13--17, 2026, San Jose, Costa Rica}
\acmDOI{10.1145/3795095.3805186}
\acmISBN{979-8-4007-2487-9/2026/07}

\begin{document}

\title[Replicating Picbreeder with Large Vision-Language Models]{In Search of the Ingredients of Open-Endedness:\\Replicating Picbreeder with Large Vision-Language Models}

\author{Sam Earle}
\authornote{Work partially completed during an internship at Sakana AI.}
\orcid{0000-0003-3783-9486}
\affiliation{%
  \institution{New York University}
  \city{Brooklyn}
  \state{New York}
  \country{USA}
}
\email{sam.earle@nyu.edu}

\author{Kai Arulkumaran}
\affiliation{%
  \institution{Sakana AI}
  \city{Tokyo}
  \country{Japan}}
\email{kailash@sakana.ai}

\author{Andrew Dai}
\affiliation{%
  \institution{Independent}
  \city{Dublin}
  \country{Ireland}
}
\email{adai@tcd.ie}

\author{Akarsh Kumar}
\affiliation{%
 \institution{Massachusetts Institute of Technology}
 \city{Cambridge}
 \state{Massachussettes}
 \country{USA}}
\email{akarsh@sakana.ai}

\author{Julian Togelius}
\affiliation{%
  \institution{New York University}
  \city{New York}
  \state{New York}
  \country{USA}}
\email{julian@togelius.com}

\author{Sebastian Risi}
\affiliation{%
  \institution{Sakana AI}
  \city{Tokyo}
  \country{Japan}}
\email{sebastianrisi@sakana.ai}

\renewcommand{\shortauthors}{Earle et al.}

\begin{abstract}
We are in the midst of large-scale industrial and academic efforts to automate the processes of scientific, technological and creative production through AI-driven assistants. Historically, a fundamental property of these processes in their human form has been their open-endedness: their capacity for generating a seemingly endless supply of novel and meaningful new forms. Do artificial agents have any capacity for such fruitful unguided discovery? To answer this question, we turn to Picbreeder, the canonical exemplar of human-driven open-ended search, in which users collaboratively generated a diverse library of images through interactive evolution of small neural networks. We replicate Picbreeder, replacing human users with frontier Vision Language Models (VLMs). We observe clear qualitative differences between the output of our system and the historical human baseline, and attempt to characterize them using metrics of phylogenetic complexity and visual and semantic salience and novelty. In an effort to identify some of the causal factors contributing to these differences, we study the addition of exploratory noise to the agents' selection process, of behavioral diversity between agents, and of narrative momentum in the form of memory of past actions. We make our code available at \href{https://github.com/smearle/picbreeder-vlm}{https://github.com/smearle/picbreeder-vlm}.
\end{abstract}

%

\ccsdesc[300]{Computing methodologies~Multi-agent systems}
\ccsdesc[300]{Computing methodologies~Cognitive science}

\keywords{Open-Endedness, Vision-Language Models, Picbreeder}
\newlength{\thumbsize}
\setlength{\thumbsize}{34px}
\begin{teaserfigure}
\begin{center}
\includegraphics[width=\thumbsize]{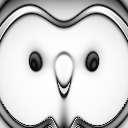}
\includegraphics[width=\thumbsize]{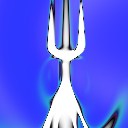}
\includegraphics[width=\thumbsize]{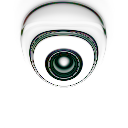}
\includegraphics[width=\thumbsize]{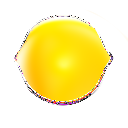}
\includegraphics[width=\thumbsize]{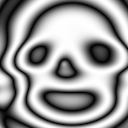}
\includegraphics[width=\thumbsize]{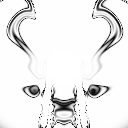}
\includegraphics[width=\thumbsize]{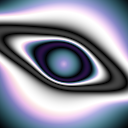}
\includegraphics[width=\thumbsize]{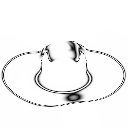}
\includegraphics[width=\thumbsize]{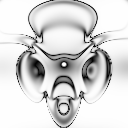}
\includegraphics[width=\thumbsize]{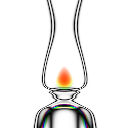}
\includegraphics[width=\thumbsize]{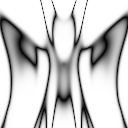}
\includegraphics[width=\thumbsize]{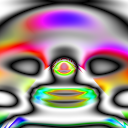}
\includegraphics[width=\thumbsize]{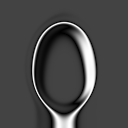}
\includegraphics[width=\thumbsize]{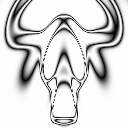}
\includegraphics[width=\thumbsize]{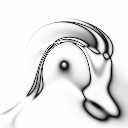}
\includegraphics[width=\thumbsize]{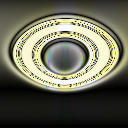}
\includegraphics[width=\thumbsize]{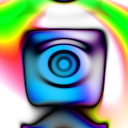}
\includegraphics[width=\thumbsize]{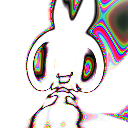}
\includegraphics[width=\thumbsize]{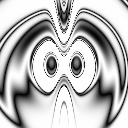}
\includegraphics[width=\thumbsize]{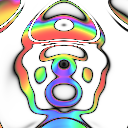}
\includegraphics[width=\thumbsize]{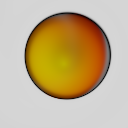}
\includegraphics[width=\thumbsize]{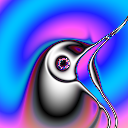}
\includegraphics[width=\thumbsize]{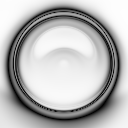}
\includegraphics[width=\thumbsize]{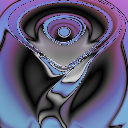}
\includegraphics[width=\thumbsize]{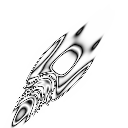}
\includegraphics[width=\thumbsize]{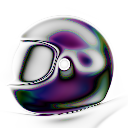}
\includegraphics[width=\thumbsize]{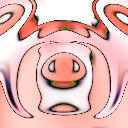}
\includegraphics[width=\thumbsize]{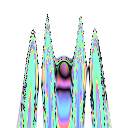}
\includegraphics[width=\thumbsize]{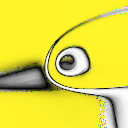}
\includegraphics[width=\thumbsize]{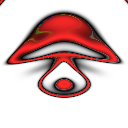}
\includegraphics[width=\thumbsize]{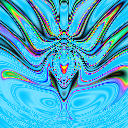}
\includegraphics[width=\thumbsize]{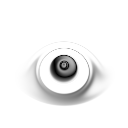}
\includegraphics[width=\thumbsize]{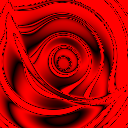}
\includegraphics[width=\thumbsize]{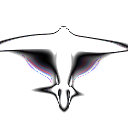}
\includegraphics[width=\thumbsize]{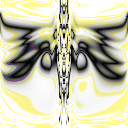}
\includegraphics[width=\thumbsize]{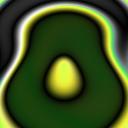}
\includegraphics[width=\thumbsize]{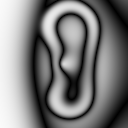}
\includegraphics[width=\thumbsize]{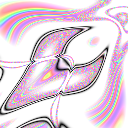}
\includegraphics[width=\thumbsize]{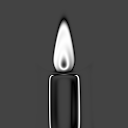}
\includegraphics[width=\thumbsize]{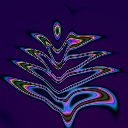}
\includegraphics[width=\thumbsize]{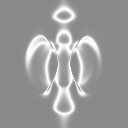}
\includegraphics[width=\thumbsize]{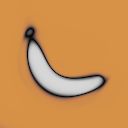}
\end{center}
\caption{Large Vision-Language Models play Picbreeder and discover novel images. Cherry-picked examples.}
\end{teaserfigure}

\received{20 February 2007}
\received[revised]{12 March 2009}
\received[accepted]{5 June 2009}

\maketitle

\section{Introduction}


Open-ended processes of learning and discovery are crucial for civilization. 
In science, mathematics, art, and technology, the most significant breakthroughs are often as much the result of serendipity and curiosity as of optimizing explicit objectives.
In contrast to the prevailing paradigm of machine learning, which seeks a single optimal solution, open-ended processes are divergent and produce an ever-growing phylogeny of novel artifacts.
Natural evolution is an example---its computational counterparts largely are not.
Indeed, creating credibly open-ended artificial systems remains an open problem~\citep{stanley2017open,stepney2024open}. 


Beyond the monolithic examples above, we can identify some real-world microcosms of open-endedness.
Of particular interest are human-in-the-loop computational systems.
Picbreeder \citep{secretan2008picbreeder,secretan2011picbreeder} is the canonical example, in which users collaboratively create interesting images via interactive evolution.
Branching from each other's creations, they continually grow an archive of artifacts according to their own whims and preferences.
If we were able to recreate such a system 
in a purely computational substrate,
it could serve as a kind of model organism, 
allowing us to experiment with its components and parameters so as to better understand the building blocks of open-endedness.

In this paper we describe a fully artificial recreation of Picbreeder, in which we use Vision-Language Models (VLMs) \cite{bordes2024introduction} in place of humans.
We analyze this system both quantitatively and qualitatively, and vary key components to understand how its output is affected.
Concretely, we answer the simple question: ``what happens when a VLM plays Picbreeder?''
More broadly, we propose a strategy for understanding open-ended processes by replicating them artificially, replacing humans with AI agents without explicit objectives.
Our research questions ask which design choices allow the system to create a meaningful diversity of artifacts. Namely:

\begin{enumerate}
    \item Do VLM agents need access to \textbf{history}? Does access to a context/memory of past actions encourage productive divergence by allowing them to recognize and steer away from existing patterns in the system?
    Or does this increased exposure reinforce existing biases, leading to mode collapses?
    \item Do VLM agents need explicit \textbf{exploration} strategies to help them explore the space of artifacts more effectively, by forcing the agent to parts of the search space they otherwise would not have visited?
    Or are they inherently capable of balancing discovery and optimization?
    \item Do we need a \textbf{multi-agent} system? Does the simulation of multiple personalities produce open-ended creative/competitive dynamics, or does it merely define a set of fixed attractors in the search space?
\end{enumerate}

In sum, we find that small amounts of exploratory noise can increase diversity of generated archives, but at the cost of the quality of images therein; surprisingly little history is necessary for optimal performance, with greater context lengths leading to pathological behavior; and increasing the number of unique agents contributes to exploration without sacrificing quality according to our quantitative metrics, but results in the propagation of nondescript, noisy, and potentially adversarial images among the archive (\autoref{fig:traits_1k_mush}).

\begin{figure*}
\begin{subfigure}{\linewidth}
\centering
\begin{subfigure}{.31\linewidth}
\includegraphics[width=\linewidth]{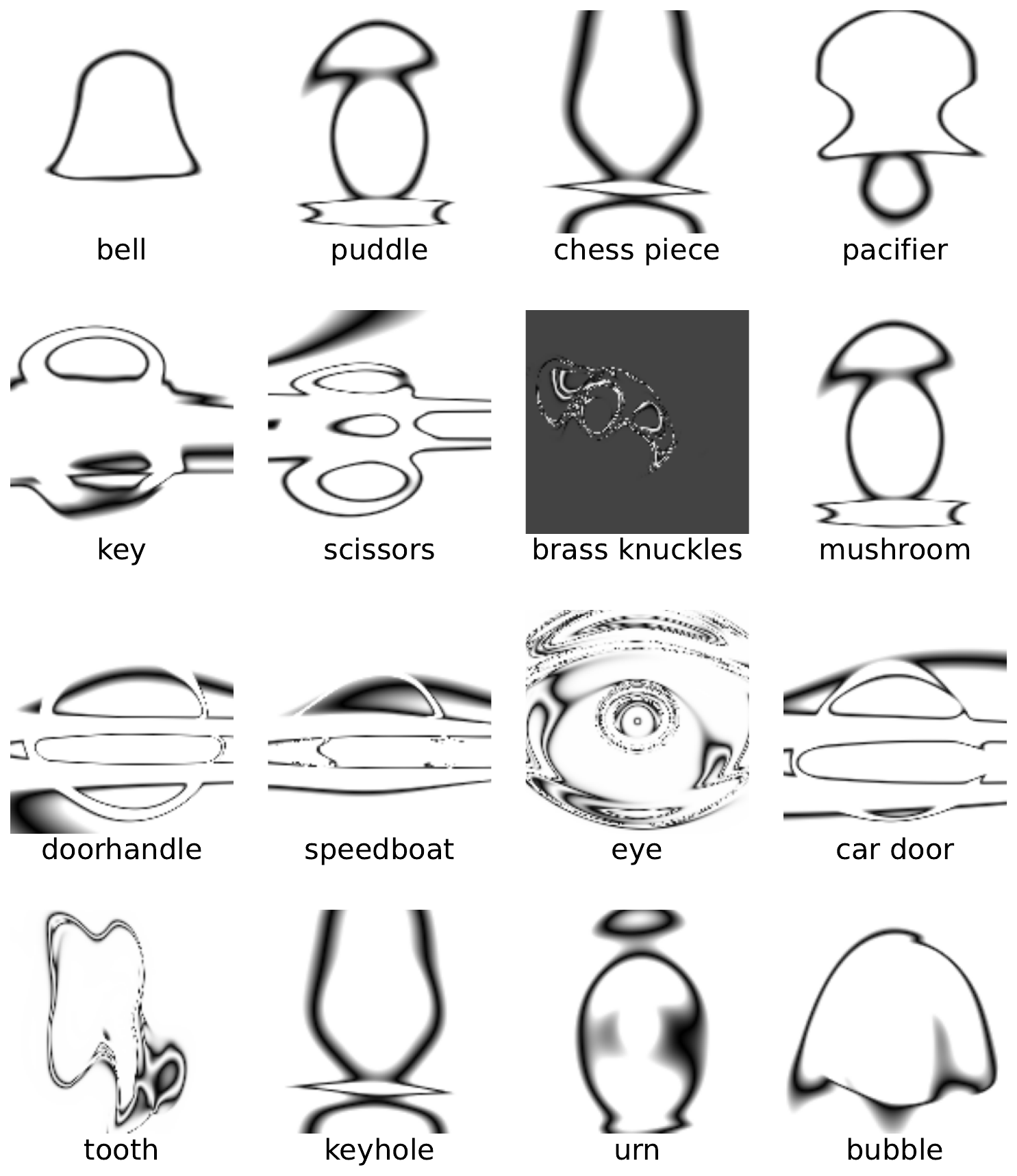}
\caption{$NA = 10$}
\label{fig:traits_grids_nouniest_10}
\end{subfigure}
\hfill
\begin{subfigure}{.31\linewidth}
\centering
\includegraphics[width=\linewidth,fbox={1pt 0pt}]{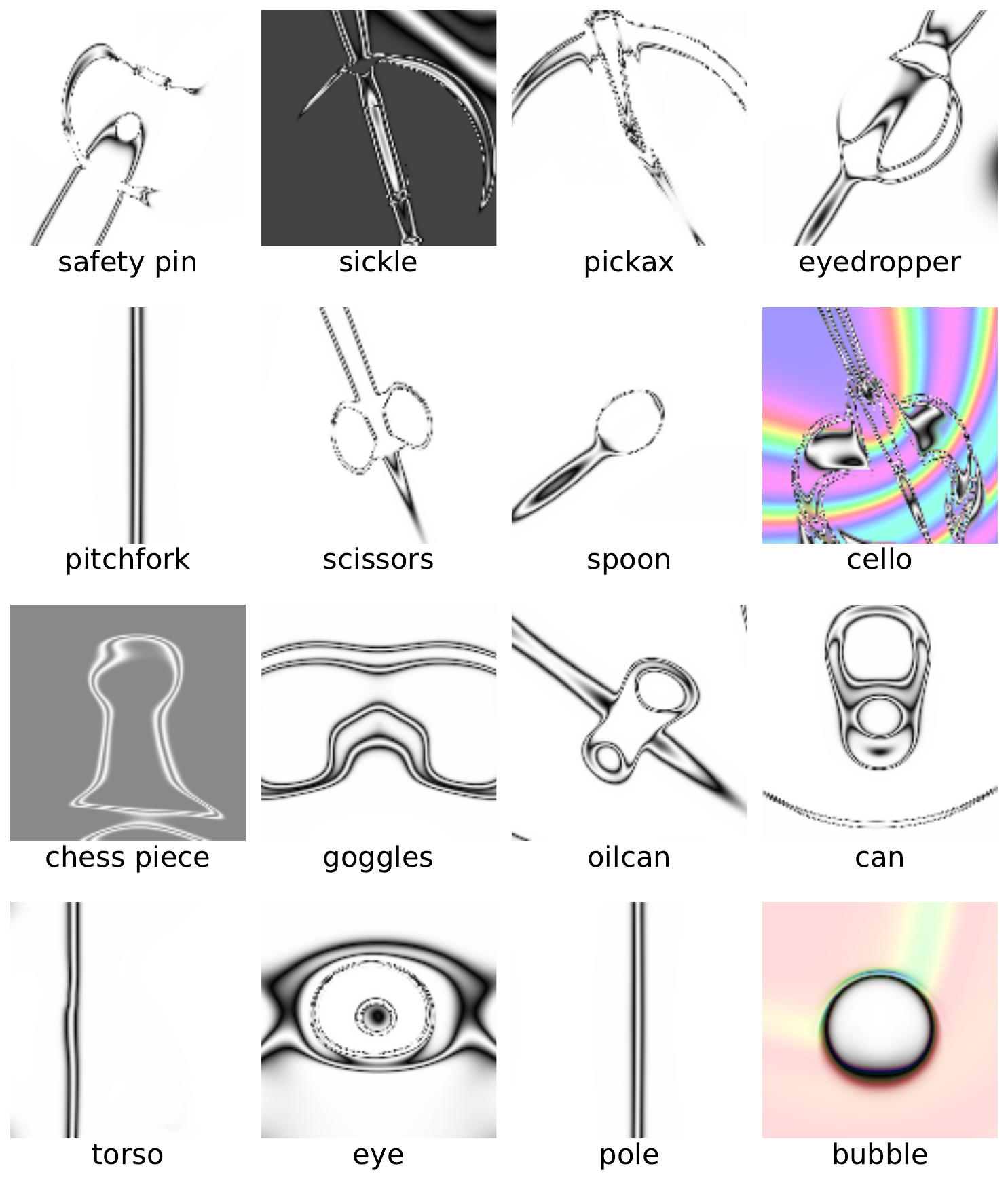}
\caption{$NA = 100$}
\label{fig:traits_grids_nouniest_100}
\end{subfigure}
\hfill
\begin{subfigure}{.31\linewidth}
\centering
\includegraphics[width=\linewidth]{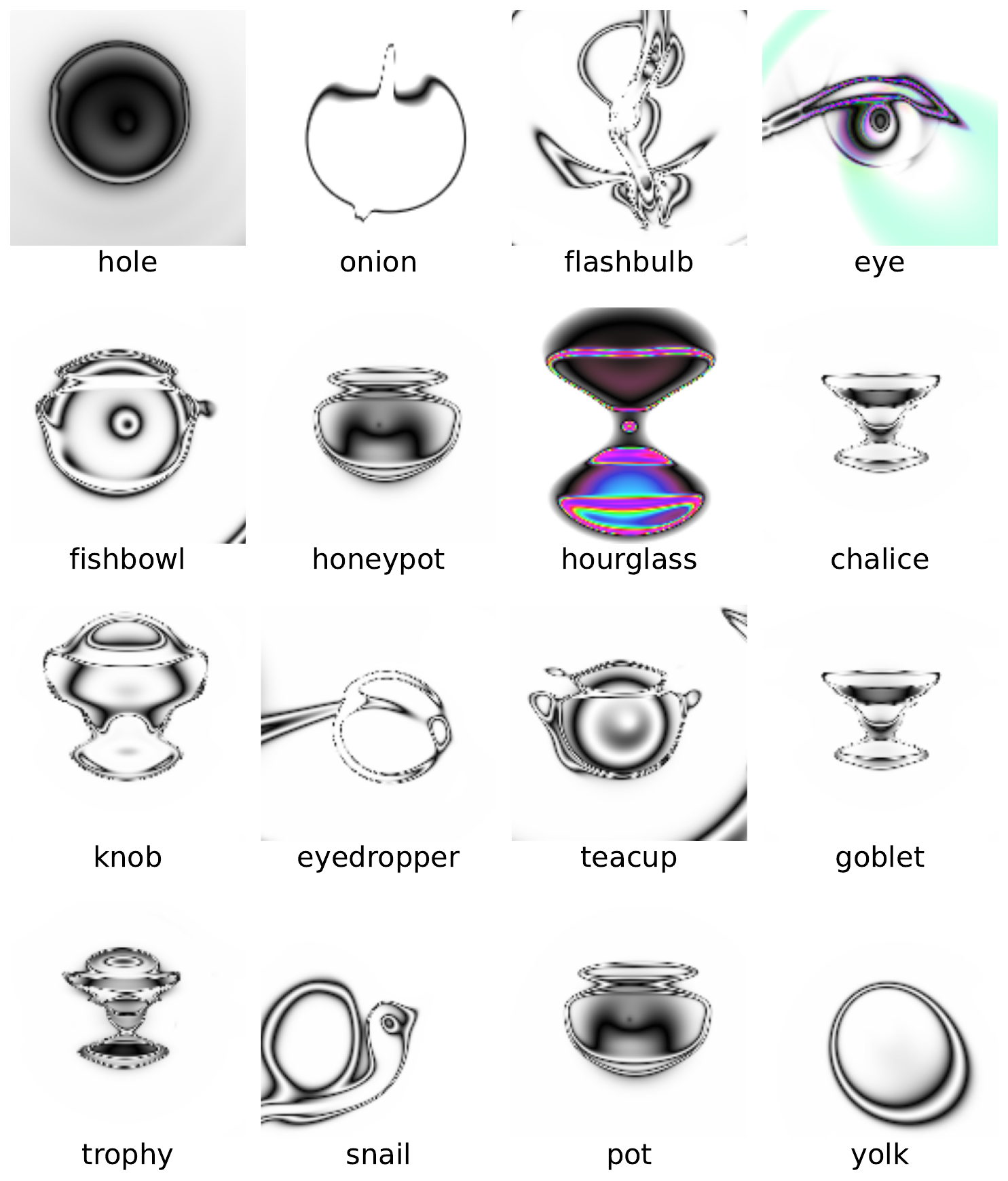}
\caption{$NA = 1,000$}
\label{fig:traits_grids_nouniest_1000}
\end{subfigure}
\caption{
Most semantically salient images in the archive, from seeds with the highest Semantic Recall.
}
\end{subfigure}
\begin{subfigure}{\linewidth}
\label{fig:traits_grids_nouniest}
\centering
\begin{subfigure}{.31\linewidth}
\includegraphics[width=\linewidth]{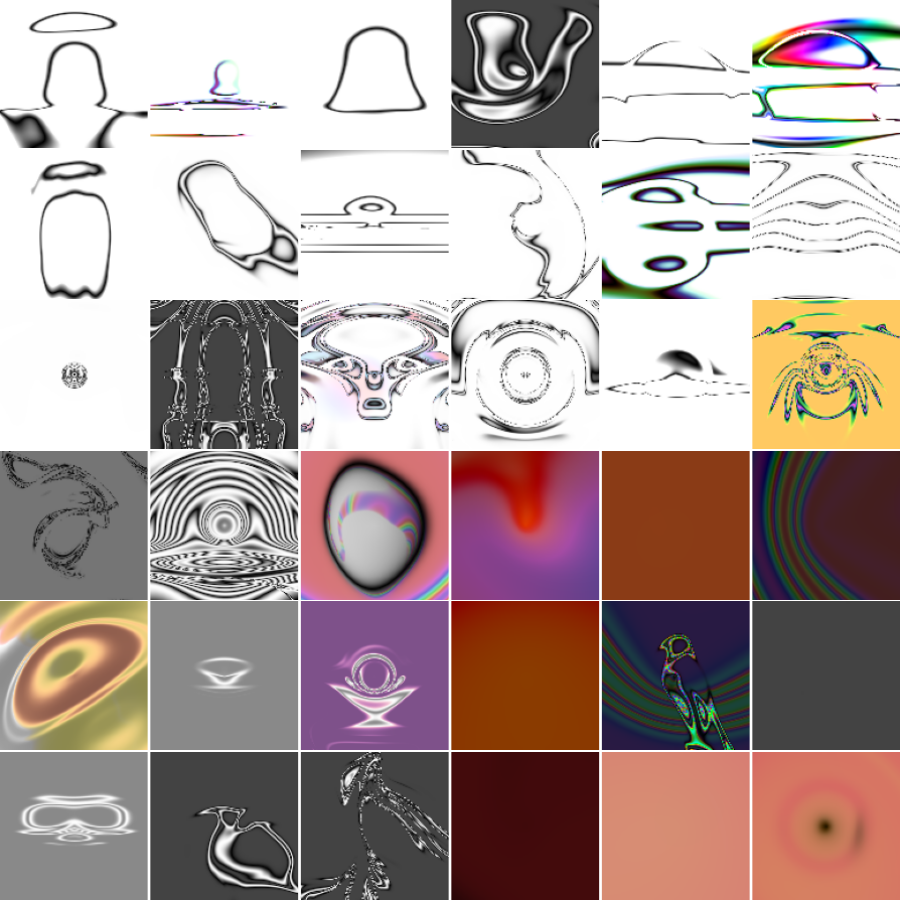}
\caption{$NA = 10$}
\label{fig:traits_grids_representative_10}
\end{subfigure}
\hfill
\begin{subfigure}{.31\linewidth}
\centering
\includegraphics[width=\linewidth]{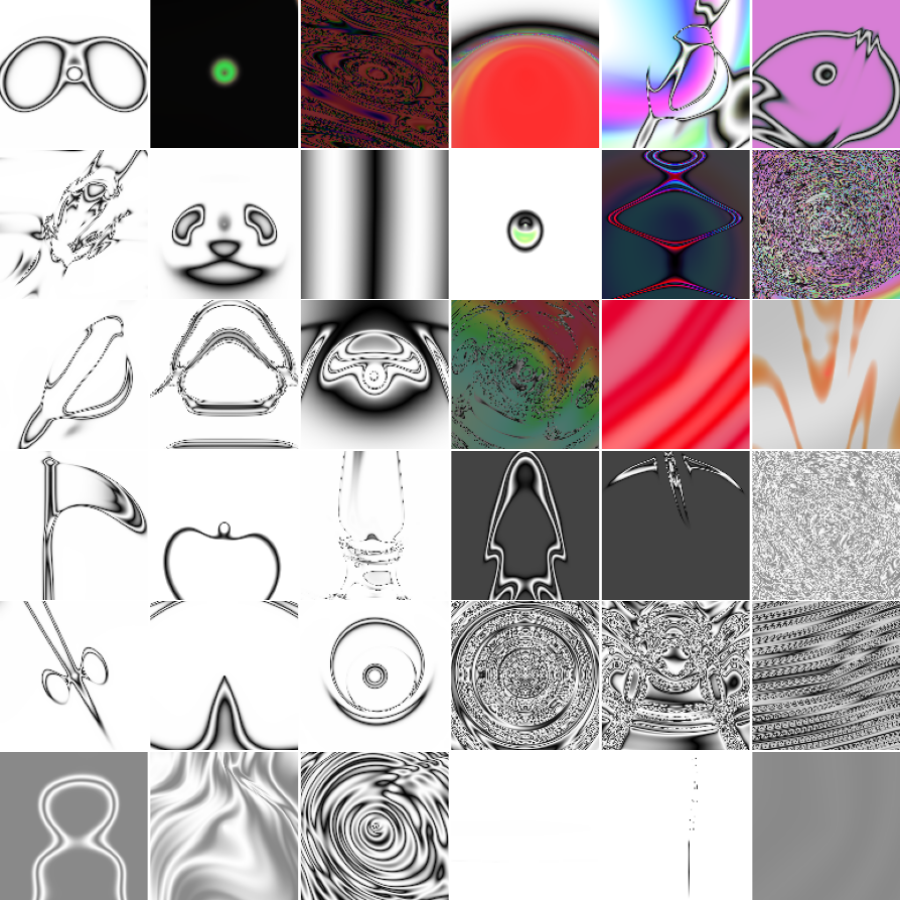}
\caption{$NA = 100$}
\label{fig:traits_grids_representative_100}
\end{subfigure}
\hfill
\begin{subfigure}{.31\linewidth}
\centering
\includegraphics[width=\linewidth,fbox={1pt 0pt}]{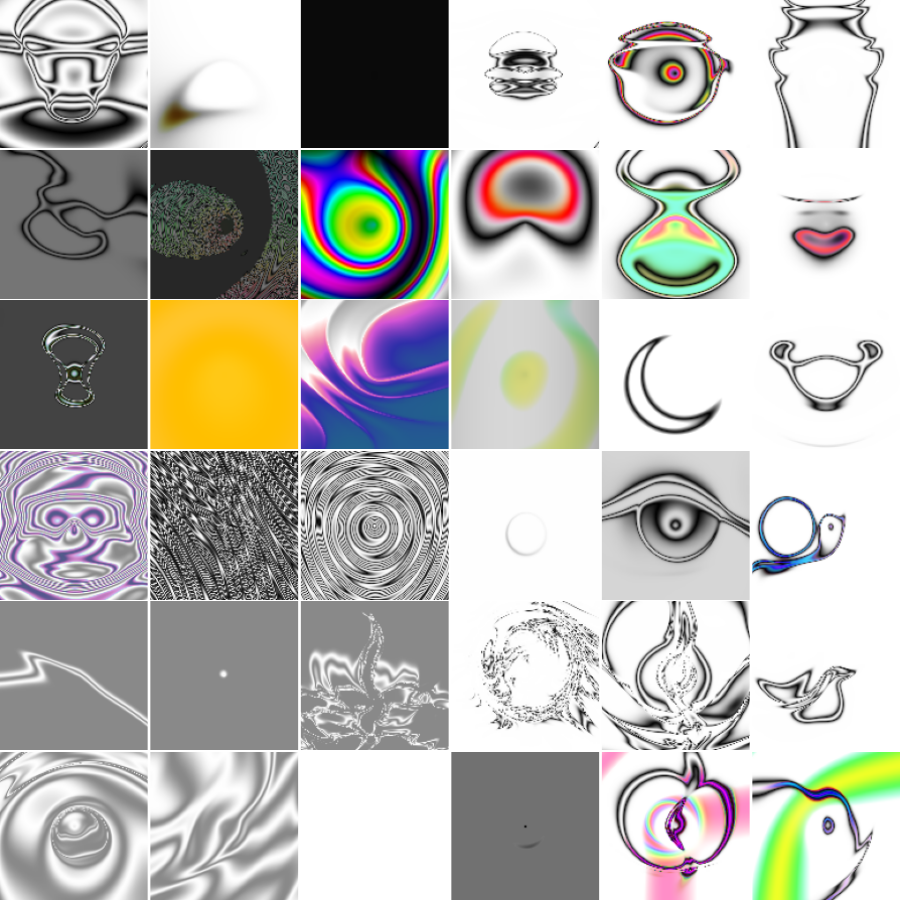}
\caption{$NA = 1,000$}
\label{fig:traits_grids_representative_1000}
\end{subfigure}
\caption{Visually representative images from the archive, from seeds with the highest Visual Coverage.}
\label{fig:traits_grids_representative}
\end{subfigure}
\caption{Qualitative effect of varying the Number of Agents ($NA$), by sampling from variably-sized pools of (LLM-generated) personality traits and prepending these to system prompts during VLM-Picbreeder sessions. Archives with highest Semantic Recall (\ref{fig:traits_grids_nouniest_100}) and Visual Coverage (\ref{fig:traits_grids_representative_1000}) are outlined.}
\label{fig:traits_grids}
\end{figure*}

\section{Related Work}








Picbreeder \citep{secretan2008picbreeder,secretan2011picbreeder} was a tool for interactive evolutionary art generation.
Whereas prior such tools were largely single-user, and user fatigue limited the length of human-in-the-loop evolutionary runs available for study, Picbreeder distributed interaction time across users by allowing them to interactively evolve a shared pool of elites.
Thanks to the strange nature of moving through its space of CPPN-generated images \citep{stanley2007compositional}, notable discoveries often seemed to hinge on moments of serendipity. 
The public experiment left a detailed trace in a single artifact (\autoref{sec:human_data}), provided a relatively constrained action space to users (\autoref{sec:picbreeder_vlm}), and had an open-ended quality about it \citep{stanley2015greatness}.
By automating away its users, we might hope to build salient computational models of intrinsic motivation \citep{oudeyer2007instrinsic}, autotelism \citep{colas2023augmenting}, and creativiy \citep{soros2024creativity}.


Prior work has indeed evolved CPPN-images in this spirit.
But even with images from the human experiment as known targets, single-objective evolution fails \citep{woolley2011deleterious}; though adding quality diversity exploration helps \citep{gaier2019quality}, and gradient descent on a network with predefined topology works well \citep{kumar2025questioning}.
Closest to our work is The Innovation Engine \citep{nguyen2016understanding}, where novelty search is guided by distances in a pre-trained image classifier's latent space.
Our work is similarly bound to a pretrained model, but here the original Picbreeder interaction pattern structures search without any explicit global objective; it is the behavior of agents that shapes discovery.
Relative to this prior work, our system represents a step-change in terms of the quality and interestingness of synthetically generated images.

This is by no small measure owing to the power of LLMs and VLMs, which have recently been deployed extensively in evolutionary and quality diversity search algorithms with notable success---as selection operators, prompted to seek interesting or high-quality artifacts \citep{zhang2024omni,klissarov2023motif,faldor2024omniepic}, as evaluators of behavior, responsible for delineating a space of meaningfully diverse phenotypes \citep{bradley2023quality,pourcel2023aces}, and as mutation operators over complex (language, code) genomes \citep{lehman2023evolution,meyerson2023language}.
Rather than using these models as new building blocks in existing systems, we take a more agentic approach \citep{jiang2026agentic}, and allow them to set their own objectives within a structured sandbox.

\section{Methods}

Our primary aim is to faithfully replicate the human Picbreeder experiment in purely computational form.
We do not seek to necessarily replicate the \textit{results} of Picbreeder---an archive of images, the quality of their representation, or the genetic relationships between them---but rather the \textit{conditions} that enabled open-ended discovery in the original system.
We provide minimal guidance to the VLMs at the helm of our system, instead allowing them to explore according to their own preferences, context, and a brief description of the system's operation.


\subsection{Re-implementing Picbreeder}
Using the neat-python library \citep{McIntyre_neat-python}, we carefully follow Picbreeder's use of using CPPNs \citep{stanley2007compositional} for representing images, and the NeuroEvolution of Augmenting Topologies (NEAT) algorithm \citep{stanley2002evolving} for evolving them.
Each CPPN is a neural network that takes as input an $(x, y, r)$ tuple of coordinates, where $x$ and $y$ are 2D coordinates, and $r$ is the distance from the center of the image (to facilitate radial symmetry).
The CPPN outputs hue, saturation, and brightness for each input tuple; in our experiments, we fix the resolution of generated images to $128\times 128$ during evolution.

One critical nuance in Picbreeder is its handling of structure/color subnetworks within the CPPN.
In initial random networks, the brightness node is given outgoing connections to the hue and saturation nodes.
This biases initial images toward color gradients that follow or reflect grayscale structure.
Activation functions at output nodes are fixed---with sigmoid for brightness, and identity elsewhere---while hidden nodes are assigned random, mutable activations from among sigmoid, sine, cosine, and identity.
To produce a grayscale image (when Picbreeder's ``color mode'' is toggled off), we sample exclusively from the brightness node.
Otherwise, we map the activations of the hue and saturation to $[0, 1]$ by wrapping and clamping, respectively, before conversion to RGB.
Connection weights are marked as belonging to either the structure or color subnetwork.
When the user is in structure- or color-only mutation modes, they may only mutate or add weights belonging to these subnetworks.

\subsection{Historical Picbreeder data}
\label{sec:human_data}

We use the dataset from \citep{kumar2025questioning}, which contains the complete lineages of a large number of the images published to the Picbreeder website between its launch in 2008 and its death around 2016.
This amounts to $9,758$ published images and their ancestry, 
allowing us to reconstruct the full phylogenetic tree of published images for comparison against that of VLM-driven Picbreeder.
Because these lineage files are ordered by publication time, we can retroactively plot various metrics over time---with archive growth---for fine-grained temporal comparison against partial VLM runs (each comprising a few thousand publications, in our experiments).

\subsection{Playing Picbreeder with VLMs}
\label{sec:picbreeder_vlm}

The \textit{session} is the core unit of the Picbreeder loop.
It begins when an agent chooses to either branch an image from the archive
 or begin with a fresh, randomly initialized population of CPPNs.
Following Picbreeder's online interface, the agent may select a single CPPN-image for branching, producing a population of offspring resulting from random mutations of the selected parent.

At the following step, the agent is presented with the resultant initial population---either of mutants resulting from branching, or of random initial CPPN-images---and asked to select one or several as parents for the next generation.
Subsequent steps proceed in the same way, with random mutation---and crossover, in the case of multiple parents---applied to produce the subsequent generation.
Each population comprises 15 CPPN-images; exact copies of parent(s) are always included among their offspring.
At the 20\textsuperscript{th} generation of evolution in the session, the agent is asked to select an image for publication to the archive and give it a title.\footnote{We initially allowed agents more freedom over when to publish, but agents chose to publish rapidly and end their session, hence we enforced longer sessions. The selected length of $20$ generations is just shy of the human mean (\autoref{fig:session_length_dist}, \autoref{tab:session_length_stats}).}

At each step, the agent may
alternatively toggle color mode.
Random initial populations default to grayscale, and branched images retain the color mode under which they were published.
When selecting parents, the agent may additionally adjust the strength of mutation (which defaults to $0.5$) to any value in $[0.01,1]$.
When color mode is on, the agent may toggle mutation mode between color- or structure-only, or both (when color mode is off, mutations default to structure-only).
All of these controls and their defaults mirror those of the original Picbreeder interface.

\begin{figure}
\begin{subfigure}{1\linewidth}
\centering
\includegraphics[width=.61\linewidth]{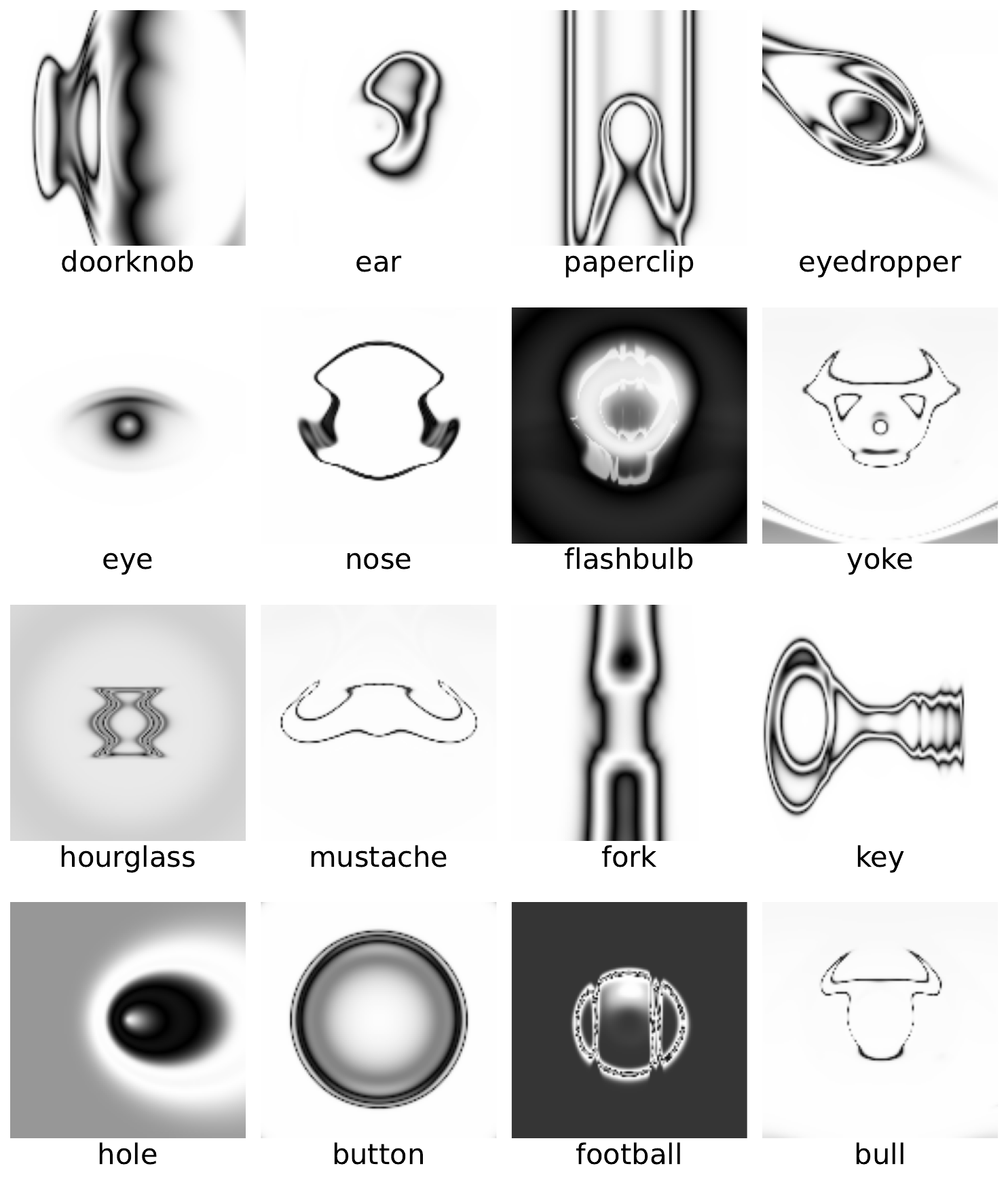}
\caption{Most semantically salient images}
\label{fig:human_grids_nouniest}
\end{subfigure}
\begin{subfigure}{1\linewidth}
\centering
\includegraphics[width=.61\linewidth]{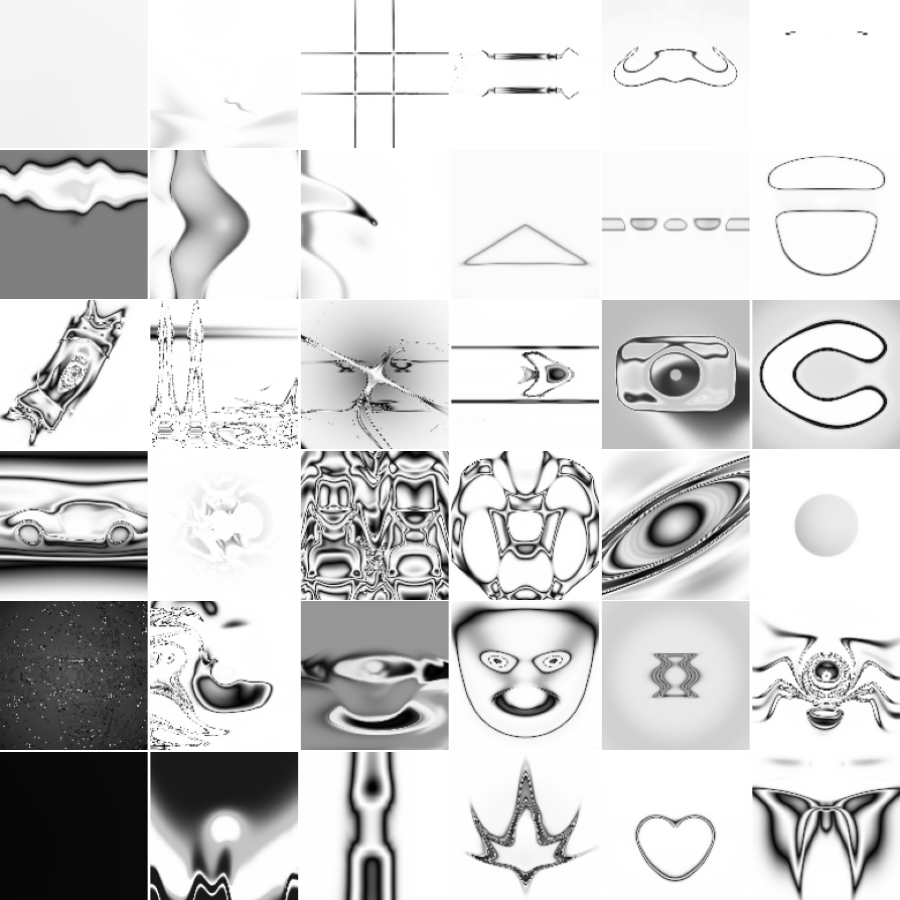}
\caption{Visually representative images}
\label{fig:human_grids_representative}
\end{subfigure}
\caption{Samples from the historical human archive after 2,000 user sessions.}
\label{fig:human_grids}
\end{figure}

During branching, agents are presented with a $100$-image sample of the archive of images published thus far.
This is broken down in the following $5$ equal parts: a set of ``top-rated'' images, having accrued the highest (VLM-generated) ratings; a set of ``best new'' images, comprising the top-rated of the $100$ most recently published; a set of ``most branched'' images, comprising those most selected for branching; a set of ``latest'' images, comprising those most recently published; and a set of ``random'' images, drawn uniformly from over the entire archive.
These sets are mutually exclusive and generated in this order, such that a member of an earlier set cannot also belong to a later one (e.g. a ``top-rated'' image may not also appear as a ``best new'' image).
This mimics the categories of images presented to users on the home screen of the original Picbreeder website (\autoref{fig:picbreeder_homepage}; save for the ``editor's choice'' category).
While these initial sets of images were smaller on the original website ($8$ in each), human users were able to selectively ``see more'' of a given category; we split the difference by presenting larger static sets.

To solicit ratings for each image published to the collaborative archive, we ask a new VLM instance (with no context of prior interactions with the system) to rate a subset of images from the archive with integer scores between $1$ and $5$.
This sample is drawn in the same way as the branching sample described above.
This ensures that newly rated images are compared against a variety of existing images in the archive.
We initiate this rating process after every $5$ new publications, once the archive has reached a size of at least $100$ images.
During branching, selection, publication, and rating, agents are prompted to provide 1-2 sentences of rationale for their decisions, allowing us a (potentially fallible) window into their decision-making process, and to observe failure modes, e.g., where agents may become overwhelmed by their context and begin mixing up the positions of images or mischaracterizing what these images depict.
We run 10 VLM agents in parallel, which branch from, publish to, and rate a shared archive.

\section{Metrics \& Interventions}

\begin{figure}
\begin{subfigure}{0.88\linewidth}
\includegraphics[width=\linewidth,trim={0 45 0 45},clip]{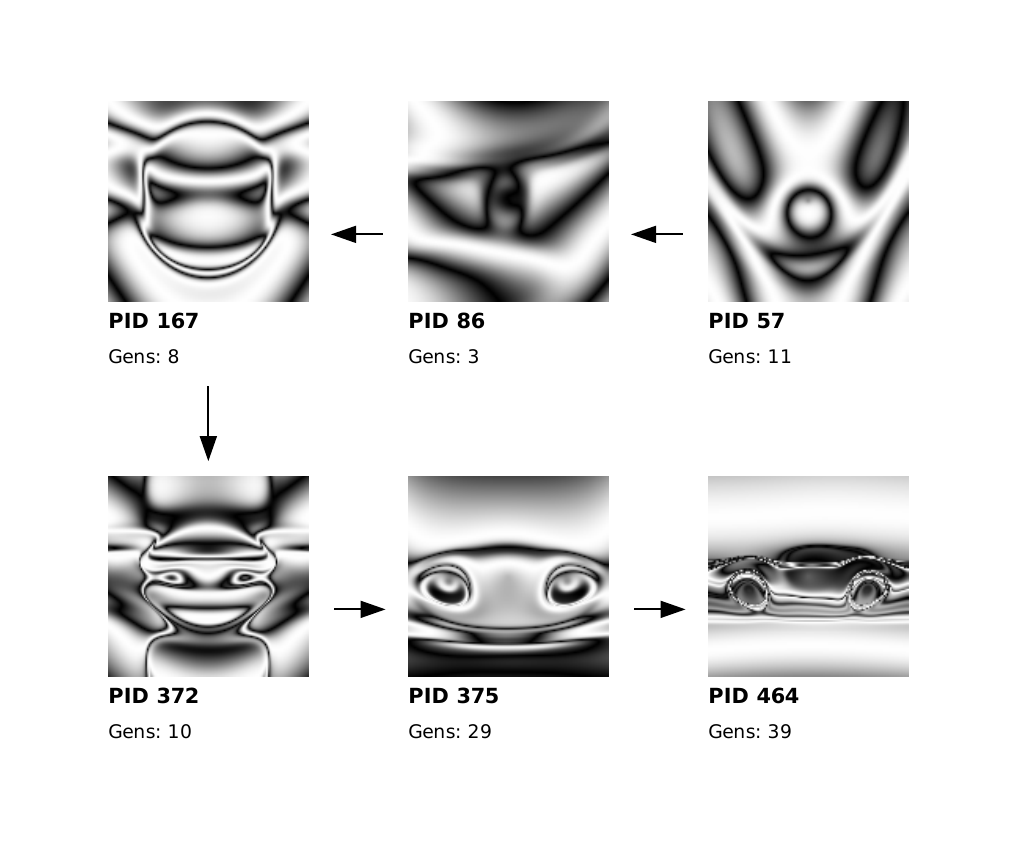}
\caption{Lineage of human-discovered car}
\end{subfigure}
\begin{subfigure}{0.88\linewidth}
\vspace{15pt}
\includegraphics[width=\linewidth,trim={0 45 0 45},clip]{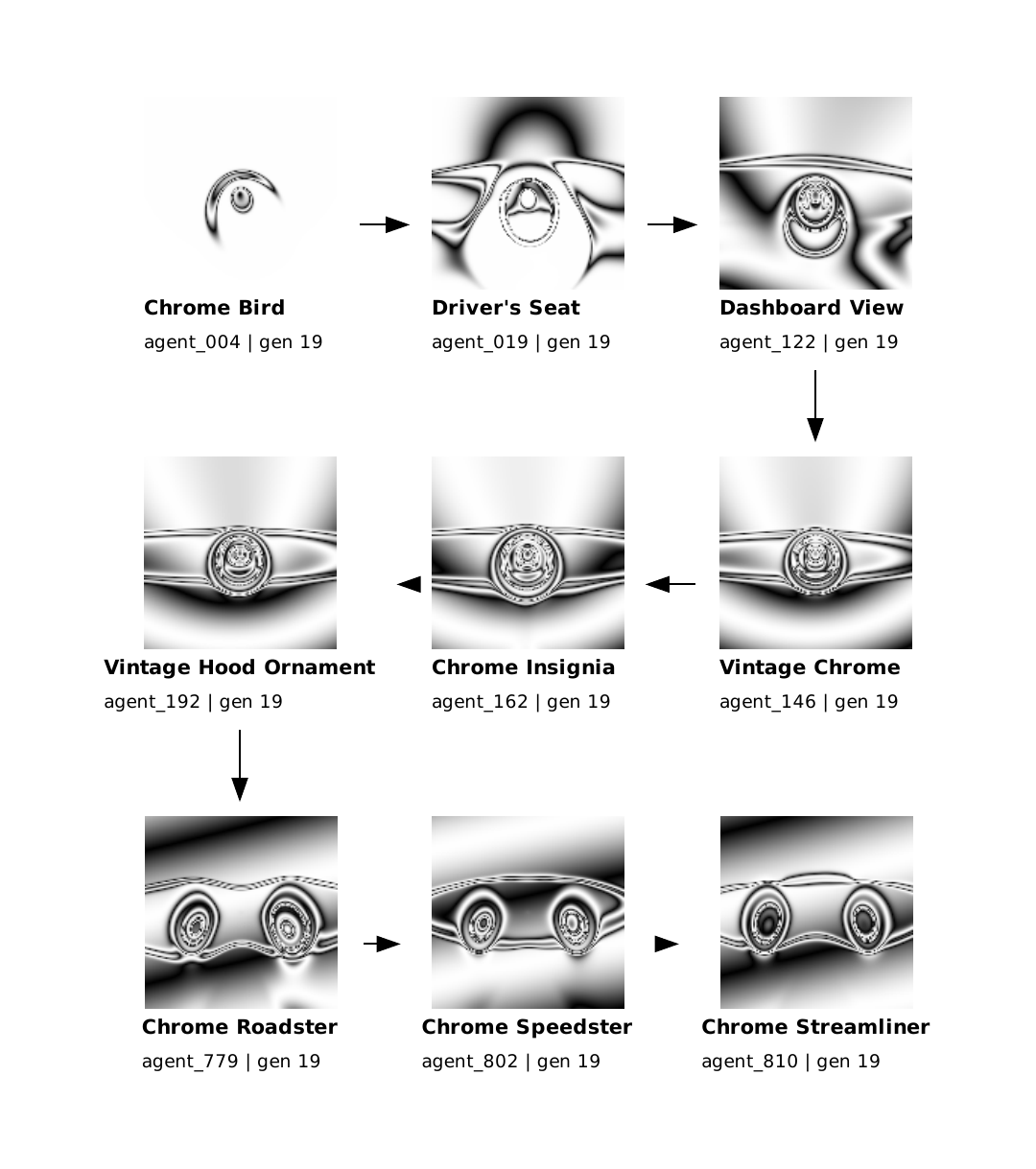}
\caption{Lineage of VLM-discovered car}
\end{subfigure}
\caption{
Ancestral lineages of semantically similar images generated by human and VLM   interactions with Picbreeder (only published images are displayed). Humans seem to take larger steps in semantic space (from face, to eye, to frog, to car) than VLMs (from abstract bird, to car seat, to dashboard, to hood ornament, to car).
}
\label{fig:lineages}
\end{figure}

Instinctively, the historical human archive is more evocative and diverse than what is produced by VLMs.
We can see this both by comparing (representative samples of) archives from either progeny (\autoref{fig:traits_grids}, \autoref{fig:human_grids}); and comparing ancestral lineages within them, tracing back from two quite similar images---an instance of convergent human- and VLM-driven evolution (\autoref{fig:lineages}).
Humans tend to take bigger leaps between publications and land on sharper, more refined images.
But why?
What is this \emph{x-factor}---this quality of boldness and discernment?
In this section, we establish evaluation metrics that attempt to quantify this x-factor, and experimental interventions that might allow us to come closer to replicating it synthetically.

\subsection{Evaluation Metrics}

Our evaluation metrics comprise attempts to abstract and mechanize what we think gives the human archive its special quality.
In particular, we try to measure the degree of fidelity with which agents are able to depict a diverse set of visually/semantically distinct forms.
We can think of these metrics roughly as capturing the quality (recall) and diversity (k-covering radii in embedding spaces and $J^1$ index of Tree Balance) of the Picbreeder archive.

\subsubsection{Semantic Recall}

To measure Semantic Recall---the system's capacity to rediscover a set of known common objects---we gather a large list of nouns/noun phrases that can plausibly be depicted as images. Here, we use the $1,824$ unique class names in the THINGS dataset \citep{hebart2023things}, after deduplication.
In a joint text-vision embedding space (we use SigLIP2-B \citep{tschannen2025siglip}), we embed each of these classes, and each of the images published to the run of an archive during a (human- or VLM-driven) Picbreeder run.
We then compute the cosine distance between each image and each class in this embedding space.
For each class, we take the minimum distance between it and any image, then sum this value over all classes.

\begin{table*}[]
\begin{tabular}{llcccc}
\toprule
Sweep & Condition & Semantic Recall & Visual Coverage & Semantic Coverage & Tree Balance ($J^1$) \\
\midrule
\multirow{6}{*}{Random Action Prob. ($\epsilon$)} & \cellcolor[gray]{0.9}0.0 & \cellcolor[gray]{0.9}0.087 $\pm$ 0.001 & \cellcolor[gray]{0.9}0.614 $\pm$ 0.002 & \cellcolor[gray]{0.9}0.696 $\pm$ 0.004 & \cellcolor[gray]{0.9}0.235 $\pm$ 0.026 \\
 & 0.05 & 0.086 $\pm$ 0.001 & 0.619 $\pm$ 0.011 & 0.702 $\pm$ 0.006 & 0.246 $\pm$ 0.023 \\
 & 0.25 & \textbf{0.088 $\pm$ 0.001} & 0.638 $\pm$ 0.009 & \textbf{0.717 $\pm$ 0.004} & 0.249 $\pm$ 0.022 \\
 & 0.5 & 0.085 $\pm$ 0.001 & 0.633 $\pm$ 0.007 & 0.709 $\pm$ 0.005 & 0.260 $\pm$ 0.013 \\
 & 0.75 & 0.084 $\pm$ 0.001 & \textbf{0.639 $\pm$ 0.004} & 0.706 $\pm$ 0.003 & \textbf{0.303 $\pm$ 0.010} \\
 & 1.0 & 0.082 $\pm$ 0.001 & 0.610 $\pm$ 0.013 & 0.700 $\pm$ 0.003 & 0.275 $\pm$ 0.030 \\
\midrule
\multirow{5}{*}{Context Length ($CL$)} & 0 & 0.082 $\pm$ 0.001 & 0.527 $\pm$ 0.009 & 0.632 $\pm$ 0.006 & 0.305 $\pm$ 0.027 \\
 & \cellcolor[gray]{0.9}1 & \cellcolor[gray]{0.9}\textbf{0.087 $\pm$ 0.001} & \cellcolor[gray]{0.9}\textbf{0.614 $\pm$ 0.002} & \cellcolor[gray]{0.9}0.696 $\pm$ 0.004 & \cellcolor[gray]{0.9}0.235 $\pm$ 0.026 \\
 & 2 & 0.083 $\pm$ 0.001 & 0.583 $\pm$ 0.004 & 0.675 $\pm$ 0.006 & 0.339 $\pm$ 0.033 \\
 & 10 & 0.079 $\pm$ 0.003 & 0.512 $\pm$ 0.021 & 0.661 $\pm$ 0.010 & 0.331 $\pm$ 0.040 \\
 & 20 (Full) & 0.083 $\pm$ 0.001 & 0.595 $\pm$ 0.010 & \textbf{0.697 $\pm$ 0.003} & \textbf{0.350 $\pm$ 0.018} \\
\midrule
\multirow{4}{*}{Num. Agents ($NA$)} & \cellcolor[gray]{0.9}0 & \cellcolor[gray]{0.9}0.087 $\pm$ 0.001 & \cellcolor[gray]{0.9}0.614 $\pm$ 0.002 & \cellcolor[gray]{0.9}0.696 $\pm$ 0.004 & \cellcolor[gray]{0.9}0.235 $\pm$ 0.026 \\
 & 10 & 0.086 $\pm$ 0.002 & 0.605 $\pm$ 0.012 & 0.698 $\pm$ 0.011 & 0.373 $\pm$ 0.030 \\
 & 100 & \textbf{0.089 $\pm$ 0.001} & 0.659 $\pm$ 0.006 & 0.710 $\pm$ 0.009 & 0.473 $\pm$ 0.022 \\
 & 1000 & 0.088 $\pm$ 0.001 & \textbf{0.665 $\pm$ 0.004} & \cellcolor[rgb]{0.75, 1, 0.75}{\textbf{0.734 $\pm$ 0.013}} & \textbf{0.476 $\pm$ 0.013} \\
\midrule
\multirow{2}{*}{Baselines} & Random & 0.080 $\pm$ 0.001 & 0.612 $\pm$ 0.005 & 0.692 $\pm$ 0.002 & \cellcolor[rgb]{0.75, 1, 0.75}{0.540 $\pm$ 0.003} \\
 & Human & \cellcolor[rgb]{0.75, 1, 0.75}{0.089 $\pm$ 0.000} & \cellcolor[rgb]{0.75, 1, 0.75}{0.681 $\pm$ 0.000} & 0.730 $\pm$ 0.000 & 0.363 $\pm$ 0.000 \\
\bottomrule
\end{tabular}
\caption{
Summary of results of various interventions on VLM-driven Picbreeder in terms of their impact on our evaluation metrics (mean $\pm$ standard error).
Overall best results highlighted in green; best results among each hyperparameter sweep appear in bold. The default setting---recurring across sweeps---is highlighted in grey.}
\label{tab:results}
\end{table*}

\subsubsection{Visual Novelty}

To measure the visual novelty of the generated archive, we embed all published images into an image embedding space (we use SigLIP-2-B-alignet \citep{muttenthaler2025aligning}).
We then use greedy farthest-point sampling \citep{gonzalez1985clustering}---starting with a set that includes a random datapoint, then repeatedly adding to this set the point that has the greatest minimum distance with any point in the set---to generate a set of $k$ representative images.
Given these $k$ representatives, we then grow spheres about these points with equal radii, until all embedded images are contained within some sphere.
The resultant radius is the $k$-covering radius (we report results for $k=100$).
To visualize archives (or $k$ representative points from these archives), we use Rasterfairy \citep{rasterfairy} to render images in a rectangular grid, arranged therein to reflect their relative distances in embedding space, allowing for more visually intuitive, ordered snapshots of the archive.

\subsubsection{Semantic Novelty}

To measure the semantic novelty of the generated archive---the number of semantically distinct, potentially novel forms---we have a VLM (gemini-2.5-pro) generate short (1-sentence) captions for each image in the archive, then map these captions to a text embedding space (gemini-embedding-001 \citep{lee2025gemini}) and measure 
the $k$-covering radius at $k=100$
over these points.

\subsubsection{Analysis of Phylogenetic Trees}

We construct the phylogenetic trees of all published images, treating as roots those images that resulted from sessions that started from a random initial population of CPPN-images, and defining child-parent relationships between any published image (child) that resulted from a session that began by branching a previously-published image (parent).
We then compute the $J^1$ index, a robust measure of Tree Balance \citep{lemant2022robust}.

\subsection{Experimental Interventions}

Our interventions on the VLM-driven Picbreeder pipeline derive from reasoning about what might give humans an edge over VLMs in matters of creative, open-ended discovery. 
We identify a few candidates, and propose analogous components in the VLM agent that might be modulated to recapture this edge. 

\subsubsection{Memory and Context Length}

First, the way in which humans \textit{remember} their recent past experiences would seem quite distinct from either of the two modes VLMs have for doing the same, namely, via storage in their weights (during training) or in their context (during inference).
When a human plays Picbreeder, they are inevitably exposed to a sample of the online archive before embarking on an evolutionary session, and this impression will surely influence their judgements about the novelty of newly-generated images.
The same goes for their selections and all candidate CPPN-images viewed during their session.
But these past impressions are neither ``wired'' in their synapses, nor continually presented to them all-at-once on a single display.

Still, we wonder what effect \textit{memory} might have on our VLM Picbreeder agents, and ask this question by simulating something like the latter case---controlling the number of previous system interaction steps that are included in a VLM's context.
In the simplest setting, the VLM is only presented with the current population, and cannot see any prior populations or its decisions pertaining to these---we refer to this as a \textit{Context Length} ($CL$) of 0.
In the default setting, with $CL=1$, the agent can see the current turn and the one prior.
In general, when the branching step is captured by the context window as defined here (e.g., when an agent with $CL=1$ is faced with its first population after branching), the agent is able to see this sample of the online archive.
The context window never extends back beyond the branching step---every user session belongs to a ``fresh'' instance of the VLM.

We select $CL=1$ as our default because it is maximally cheap, 
while still giving the agent the opportunity to notice if it might be stuck in a local minimum, repeating its selection decision across multiple turns.
At the extreme of maximum rememberance, $CL=20$, the agent's context will always include the full chat history of the current session, including its branching step.
In this case, we append a special directive to the prompt which asks the agent, when justifying its publication decision, to additionally explain how this publication is novel with respect to the initial archive sample (\autoref{fig:prompt_addendum}).
This is motivated by the empirical observation that agents with less context often publish identical copies, or minor variations of, the same image dozens of times.

We expect that larger context lengths should help agents escape local minima and implicitly incentivize them to explore more aggressively.
On the other hand, we're wary of the possibility that overwhelming agents with excessively lengthy prompts might limit their ability to effectively reason and discern among images.

\subsubsection{Exploration and Selection Noise}

In human decision-making, given the inherent noisiness and complexity of the physical world, a great many auxiliary random variables may come into play.
This may be all the more consequential in Picbreeder, where the CPPN mutations affected by the user are themselves noisy, and the interactive evolutionary process is inherently difficult to control.
In such noisy domains, users are liable to exert less effort in long-term planning~\citep{lei2025human}, and may therefore be more likely to make decisions ``on a whim'' in a quasi-random fashion.
The machinery of large neural networks, by contrast, elides much of the complexity of the physical world, and LLMs may be even more limited than humans in their ability to emulate true randomness \citep{harrison2024comparison,van2024random}.
It's also conceivable that they may be less prone to make decisions ``on a whim'' in general, given their having been fine-tuned to be maximally helpful in domains with verifiable solutions.


We therefore use an $\epsilon$-greedy exploration strategy \citep{sutton1998reinforcement} to inject randomness into the agents' selection process.
The hope is that this will facilitate productive exploration, allowing the agent to escape attractors in search space.
Conversely, an over-abundance of such noise will presumably lead to an archive of similarly noisy, meaningless images.

Concretely, at each selection step, with probability $\epsilon$, the VLM query is replaced with a random action, viz. the uniform random selection of a parent from the current generation and/or the random adjustment of settings (color toggle; mutation mode and strength) that would otherwise have been available to the VLM.
During an $\epsilon$-random action, color mode is toggled (forgoing the selection of a random parent) with probability 0.1.
With probability 0.2, (provided color mode is active) a new mutation mode will be uniform randomly selected from the set of 3 possible such modes.
And with probability 0.2, a new continuous mutation strength will be selected from a uniform continuous distribution over $[0,1]$.
These random actions can only occur during selection steps; even when $\epsilon=1$, a VLM is queried to make branching and rating decisions.

\subsubsection{Multiple Agents and Promptable Inclinations}

Human populations are diverse, and this diversity likely translates to myriad distinct Picbreeder ``playstyles''.
Anecdotally, one prolific user of the original Picbreeder system (``BurnedDirt'') fixated on minimal forms---clean basic shapes, patterns of straight lines---
while other users tended toward forms resembling insects or faces.
Likely thousands of distinct human users engaged with Picbreeder, but we'd be hard-pressed to find a comparable variety of distinct VLMs to deploy in our system in parallel.

We opt instead to simply prompt for diverse play-styles.
To this end, we feed the Picbreeder system prompt (which details the system and its interface for VLM agents, \autoref{fig:prompt_system}) to an LLM (gemini-3-pro-preview) and ask it to come up with distinct personality traits that may indirectly affect a user's behavior in such a system, while avoiding the specification of concrete objectives (\autoref{fig:traits_prompt}) in batches of 50 traits, up to a total of 1,000 traits, with the LLM viewing its previous 10 batches of generated traits at each step.
We then run experiments controlling for the number of such traits that we use to parameterize distinct ``agents'' during a VLM-Picbreeder run, randomly selecting this many traits at the beginning of the experiment from the overall pool.
At the beginning of each agent session, a personality prompt is drawn at random from this subset, and prepended to that agent's system prompt for the duration of their session.
A random sample of the generated prompts is given in \autoref{tab:traits}.

\section{Results \& Discussion}

Images in the human Picbreeder archive tend to be diverse, aesthetically refined, and often evocative (cf. Figs. \ref{fig:traits_grids} and \ref{fig:human_grids}).
Our evaluation metrics appear to reflect aspects of this qualitative discrepancy, with the historical human baseline dominating most (\autoref{tab:results}).
In the following sections, we detail these results, and describe correlations between changes in our evaluation metrics and qualitative changes in generated archives, with reference to additional qualitative archive samples and quantitative visualizations (Figs. \ref{fig:exploration}-\ref{fig:traits}) in the Appendix.

A fully random baseline (in which all selections, branching, and publication decisions are sampled uniformly) acts as a lower bound, achieving low scores on all metrics, save for Tree Balance (since uniform random branching decisions lead to highly balanced trees in expectation).

In our VLM experiments, we default to Context Length $CL = 1$, random selection probability $\epsilon = 0$, and Number of Agents $NA = 1$.
We run each experiment for a total of $2,000$ sessions (resulting in an archive of as many images), repeat each experiment with $6$ random seeds, and report/plot the mean and its standard error over these seeds.

Qualitatively distinct design patterns result from different models, even within the same family (\autoref{fig:model_grids}). Perhaps surprisingly, gemini-3-pro underperforms 2.5 qualitatively and quantitatively (\autoref{fig:model_recall}), developing deeper perfectionist fixations, while 2.5-flash and flash-lite tend increasingly toward high-frequency noise.
qwen3-vl-8b and qwen3-vl-30b-fp8 \citep{Qwen3-VL}---which we run locally \citep{kwon2023efficient}---collapse completely toward it.
We use gemini-2.5-pro \citep{comanici2025gemini} as the default for our experiments.

\subsection{Exploration}

Without exploratory noise ($\epsilon=0$), the VLM-generated archives are exceedingly likely to contain many dozen insignificant variations of the same form.
We see this quantitatively (\autoref{fig:exploration_semantic_coverage}), where $\epsilon=0$ is nearly as weak as Random in terms of Semantic Coverage; and in the visually representative sample from such an archive in \autoref{fig:exploration_grids_representative_epsilon_0}, with the repetition of a few fox- and fishbone-like forms in particular (see also \autoref{fig:attractor_grids}).\footnote{Since these representatives are chosen to be maximally distant from one another from among the set of images, this indicates that an abundance of closely related forms to these exemplars live in the archive (mode collapse).}

The use of noise in the selection process encourages exploration and seems to avoid such mode collapse (cf. the relatively diverse representatives in \autoref{fig:exploration_grids_representative_epsilon_0_25}), but this comes at a direct tradeoff w.r.t. the legibility of generated images: when $\epsilon \leq 0.25$, Semantic Recall is roughly comparable to that of $\epsilon=0$, though recall suffers under larger $\epsilon$ (\autoref{fig:exploration_semantic_coverage}). 
Though these more exploratory settings generate a greater diversity of images---as evidenced by their high visual (\autoref{fig:exploration_visual_coverage}) and semantic (\autoref{fig:exploration_semantic_coverage}) coverage---they tend to be less sharp and refined than in their noiseless counterparts.

A setting of $\epsilon=1$ drastically drops Semantic Recall score, but not to the point of matching Random, and in \autoref{fig:exploration_grids_representative_epsilon_1} we indeed note a few recognizable or interesting forms.
That is, even when VLMs are confined to only branching and rating, they can steer the evolutionary process (though perhaps very slowly) toward meaningful artifacts.\footnote{One way of thinking about $\epsilon$ is that, roughly, it increases mutation strength and lessens session length, so that when $\epsilon=1$, the agent makes only one selection choice---during branching---then the remaining 20 generations of random selection are akin to one very drastic mutation step.}

In general, we note how very imbalanced VLM-generated phylogenetic trees are relative to the human baseline (\autoref{fig:exploration_j1_index}), suggesting that VLMs tend to be far more homogeneous in their branching decisions---viz. prone to repeatedly branch from the same image(s), and less likely to start from random initial populations---as compared to human users.
Ramping up $\epsilon$ only slightly increases Tree Balance.
So, while increasing $\epsilon$ leads to individuals \textit{looking} more diverse, these individuals still tend to be related to one another.
In other words, exploratory noise increases phylogenetic diversity, but not  genetic diversity (where the latter might arguably lead to deeper and more meaningful long-term variation).
It's interesting that this selection bias is robust to noise in the first place; that even from among more diverse sets of images (and in the extreme, sets of highly noisy images), the VLM always has a clear favorite.
This kind of stubborn favoritism could be a barrier to using VLMs' as an engine of open-ended search.

\subsection{History}

Removing all history beyond the current turn ($CL = 0$) collapses Semantic Recall (\autoref{fig:history_semantic_recall}), because agents in this setting are prone to publishing duplicates (note the redundancy among visual representatives in \autoref{fig:history_grids_representative_0}).
Setting $CL = 1$ proves surprisingly effective in terms of Semantic Recall, in spite of the relative homogeneity of the archive (as discussed above when $\epsilon=1$, these two settings corresponding to the same set of experiments under default hyperparameters).
Clearly, the homogeneity of the archive when $CL = 1$ is lesser than when $CL=0$, both in terms of visual (\autoref{fig:history_visual_coverage}) and semantic (\autoref{fig:history_semantic_coverage}) diversity, suggesting that a little context goes a long way in preventing the agent from spiraling into a pattern of repeated or near-identical selections.

But such benefits neither scale nor are sustained: even incrementing to $CL = 2$ results in a large hit to Semantic Recall with hardly any gain in terms of diversity.
Indeed, the archives produced when $CL = 2$ remain similarly homogenous, though the forms appear slightly less refined, and messier, presumably owing to the detrimental effect of information-overload in the agent's context.
Increasing context to $CL=10$ again results in a sharp decrease in Semantic Recall, as well a decrease in diversity.
But it's not quite the case that the published images under $CL=10$ simply become messier: some are quite refined, and in particular we see a huge number of duplicate entries of near-photorealistic top-down views of soda cans across multiple seeds (\autoref{fig:soda}), these being almost entirely absent from runs with other $CL$.
It may be that given these larger contexts, the agent begins to fall into auto-sycophantic loops which reinforce its own predetermined objectives, collapsing diversity but sometimes resulting in a handful of refined forms.

With full history, $CL=20$, diversity/coverage scores reach an apex relative to other context lengths.
This may well be attributable to the additional, novelty-pleading prompt appended in these cases (\autoref{fig:prompt_addendum}).
But Semantic Recall, while improving over $CL=10$, does not reach the height of $CL=1$, likely owing to the often abstract and high-frequency publications in these archives.
In general, we note that increasing context pushes the agent toward busier, apparently more complex (and sometimes downright noisy) images, while lower $CL$ tends to elicit starker forms, with archives often nearly entirely absent of color.

\subsection{Multiple Agents}

Adding agents (via LLM-generated behavioral idiosyncracies) does not noticeably improve recall (\autoref{fig:traits_semantic_recall}), but results in a marked improvement in the archive's diversity in terms of visual and Semantic Coverage, in addition to Tree Balance (\autoref{fig:traits_diversity}) at high agent counts.

A small number of agents ($NA=10$) results in a drop in Visual Coverage (\autoref{fig:traits_visual_coverage}), seemingly because these agents carve the grid up into subregions corresponding to their individual preferences, and remain in these subregions since constituents thereof are always likely to appear in the 100-image archive sample during branching.
In \autoref{fig:traits_grids_representative_10}, for example, we see the work of an agent who, according to their personality prompt, is ``searching for the dry red color of terracotta clay'', and has accordingly produced a large set of near-identical solid swatches of such colors.

With a large number of agents ($NA=1,000$), we see the highest Semantic Coverage and Tree Balance of any of our experimental settings.
Further analysis is needed to determine whether these agents aren't similarly carving up the archive into a large number of subregions.
But even so, given that there is low probability that a constituent of any given one of these $1,000$ supposed subregions will appear in the archive sample during branching, these agents would likely be forced to branch from an image from beyond their comfort zone.
These dynamics of artificial collaborative friction in which agents with idiosyncratic personal agendas are forced to work with disagreeable raw materials could provide some hope of recovering the boldness of human leaps of invention; trading self-satisfaction for energizing internal conflict.

And yet, unfortunately, these many-agent grids are rife with high-frequency, uninterpretable (usually grayscale) psychedelic patterns---these distinct forms making up 10-20\% of the archive while virtually absent from other experimental settings.
It's interesting that this noise in particular---versus that of Random, or under large $\epsilon$---attains the highest Semantic Coverage by a wide margin.
Clearly, these images are something like adversarial attacks \citep{nguyen2015deep}.
In future work, a complementary metric measuring semantic variance of a single image over repeated rounds of captioning could help determine whether each such adversarial image consistently maps to a single agent-preference (and textual description) or---perhaps more likely given the collaborative friction described above---these images serve as adversarial ``hubs'', massaged by many agents to at once conform to multiple distinct preferences (and analogously mapping to many distinct textual descriptions).

\section{Conclusion}

Are large Vision-Language Models capable of open-ended discovery? Can they thereby be used to automate processes that hinge on the kind of boundless creativity that has until now been viewed as a uniquely human capacity?
Leveraging Picbreeder as a minimal substrate for the potential expression of such open-ended evolutionary processes, and faithfully placing VLMs in the role of human users, we bring these models' output into direct contact with their historical human counterparts and our own intuitions about what constitutes the special open-ended quality of this output.

We test these intuitions by modeling them computationally.
We find that, taken together, our metrics of Semantic Recall, phylogenetic Tree Balance, and Visual and Semantic Coverage capture a large part of our sense of meaningful qualitative variation among Picbreeder archives.
Accordingly, we find that a number of separate interventions---striking a balance in terms of the injection of exploratory noise into agents' interactions with the system and the amount of history provided each agent, and maximizing the effective number of such agents in terms of behavioral diversity---can lead to the appearance of increased open-ended potential in generated archives.

These experiments also reveal phenomena demanding further investigation, such as the reliable emergence of unexpected idiosyncracies or pathologies arising under certain settings, or the propagation of apparent adversarial images in the highly multi-agent setting, which may present dangers or opportunities in further scaling such systems.
Above all, experimentation under more diverse conditions (i.e., combining the separate insights gleaned from our interventions above) and at larger scales (most of the experimental settings we present here feel worthy of running for longer) are necessary to gain a better sense of the current state of VLMs' true open-ended potential.
Carefully designed user studies---casting humans as judges or (once again) as users of the system in a controlled setting---may also prove instrumental in calibrating evaluation metrics and motivating new interventions.

By scaling the methods developed here, we may begin to provide meaningful answers to the question of VLMs' capacity for human-level open-ended discovery, and develop design principles that will allow us to augment and accelerate open-ended processes---of scientific discovery, of the infinite generation of procedural interactive worlds, of human thought---in good conscience.


\bibliographystyle{plainnat}
\bibliography{refs}

\clearpage
\appendix
\renewcommand{\thefigure}{A\arabic{figure}}
\setcounter{figure}{0}
\renewcommand{\thetable}{A\arabic{table}}
\setcounter{table}{0}

\begin{figure*}
\centering
\includegraphics[width=1.0\linewidth]{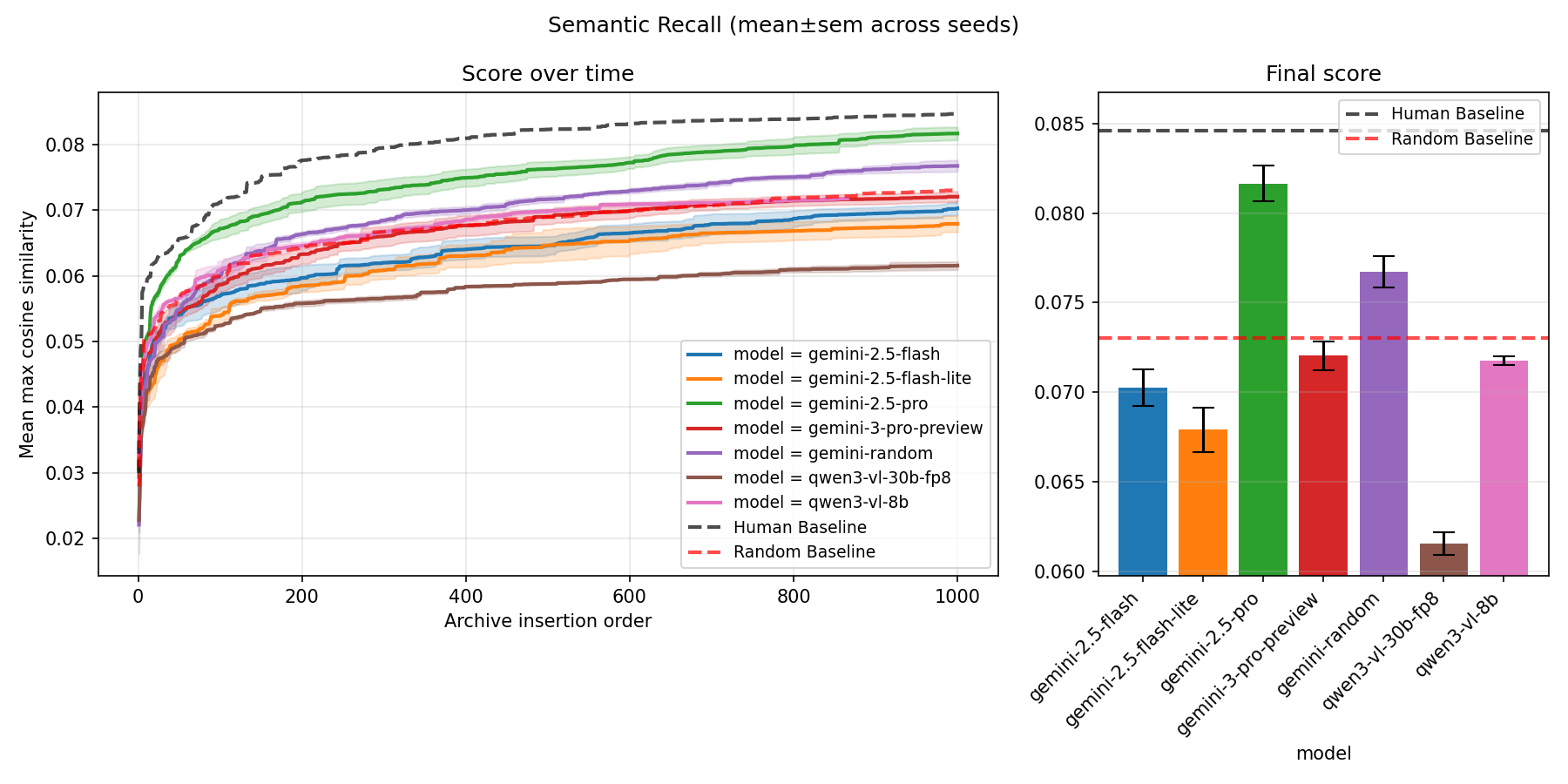}
\caption{Effect of the choice of VLM model on the Semantic Recall of the Picbreeder archive. In the gemini-random setting, each agent is randomly assigned to one of the other gemini models shown in this plot. Surprisingly, gemini-2.5-pro significantly outperforms all other model choices, including gemini-3-pro-preview.}
\label{fig:model_recall}
\end{figure*}

\newlength{\imgvsep}
\setlength{\imgvsep}{0.6em} 
\begin{figure*}
\centering
\begin{subfigure}{.31\linewidth}
\centering
\includegraphics[width=\linewidth]{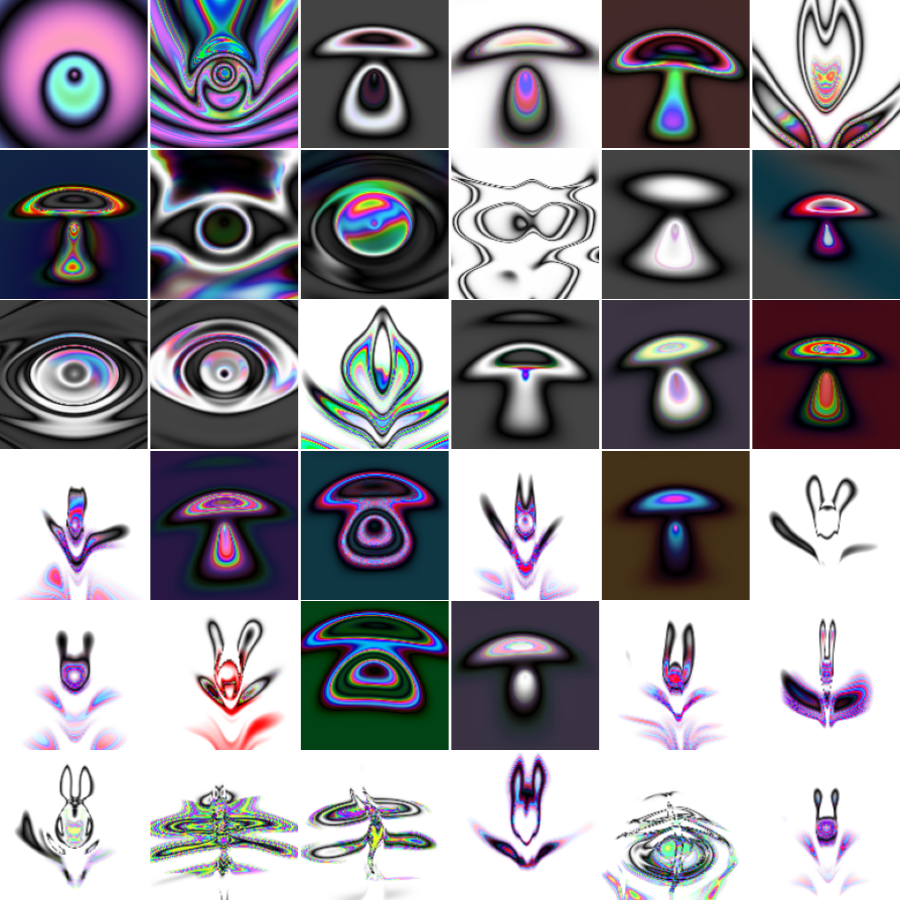}\par\vspace{\imgvsep}
\includegraphics[width=\linewidth]{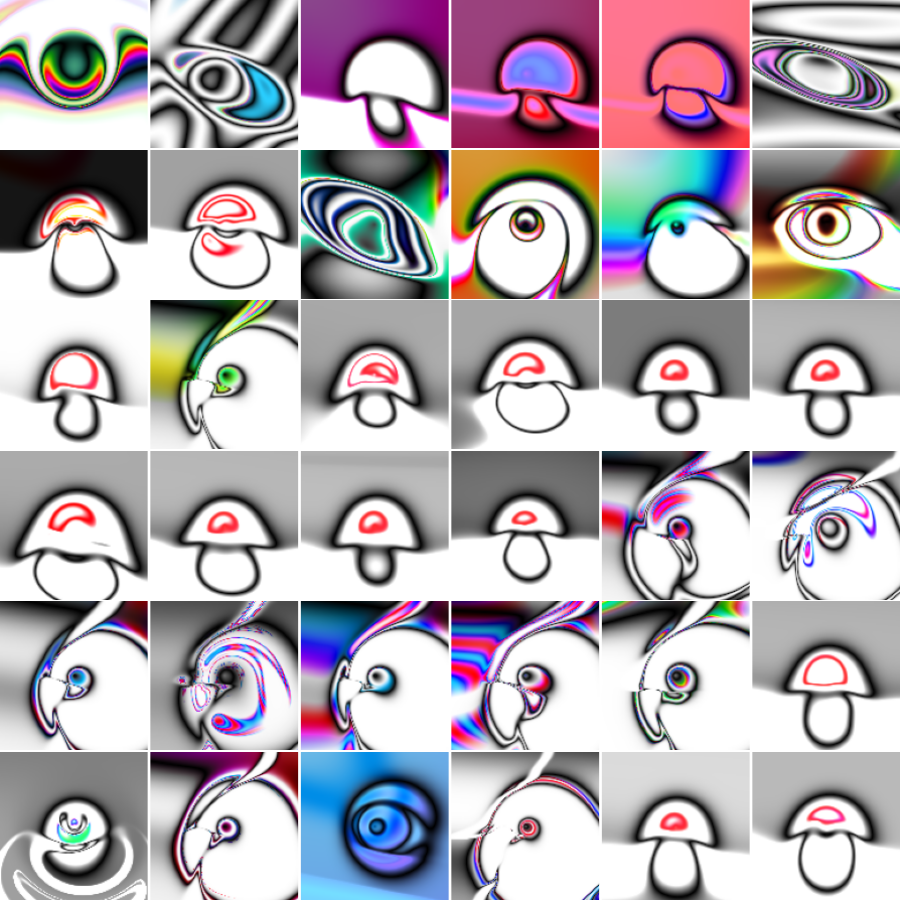}\par\vspace{\imgvsep}
\includegraphics[width=\linewidth]{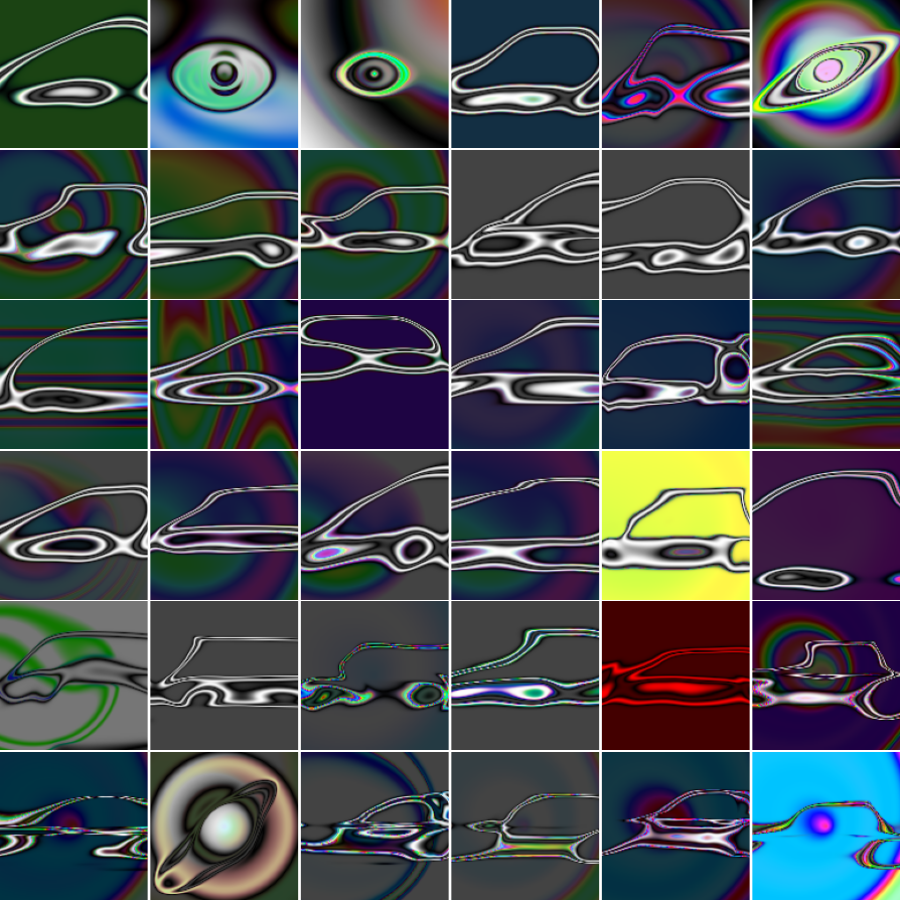}
\caption{gemini-3-pro-preview}
\end{subfigure}
\hfill
\begin{subfigure}{.31\linewidth}
\centering
\includegraphics[width=\linewidth]{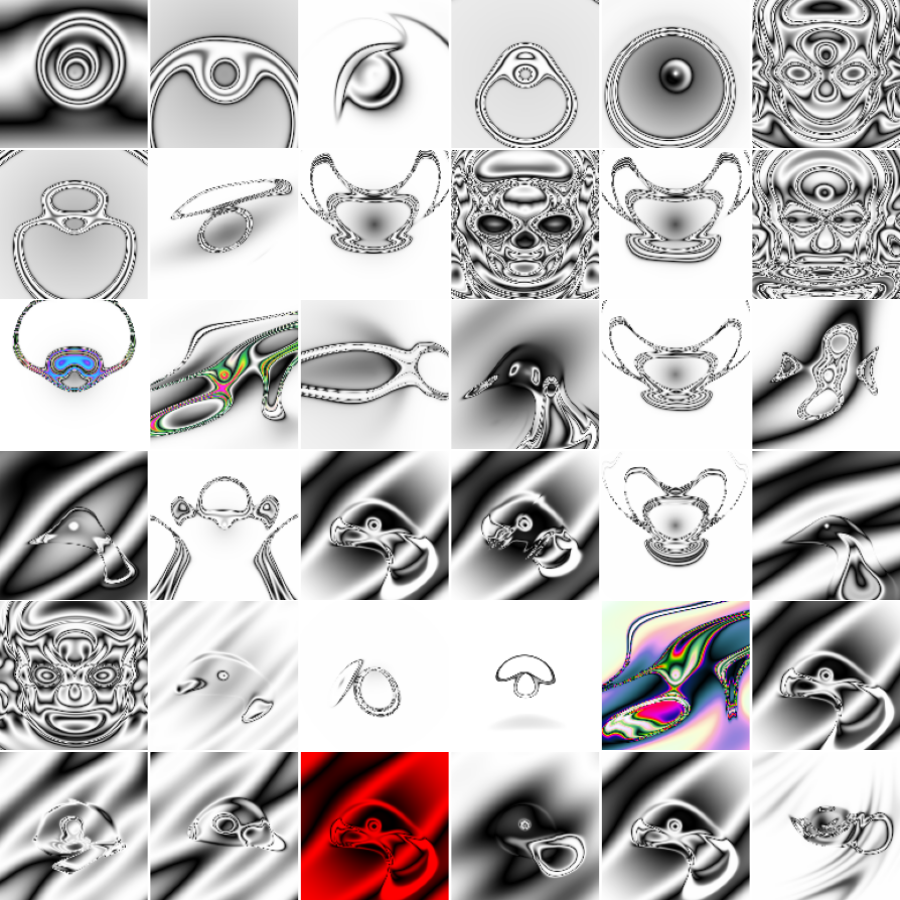}\par\vspace{\imgvsep}
\includegraphics[width=\linewidth]{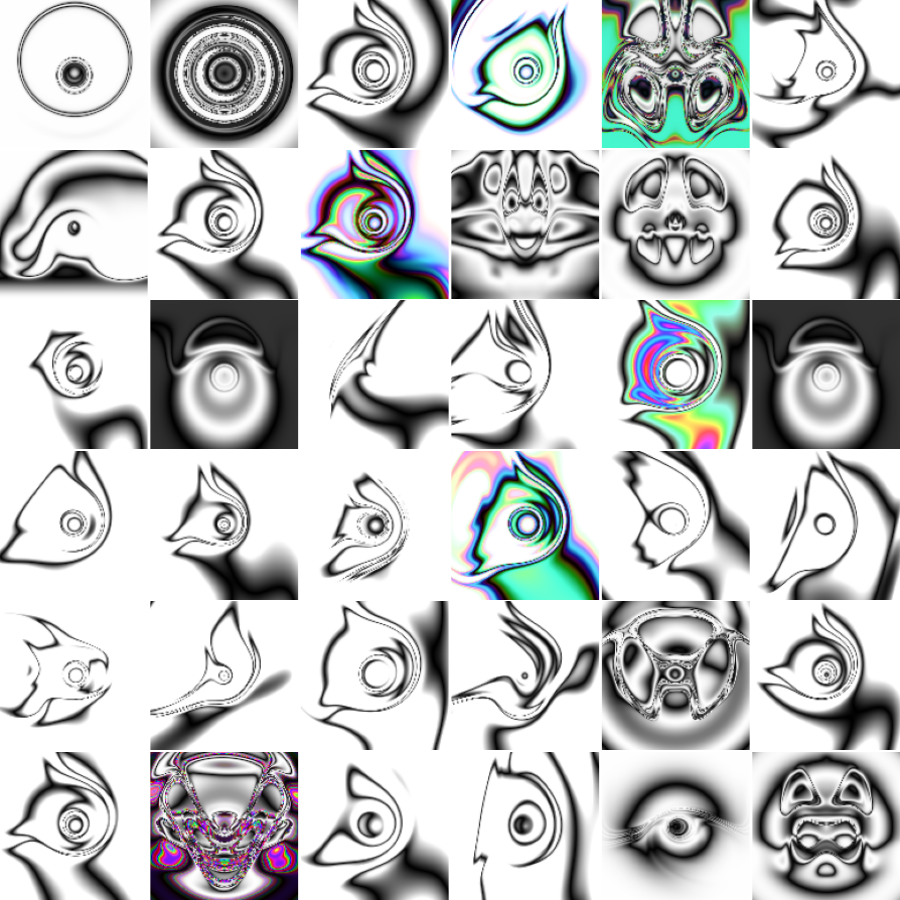}\par\vspace{\imgvsep}
\includegraphics[width=\linewidth]{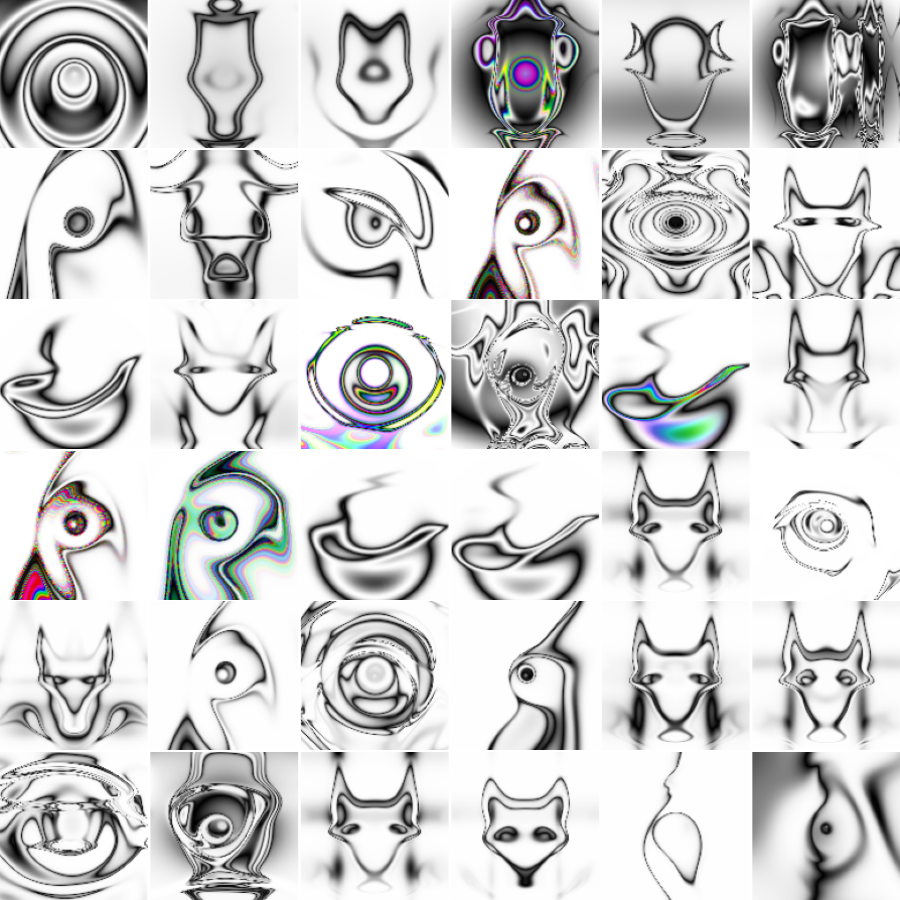}
\caption{gemini-2.5-pro}
\end{subfigure}
\hfill
\begin{subfigure}{.31\linewidth}
\centering
\includegraphics[width=\linewidth]{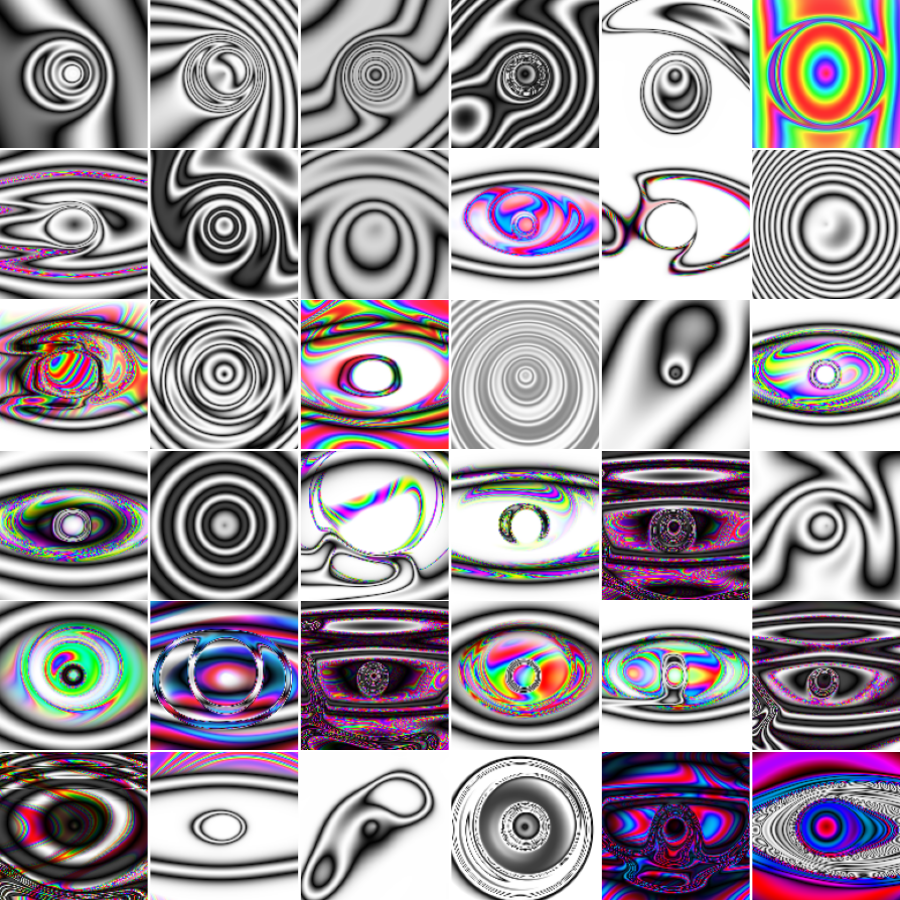}\par\vspace{\imgvsep}
\includegraphics[width=\linewidth]{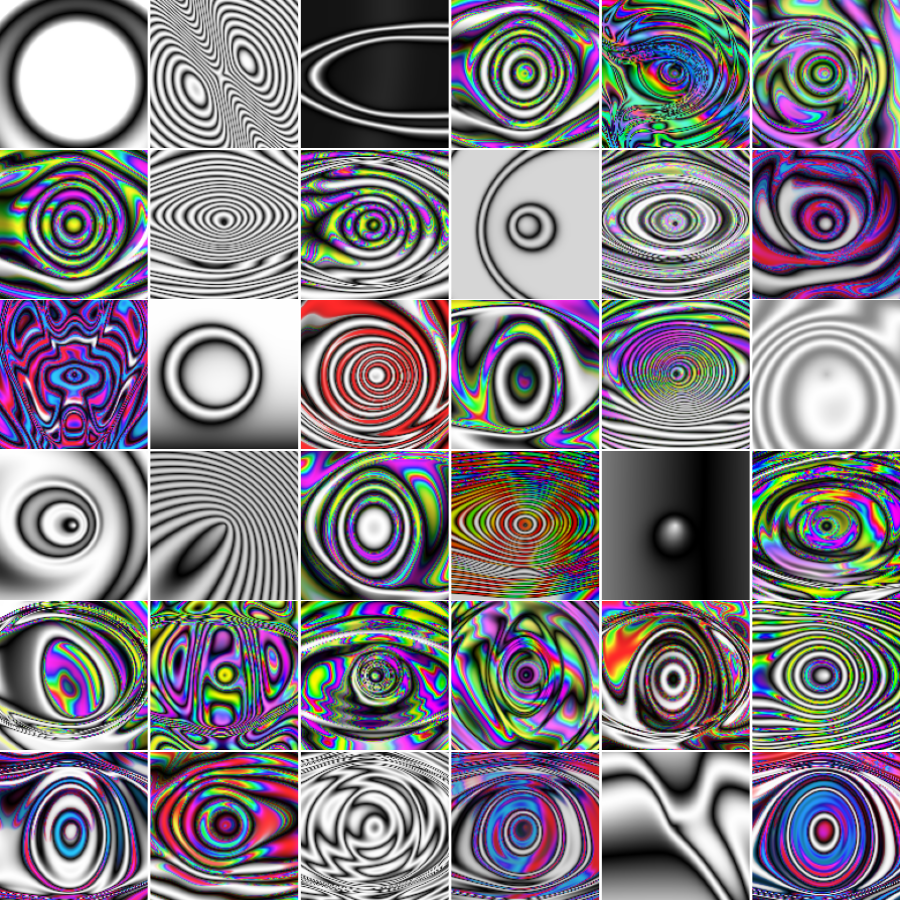}\par\vspace{\imgvsep}
\includegraphics[width=\linewidth]{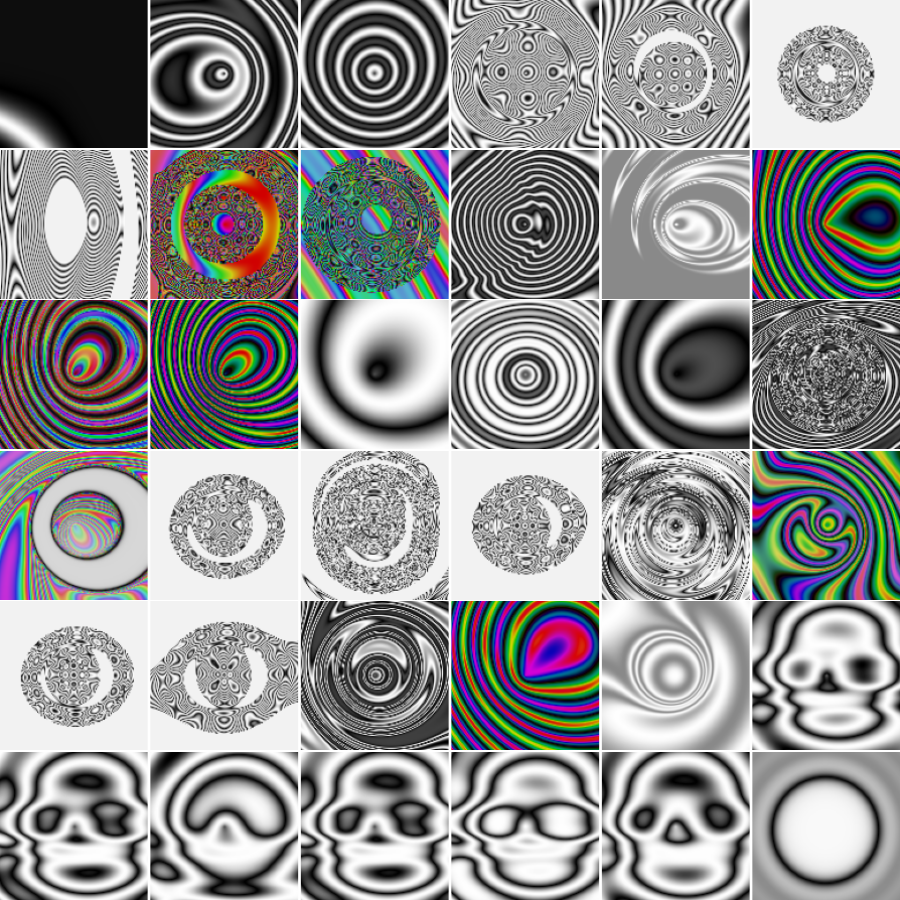}
\caption{gemini-2.5-flash-lite}
\end{subfigure}
\caption{Effect of the choice of VLM model on Picbreeder archives after 500 agent sessions. For each model, samples of 3 archives from different random seeds are shown. Samples are generated by selecting images at uniform intervals with respect to publication order. We note that \texttt{gemini-3-pro-preview} is prone to a kind of mode collapse in the collaborative archive, often obsessing over mushroom-like forms in particular. \texttt{gemini-2.5-flash-lite}, meanwhile, tends to flood the archive with abstract, high-frequency, psychedelic patterns (though it also discovers skull-like forms whose internal representations we evaluate in \autoref{fig:skull_representations} resembling the---admittedly more refined---skull generated in the human baseline.}
\label{fig:model_grids}
\end{figure*}

\section{Limitations \& Future Work}

We could let the agents evolve indefinitely. Let them restart, quit, publish multiple times, or not at all.
This is probably quite important, and likely a bottleneck on the open-ended potential of the current system.
In this work, we constrain the agent along these lines because it makes for cleaner comparisons in terms of our evaluation metrics by guaranteeing that archives of the same size will have resulted from the same number of evolutionary generations.
However, the less constrainted alternative is supported in our codebase (by specifying \verb|fixed_session_length=False|).

During preliminary experiments, we found that agents tended to make redundant publications in quick succession under this setting (even when warned against it), often seeing every minor variation upon an image with which they were already satisfied as something worth sharing with the world.
This was true even when giving agents unlimited context of their current session (though such a failure case would be even more fundamentally difficult to avoid with more limited context), which in itself would grow prohibitively expensive with unconstrainted context length (our code falls back to trimming the oldest chat turns from history when encountering model-specific token limits).

Regarding memory, we might streamline what we put into the agent's context, e.g., always keeping the archive sample present to include diversity, and/or keeping the agent's original (branching) selection (and/or subsequent selections), to allow for exploration without overwhelming the agent's context and splitting its attention to detrimental effect.

We could perhaps fine-tune VLMs on Picbreeder trajectories, giving them something like a long-term memory of past experiences with the system and the running online archive, and potentially differentially imbuing variants of the same VLM with certain behavioral preferences.

\section{API Costs}

A full Picbreeder-VLM run with full in-session memory---producing a archive with $\approx$10k human images to match our human dataset---costs around 10k USD.
To date, we've spent around 150k USD on gemini API calls while developing and using the system.

\section{Additional Results}

\begin{figure*}
\centering
\includegraphics[width=\linewidth]{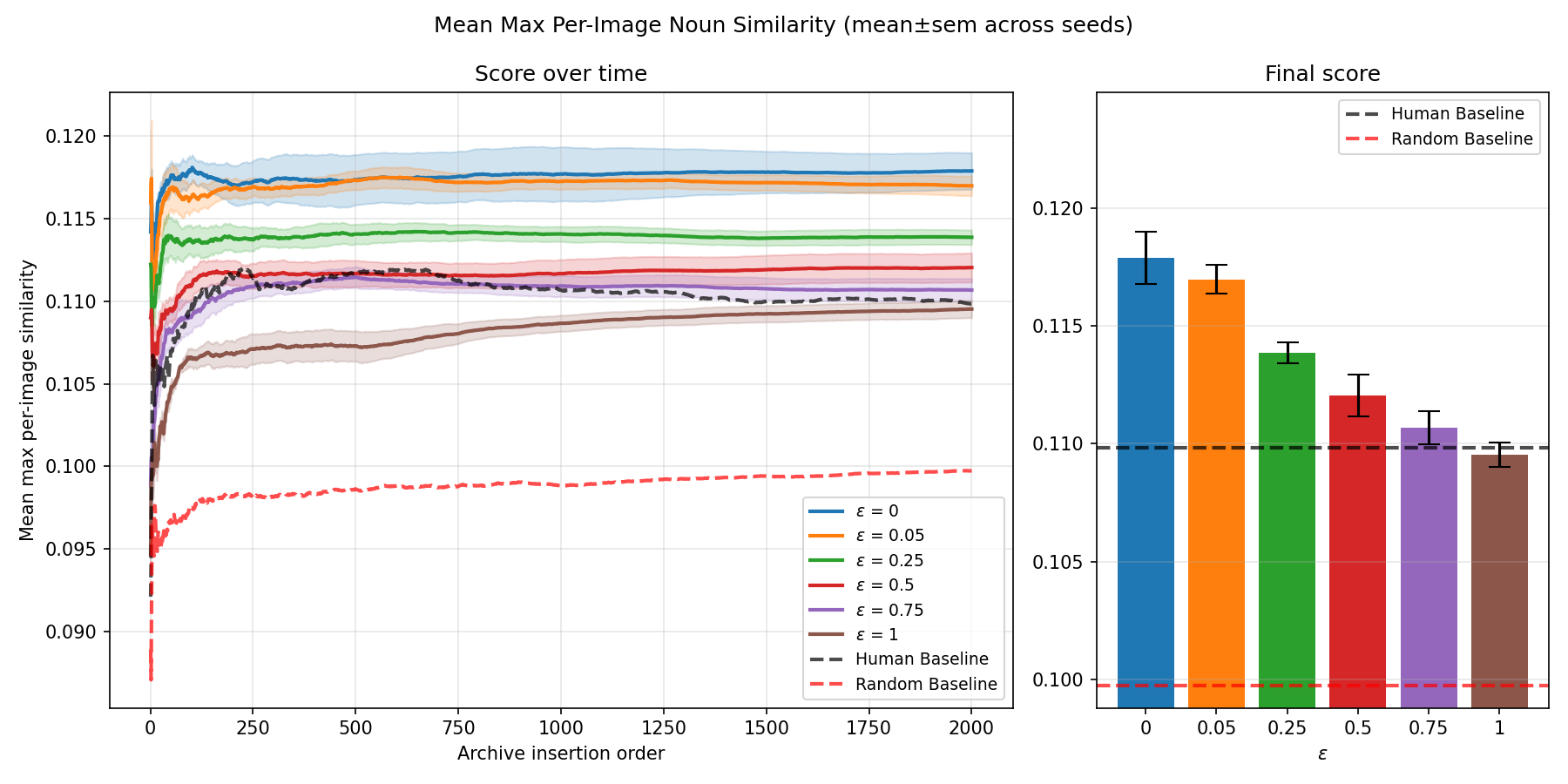}
\caption{Effect of exploration ($\epsilon$-greedy) on the Fidelity of the Picbreeder archive. Greedy strategies can game this metric by refining a small set of images and flooding the archive with near-duplicates.}
\label{fig:exploration_fidelity}
\end{figure*}
\begin{figure*}
\centering
\includegraphics[width=\linewidth]{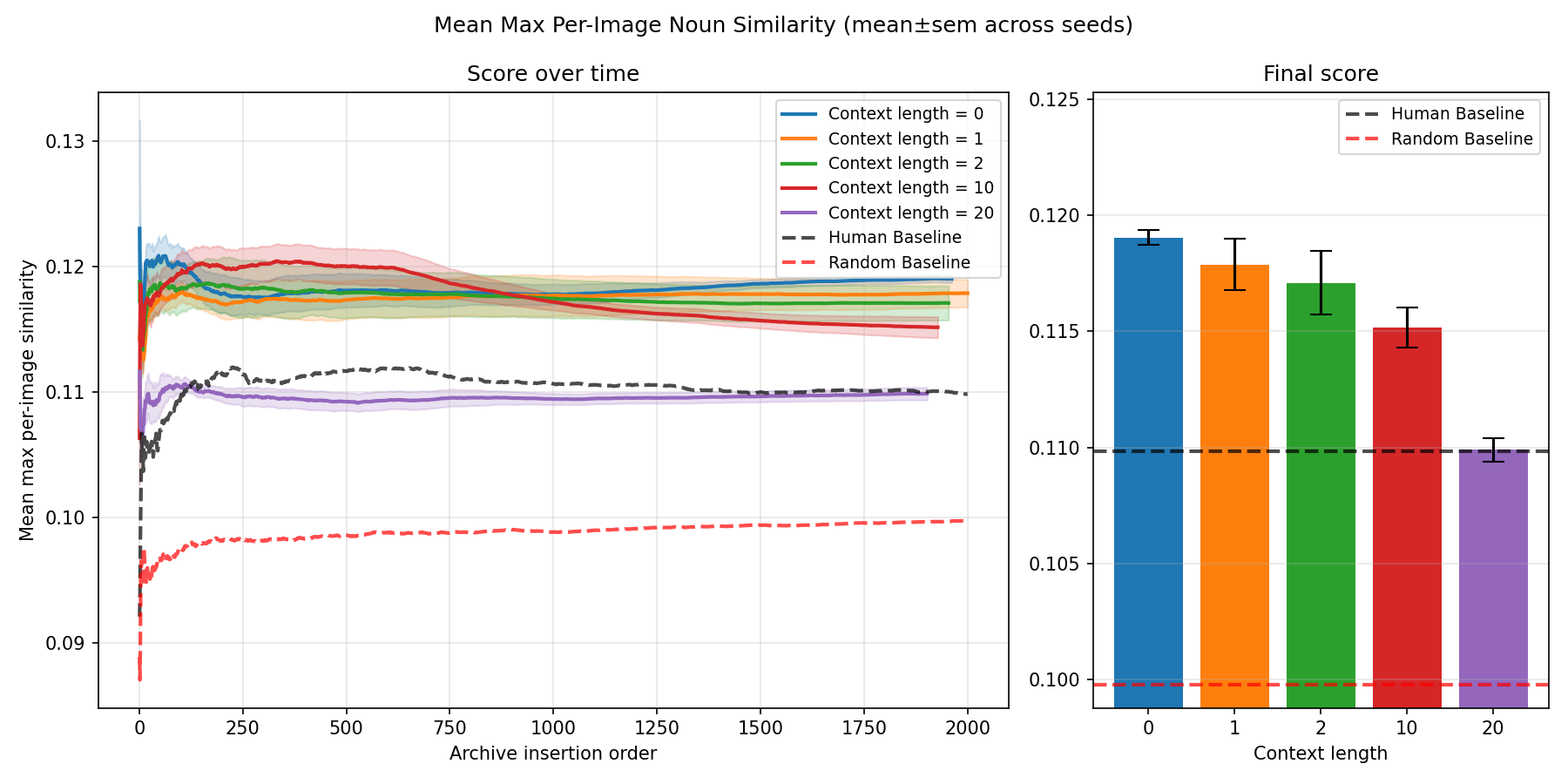}
\caption{Effect of history (number of previous actions included in an agent’s context/memory) on the Fidelity of the Picbreeder archive. Agents without history perform best by refining the current form, without any incentive to push past local optima.}
\label{fig:history_fidelity}
\end{figure*}
\begin{figure*}
\centering
\includegraphics[width=\linewidth]{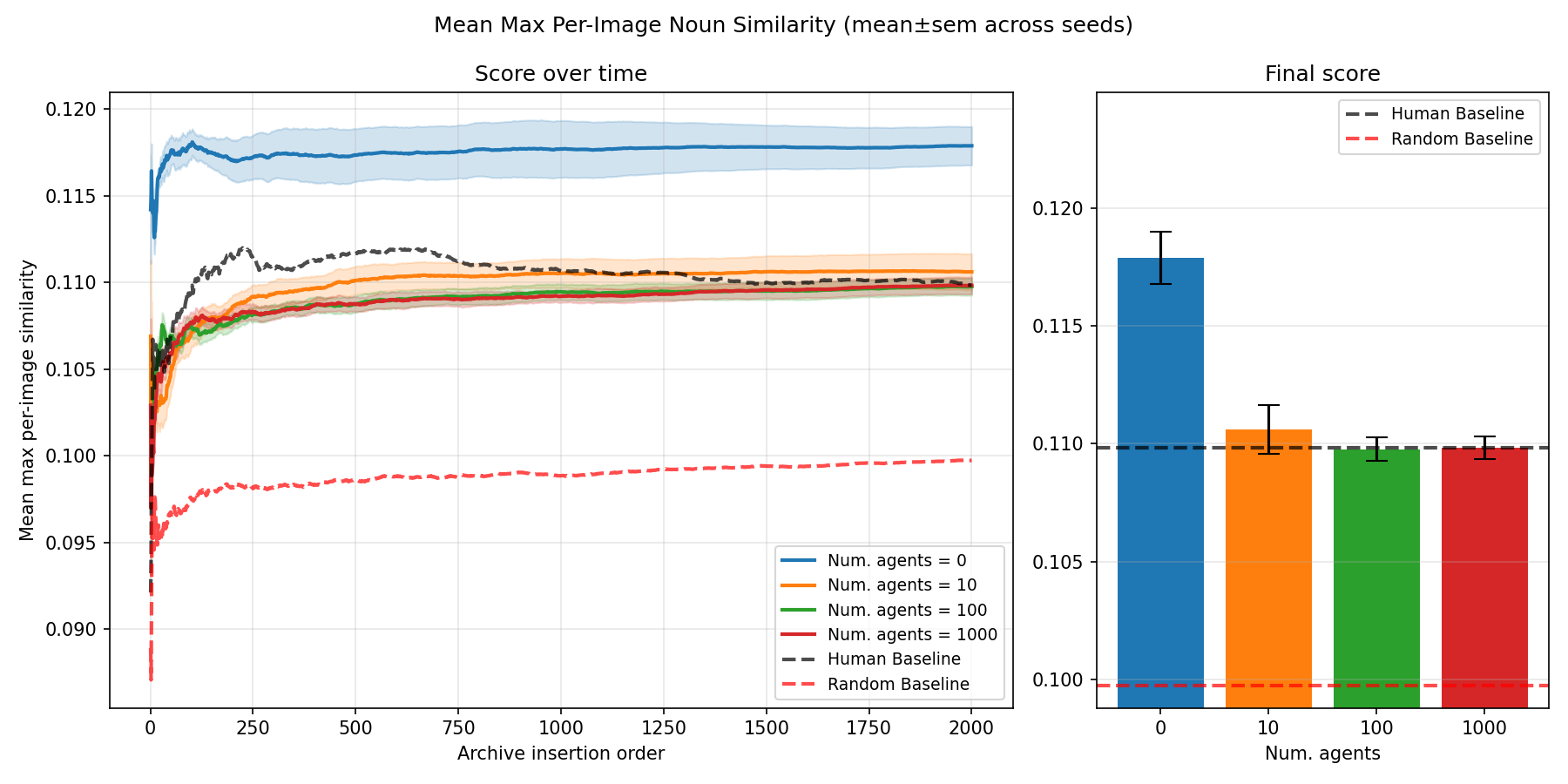}
\caption{Effect of multiple agents (number of distinct personality traits assigned) on the fidelity of the Picbreeder archive. Adding agents reduces Fidelity by inhibiting mode collapse.}
\label{fig:traits_fidelity}
\end{figure*}

Complementary to Semantic Recall, we implement a Semantic Fidelity metric, which takes the average of the best similarity in text-image embedding space between each \textit{image} and any noun (as opposed to between each \textit{noun} and any image).
Results are shown in Figs. \ref{fig:exploration_fidelity}-\ref{fig:traits_fidelity}.
However, this metric can easily be gamed by endlessly reproducing a single semantically salient image.

\subsection{Internal CPPN Representations}

\citet{kumar2025questioning} argue that optimizing toward fixed objectives via Stochastic Gradient Descent (SGD) leads to models with ``fractured, entangled'' representations.
With the human Picbreeder experiment as their counterexample, they argue that, by contrast, open-ended search results in models with ``unified'' representations more aligned with human intuition.
The models in question here are CPPNs, but their argument extends to deep neural networks in general; they point to similar ``entanglement'' among GPT-3's representations, e.g. its inability to count farm animals, in contrast to its ability to count office supplies.

In this light, the present work can be seen as asking whether we can hope to automate open-ended search over neural network substrates---sans any humans-in-the-loop---with the same beneficial outcome for the internal representations of these models.
If we can automate such a search, we may by extension be able to automate AI research itself, using VLMs to guide the training of new generations of VLMs.
In such an open-ended AI-generating system, we could imagine for example that these next-generation VLMs would be trained by their forebears with a series of diverse objectives, or on a curriculum of diverse corpora, as opposed to over a monolithic corpus with a single autoregressive objective.

\begin{figure*}
\centering
\includegraphics[width=0.5\linewidth]{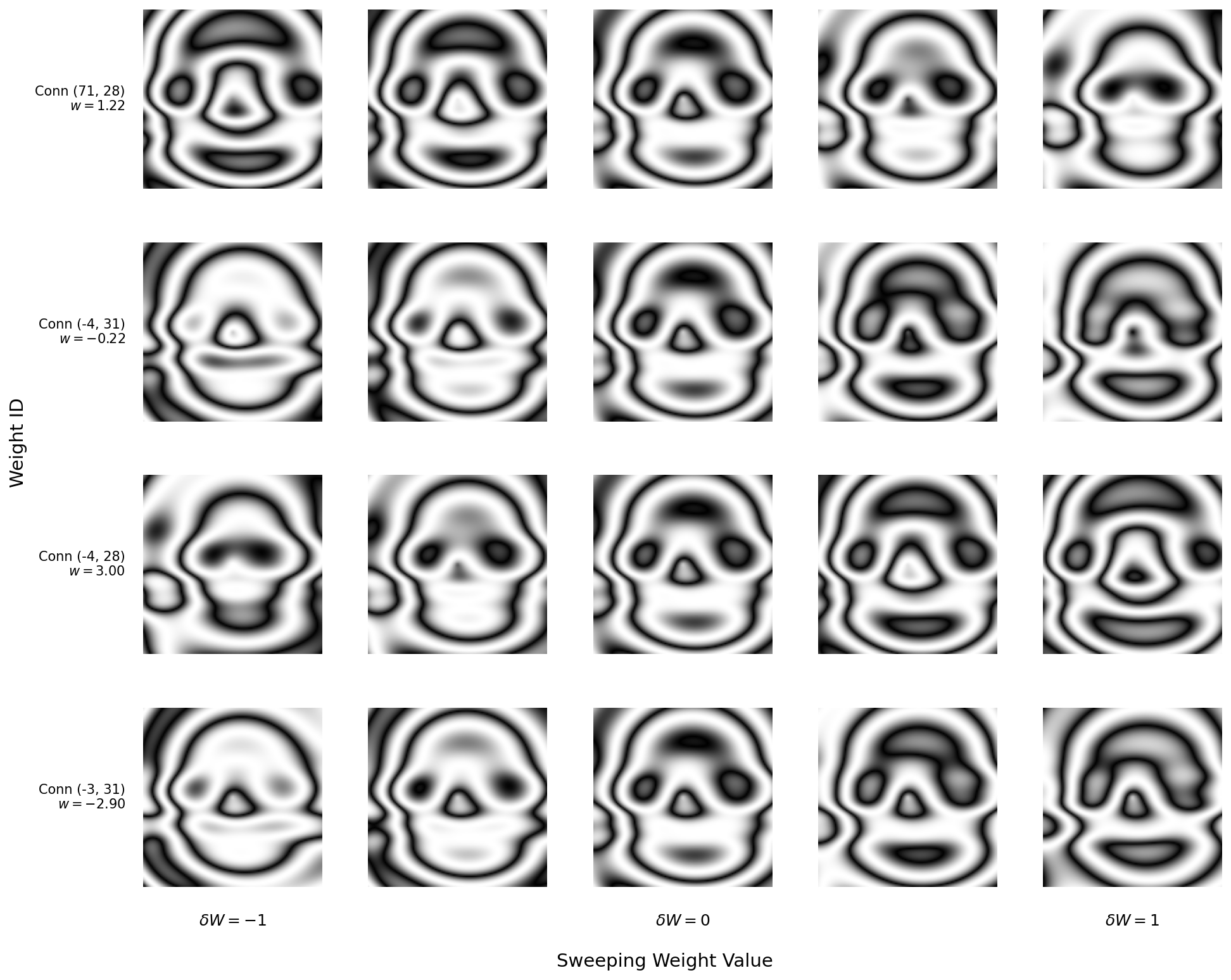}
\caption{Visualization of internal representations resulting from VLM-driven evolution. We apply perturbations from $[-1,1]$ to each weight in the CPPN, and display the weights that lead to greatest difference in terms of pixel distance from the initial image at the extremes of this range. These representations are not nearly as ``fractured'' as those resulting from SGD over a fixed-topology CPPN in \cite{kumar2025questioning}, but neither are they so neatly factorized as to correspond to features like mouths or eyes opening or closing.}
\label{fig:skull_representations}
\end{figure*}

With this question in mind, we replicate the analysis of internal CPPN representations in \citep{kumar2025questioning} on a CPPN representation of a skull resulting from VLM-driven Picbreeder in \autoref{fig:skull_representations}.
Using a CPPN that generates an image of a skull resembling that from the human experiment (though considerably less refined), we sweep the values of each individual weight of the network, adding values in $[-1,1]$ to the weight's original value.
We visualize the weight-sweeps that lead to greatest difference from the original image in terms of pixel distance at the extremes of the weight's values.
We find that weights lead to relatively smooth changes to the image, and that the perturbed images still mostly resemble skulls, whereas perturbations applied to the weights of the SGD-generated CPPN in \citep{kumar2025questioning} led to more chaotic and destructive changes to the image.
However, we don't quite find any of the clean semantic labels---like ``mouth opening'' or ``eye winking''---recovered from the human-generated CPPN in \citep{kumar2025questioning}.

Overall, this would seem to be a promising result, which suggests that by further refining the strategies introduced here for VLM-guided open-ended search, we could produce models with increasingly unified representations.
It's worth noting some potentially confounding factor though, namely the use of NEAT-style evolution compared to SGD over a fixed-topology network, and the initial difference between the skulls here and in \citep{kumar2025questioning}.

\begin{figure*}
\includegraphics[width=.4\linewidth]{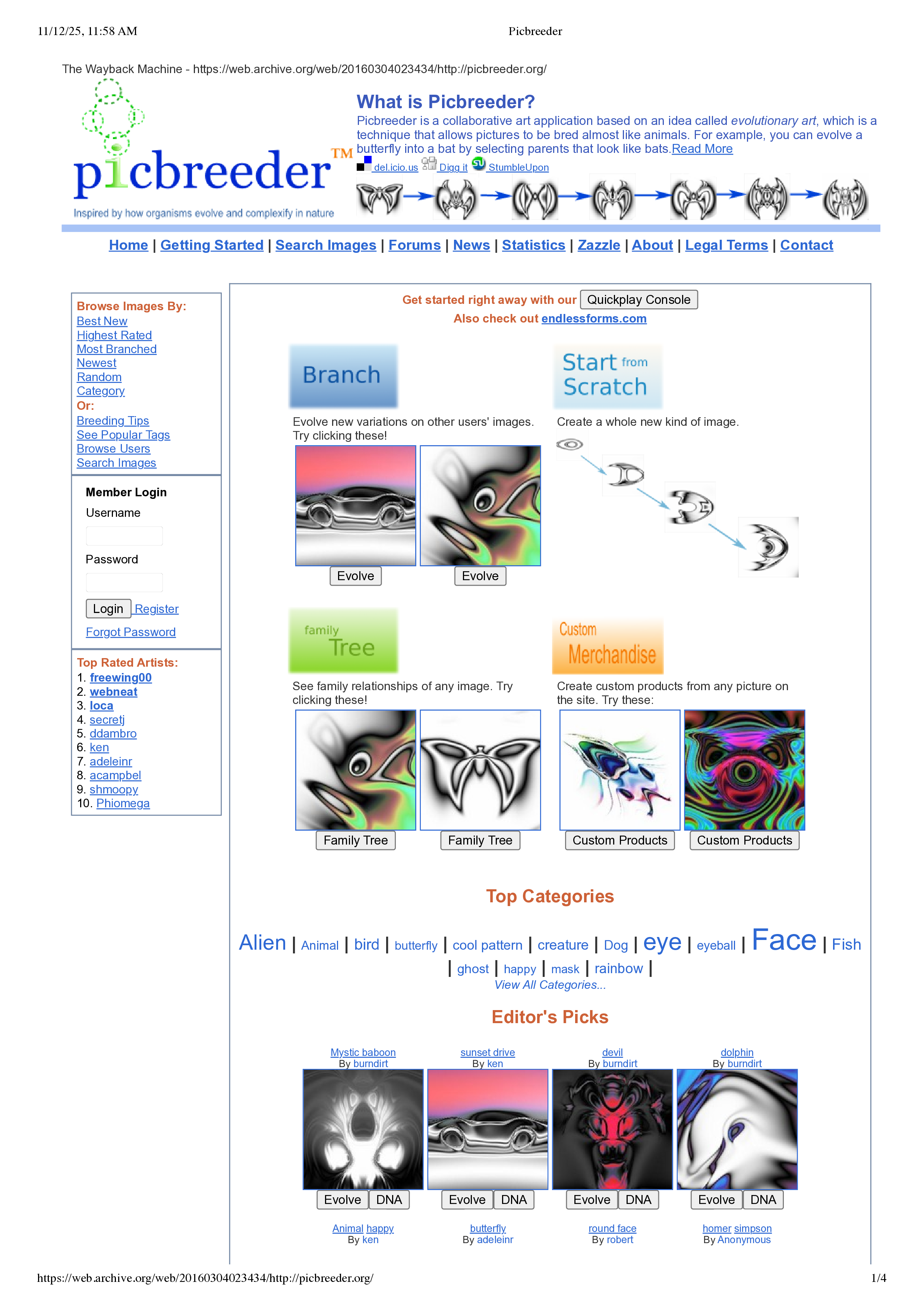}
\includegraphics[width=.4\linewidth]{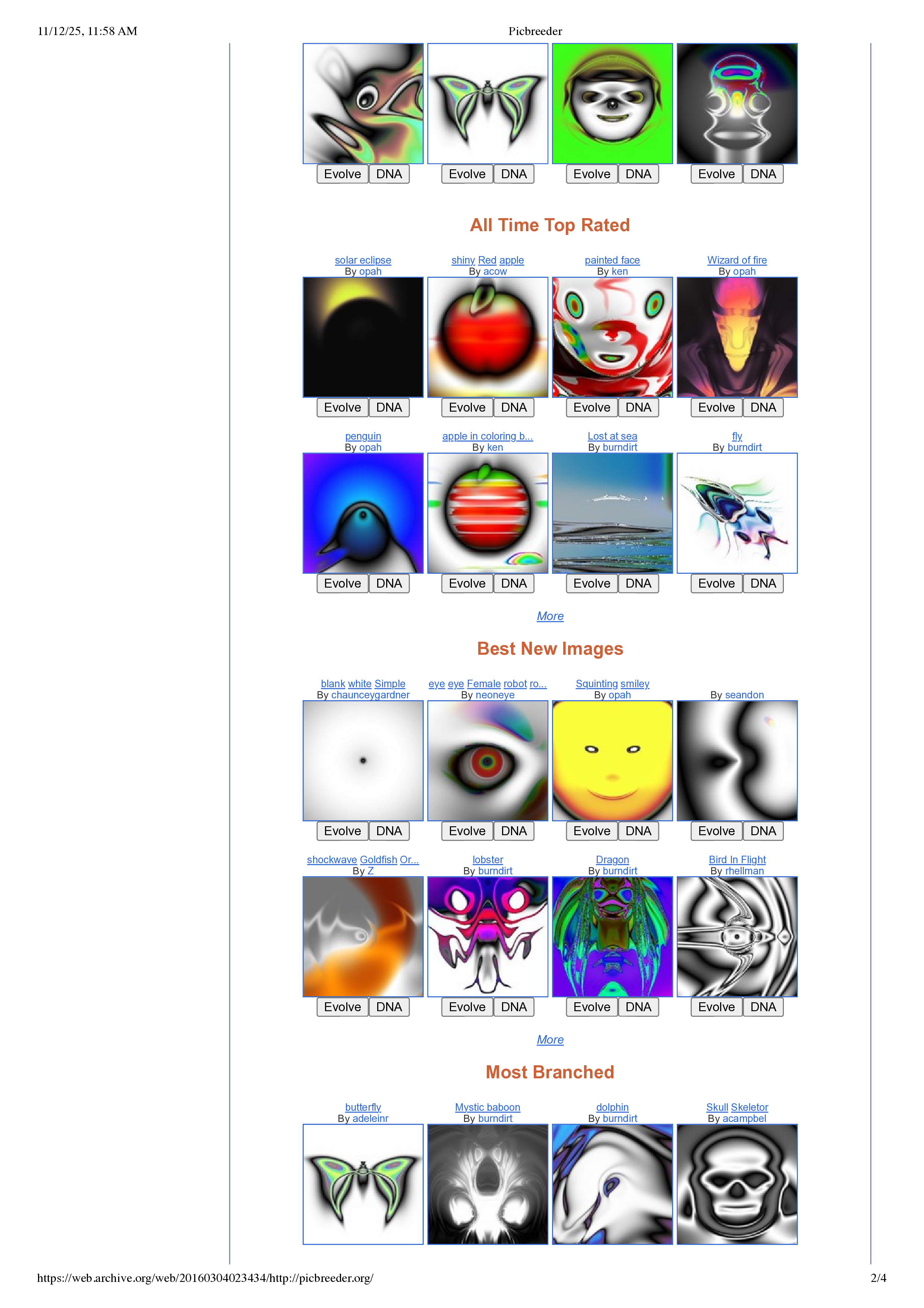}
\includegraphics[width=.4\linewidth]{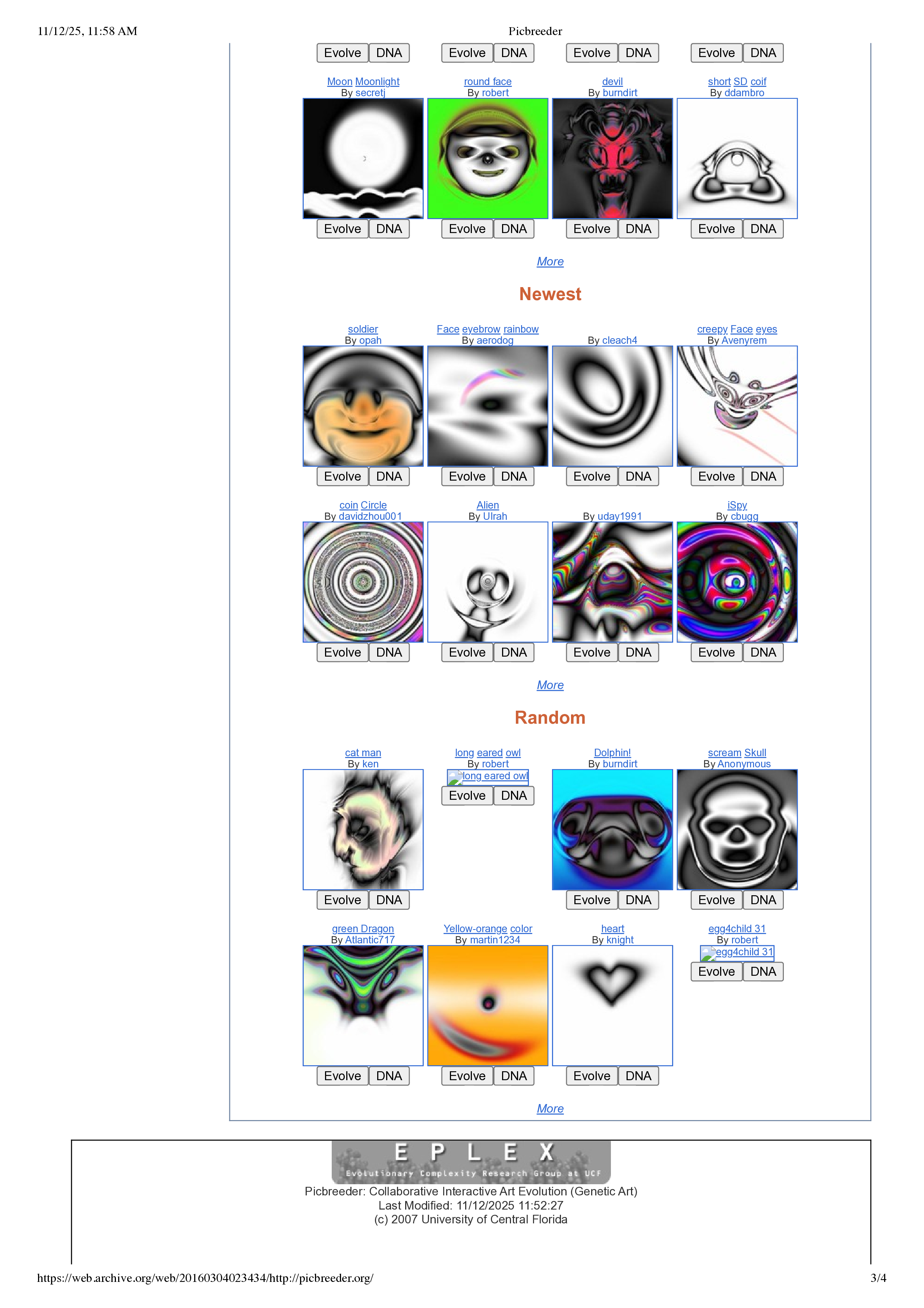}
\caption{Snapshots of the original Picbreeder webpage, recovered via the Wayback Machine. We mimic a human user's exposure to this home page by displaying, at the beginning of each VLM agent's Picbreeder session, a sample of the archive generated thus far comprising top rated, best new, most branched, and random subsamples. Absent from our re-implementation are semantic tags, ``Editor's Picks'', user information, image titles, and the ability to further browse the site/archive.}
\label{fig:picbreeder_homepage}
\end{figure*}

\begin{figure*}[h]
\includegraphics[width=0.8\linewidth]{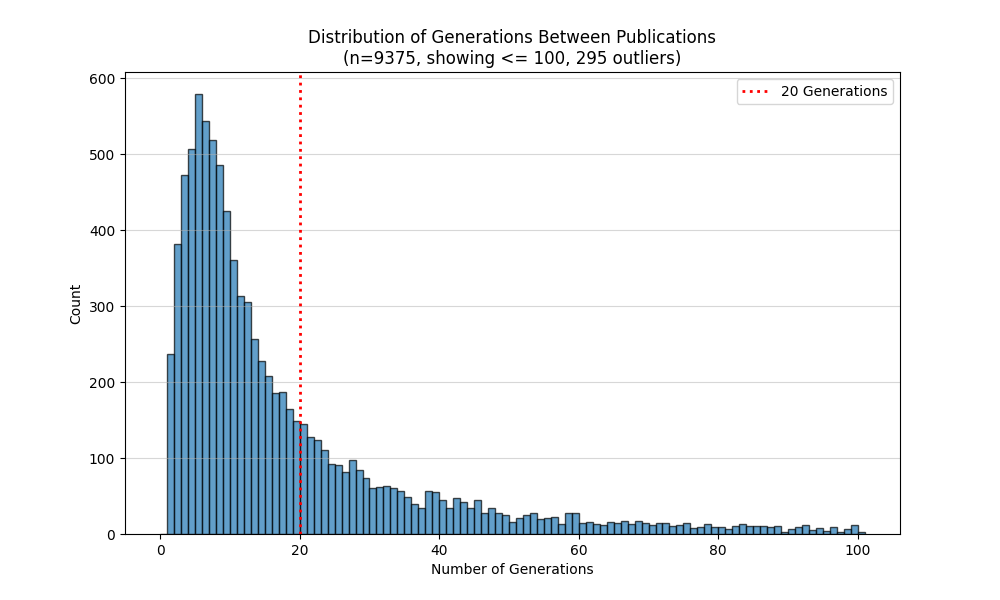} 
\caption{Distribution of human Picbreeder session lengths.}
\label{fig:session_length_dist}
\end{figure*}

\begin{table*}[h]
\begin{tabular}{lc}
    \toprule
    \textbf{Statistic} & \textbf{Value} \\
    \midrule
    Mean Generations & 21.96 \\
    Median Generations & 11.00 \\
    Min Generations & 1 \\
    Max Generations & 765 \\
    \% $\le$ 20 Generations & 70.9\% \\
    \bottomrule
\end{tabular}
\label{tab:human_gen_stats}
 
\caption{Human Picbreeder session length statistics.}
\label{tab:session_length_stats}
\end{table*}

\begin{table*}
\begin{tabular}{rp{12cm}}
\toprule
\textbf{\#} & \textbf{Trait Description} \\
\midrule
1 & You prefer images that evoke the smell of campfire smoke. \\
2 & You act like a joyful dancer moving to an unseen rhythm. \\
3 & You are searching for the visual equivalent of a wailing siren. \\
4 & You are drawn to patterns that look like intersecting searchlights. \\
5 & You prefer images that look like magnetic tape tracking errors. \\
6 & You are searching for a visual representation of 'indifference'. \\
7 & You act like a bored teenager trying to break the system boundaries. \\
8 & You prefer images that look like swirling whirlpools. \\
9 & You are searching for a shape that looks like a question mark. \\
10 & You are searching for a visual representation of 'jealousy'. \\
11 & You are searching for the specific yellow of a lemon rind. \\
12 & You are drawn to the texture of overlapping bird feathers. \\
13 & You act like a plumber searching for leaks in the visual pipes. \\
14 & You prefer images that look like they are viewed through a kaleidoscope. \\
15 & You act like an astronomer discovering a new comet. \\
16 & You act like a weary soldier marching towards a destination. \\
17 & You act like a puppeteer trying to make the static images dance. \\
18 & You act like a skeptical critic who thinks everything is 'derivative'. \\
19 & You avoid images that look like tunnels or deep pits. \\
20 & You prefer images that look like analog video feedback loops. \\
21 & You are searching for a shape that resembles a kite. \\
22 & You are searching for a visual representation of 'guilt'. \\
23 & You are drawn to the visual texture of fur or hair. \\
24 & You dislike images that feel 'heavy' and prefer those that look weightless. \\
25 & You are searching for a visual representation of the concept of 'echo'. \\
26 & You have a short attention span and frequently restart from scratch. \\
27 & You are searching for a specific curve that feels 'friendly'. \\
28 & You act like a child collecting shiny pebbles on a beach. \\
29 & You act like a sophisticated art collector bidding at an auction. \\
30 & You act like a coal miner digging in the dark for a gem. \\
31 & You are deeply drawn to imagery resembling nebulae and galaxies. \\
32 & You are drawn to the aesthetic of bad analog TV reception. \\
33 & You act like a detective revisiting a cold case. \\
34 & You prefer images that look like swallowing black holes. \\
35 & You act like a terrified mouse trying to hide in the darker parts of the image. \\
36 & You prefer images with a distinct central void or empty space. \\
37 & You are searching for the visual equivalent of the smell of cinnamon. \\
38 & You have a deep love for fractals and self-simulating patterns. \\
39 & You act like a meticulous watchmaker examining gears and springs. \\
40 & You prefer images that look like glow-in-the-dark stars on a ceiling. \\
\bottomrule
\end{tabular}

\caption{Sample of LLM-generated personality traits used when $NA > 0$. To generate these traits, we give gemini-3-pro-preview the Picbreeder VLM system prompt (\autoref{fig:prompt}) and ask for personality traits that may implicitly affect an agent's behavior on this task (\autoref{fig:traits_prompt}).}
\label{tab:traits}
\end{table*}

\begin{figure*}
\centering
\begin{subfigure}{\linewidth}
\includegraphics[width=1.0\linewidth]{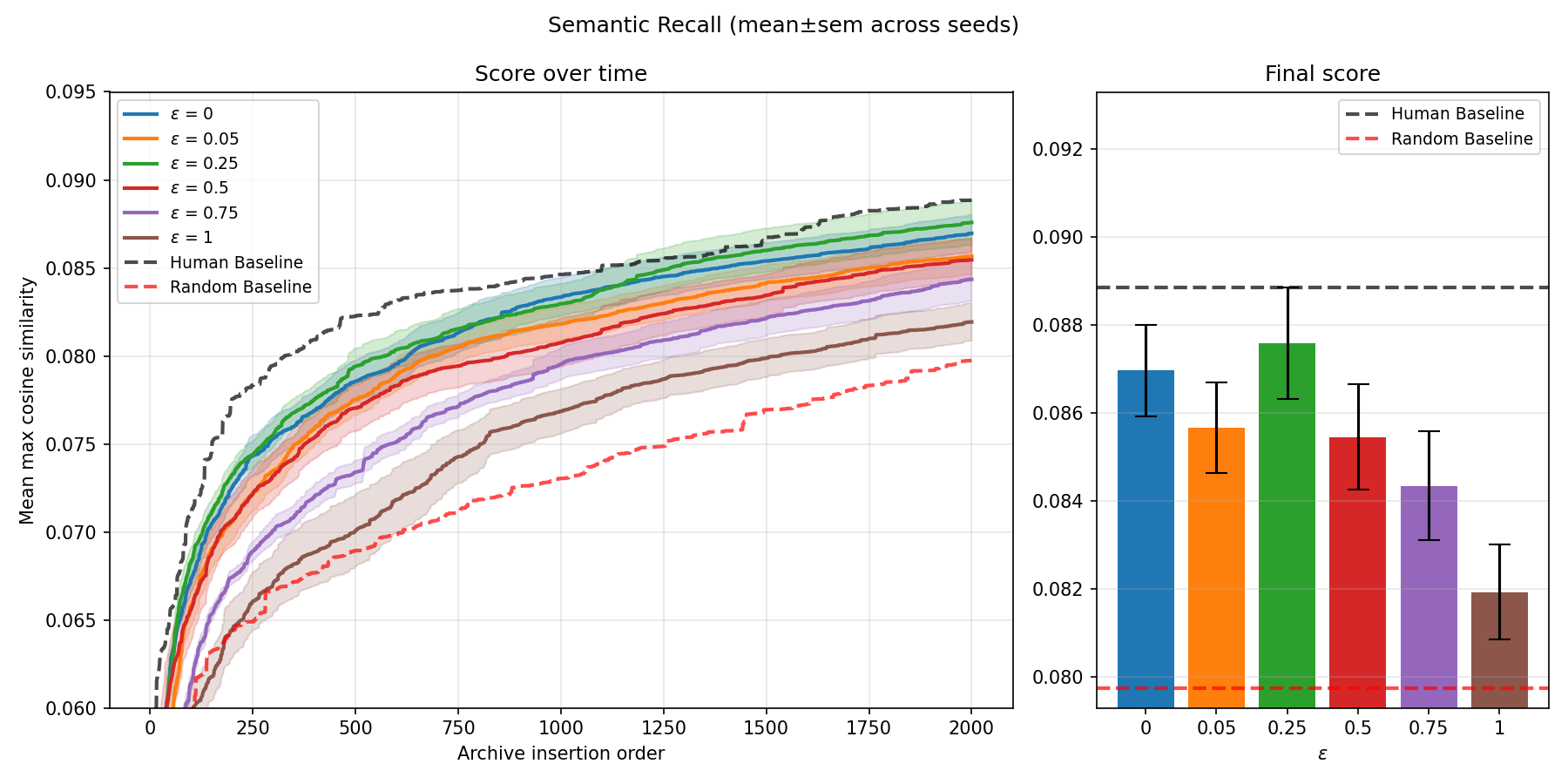}
\caption{Semantic Recall within the Picbreeder archive over the course of collaborative evolution.}
\label{fig:exploration_semantic_recall}
\end{subfigure}
\begin{subfigure}{\linewidth}
\centering
\begin{subfigure}{.31\linewidth}
\includegraphics[width=\linewidth]{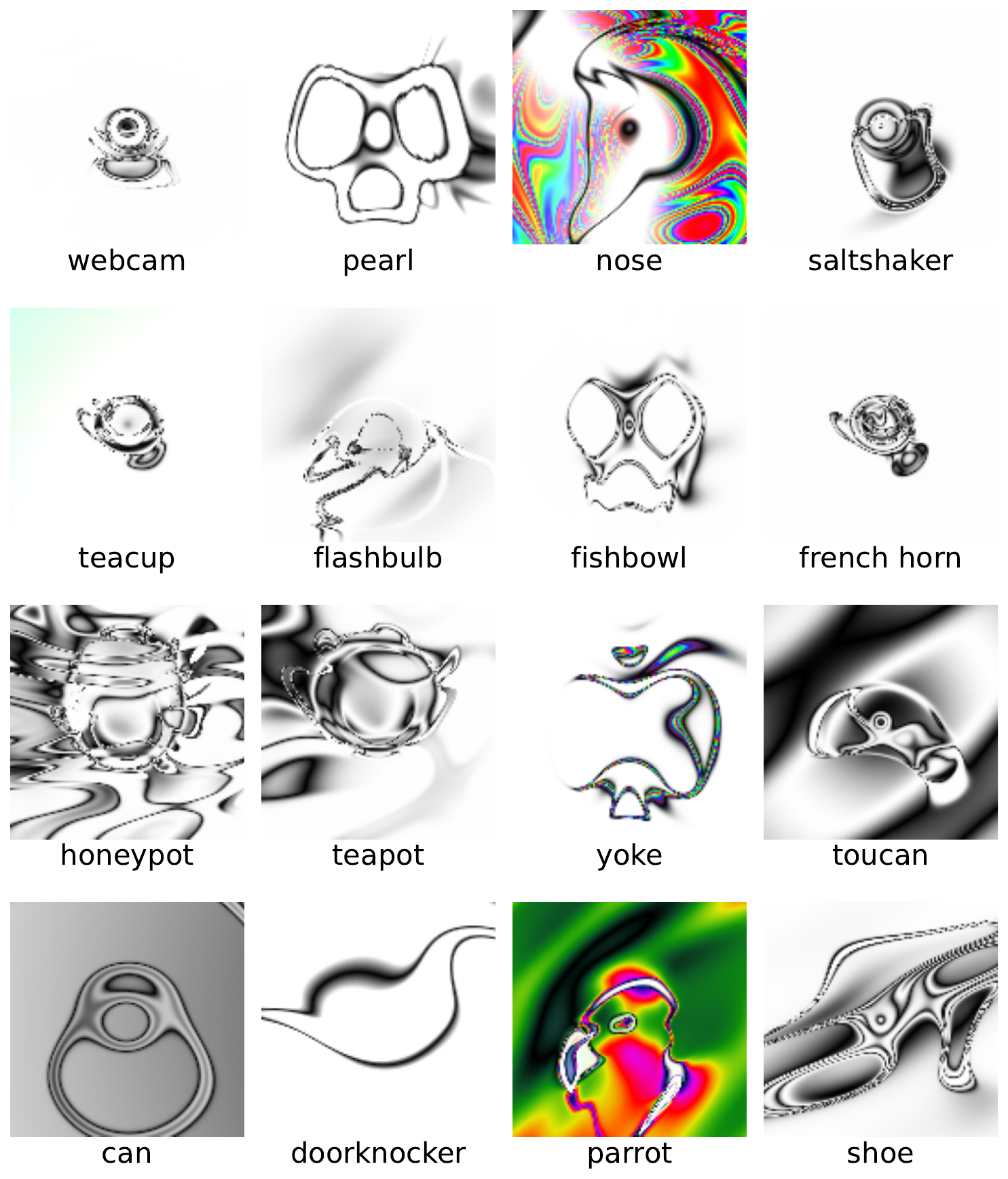}
\caption{$\epsilon$ = 0}
\end{subfigure}
\hfill
\begin{subfigure}{.31\linewidth}
\centering
\includegraphics[width=\linewidth,fbox={1pt 0pt}]{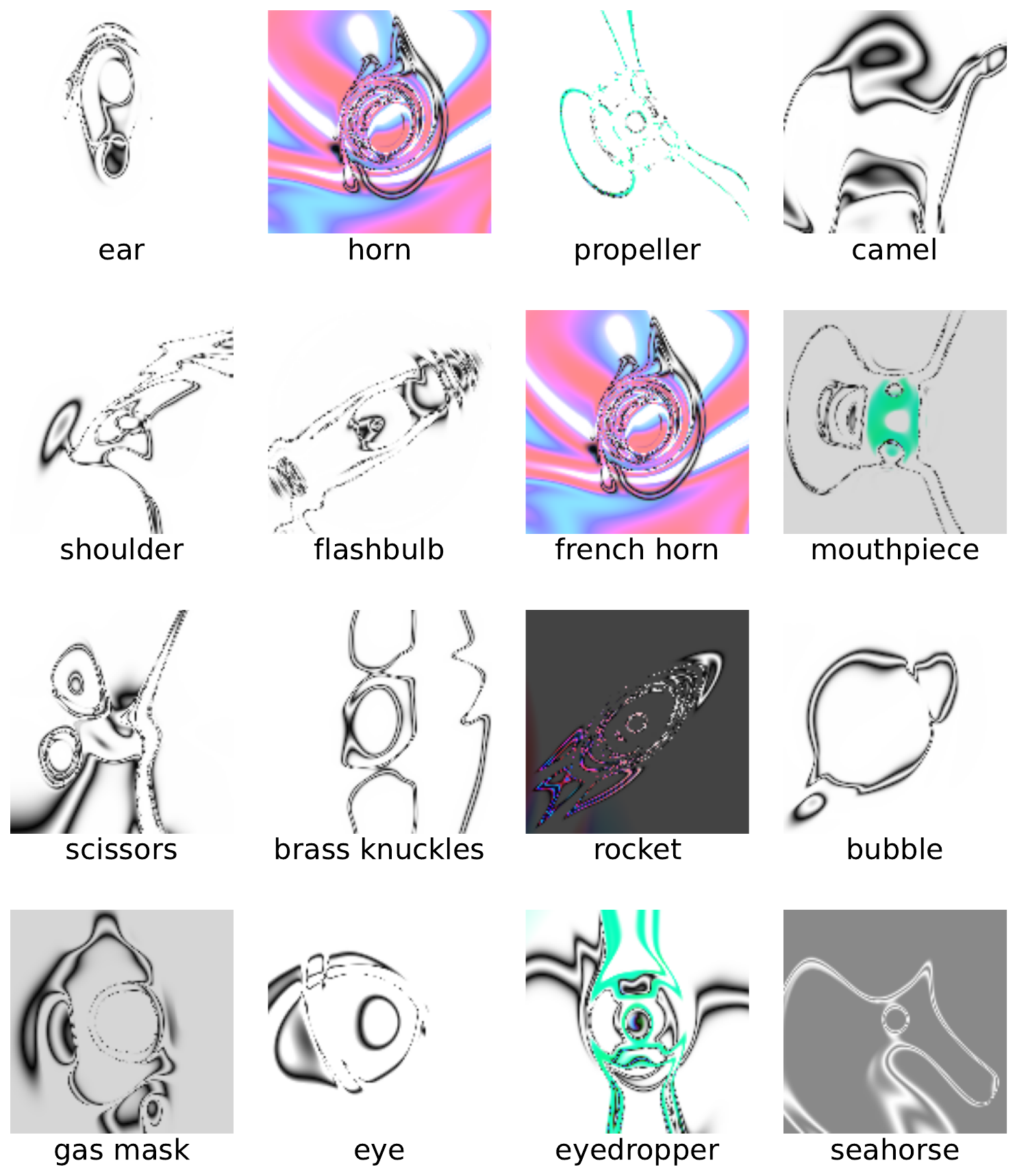}
\caption{$\epsilon$ = 0.25}
\label{fig:exploration_grids_nouniest_epsilon_0_25}
\end{subfigure}
\hfill
\begin{subfigure}{.31\linewidth}
\centering
\includegraphics[width=\linewidth]{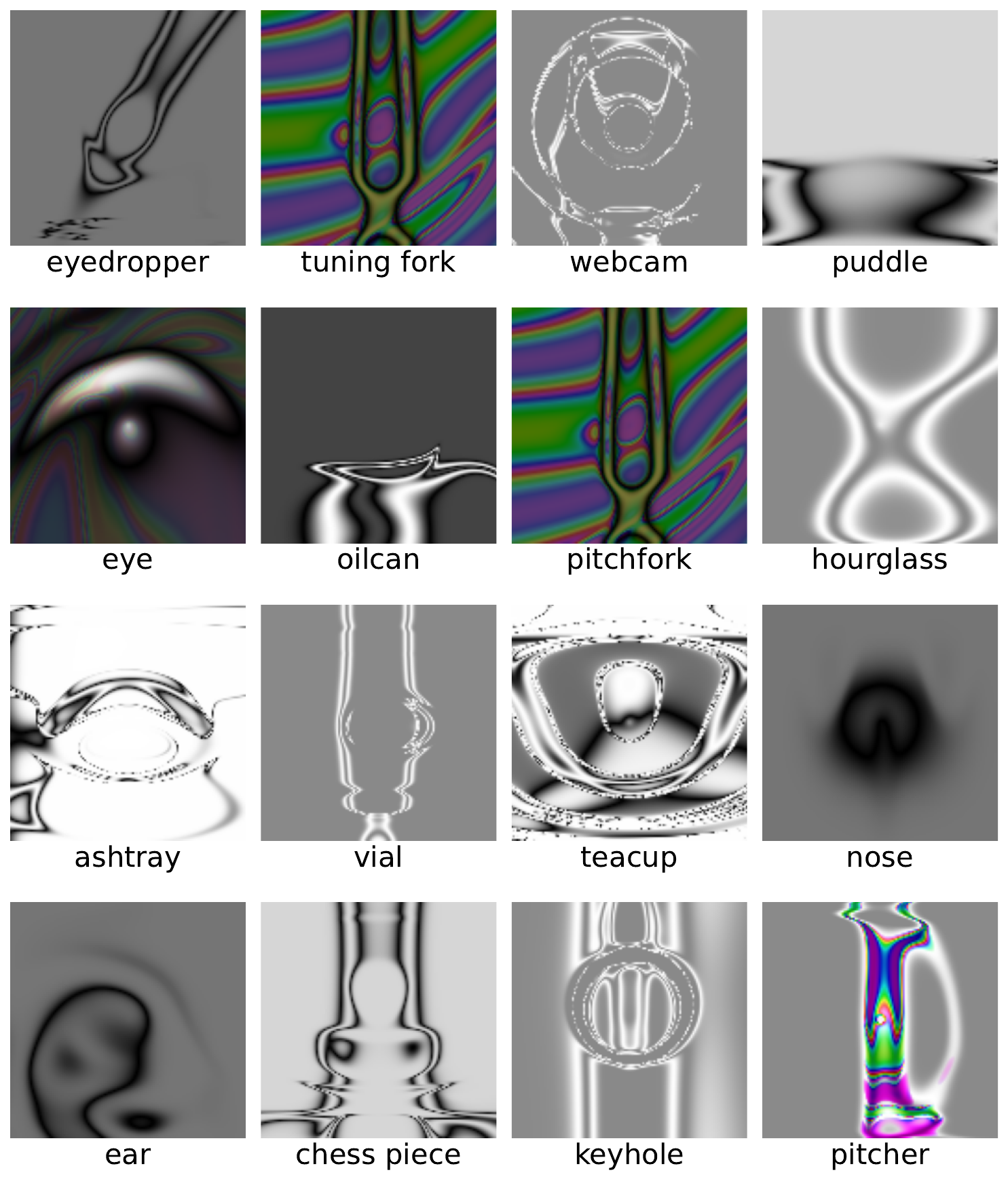}
\caption{$\epsilon = $ 1}
\end{subfigure}
\caption{Most semantically salient images in the archive, from seeds with the highest Semantic Recall. Archive with highest Semantic Recall (\autoref{fig:exploration_grids_nouniest_epsilon_0_25}) is outlined.}
\label{fig:exploration_grids_nouniest}
\end{subfigure}
\caption{Effect of exploration ($\epsilon$-greedy) on Semantic Recall within the Picbreeder archive. Forcing agents to take some random selections (i.e., with probability 0.25) can improve the quality of the archive, with Semantic Recall score approaching that of the historical human baseline. Large amounts of random parent selection ($\epsilon \geq 0.5$), and a fully random baseline (in which branching, publication, and archive rating decisions are also random) are detrimental.}
\label{fig:exploration}
\end{figure*}

\begin{figure*}
\begin{subfigure}{\linewidth}
\begin{subfigure}{0.33\linewidth}
\centering
\includegraphics[width=1.0\linewidth]{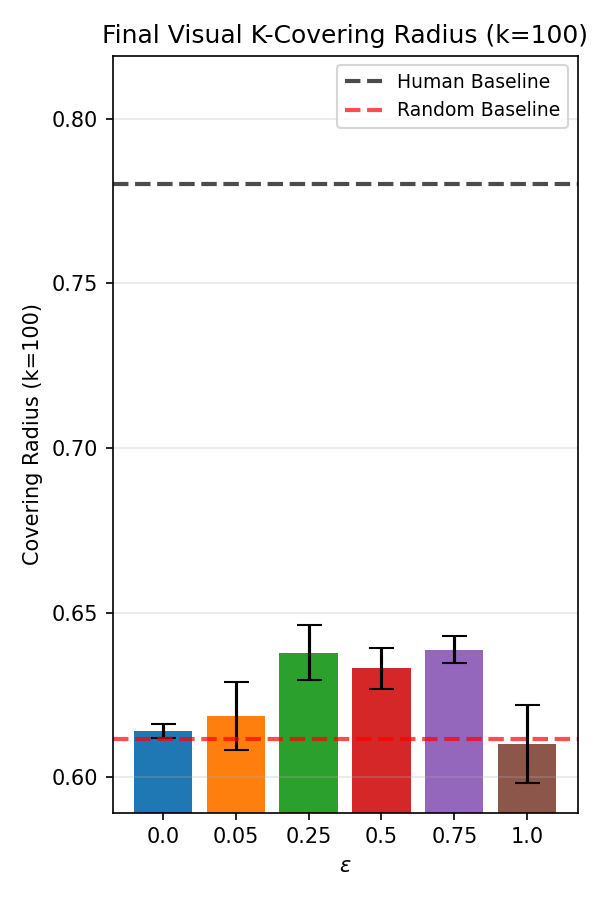}
\caption{Visual Coverage}
\label{fig:exploration_visual_coverage}
\end{subfigure}
\begin{subfigure}{0.33\linewidth}
\centering
\includegraphics[width=1.0\linewidth]{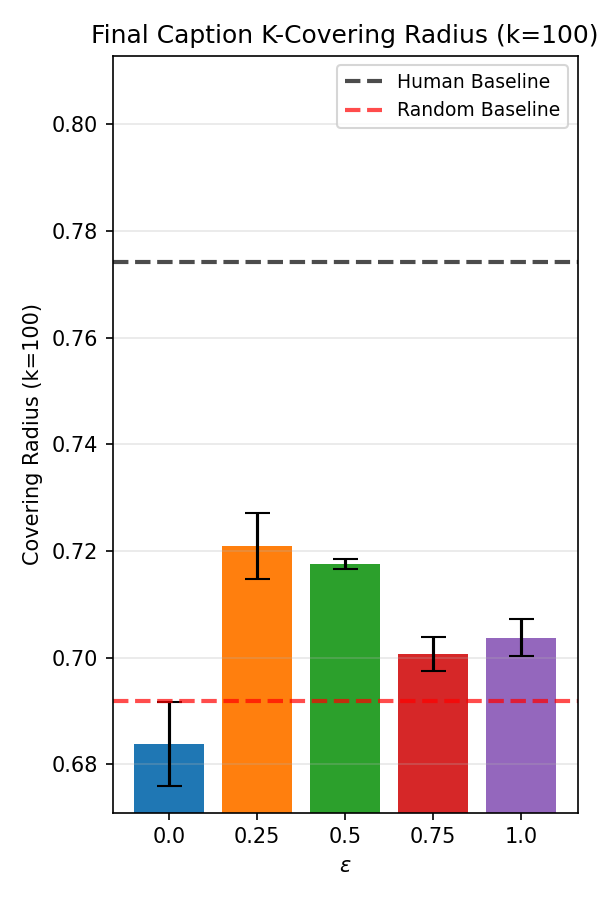}
\caption{Semantic Coverage}
\label{fig:exploration_semantic_coverage}
\end{subfigure}
\begin{subfigure}{0.33\linewidth}
\centering
\includegraphics[width=1.0\linewidth]{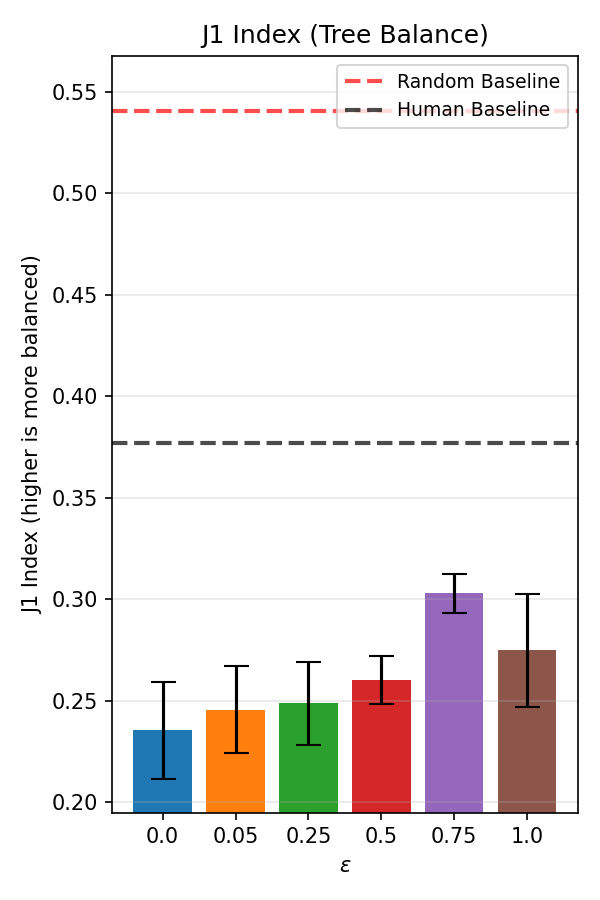}
\caption{Phylogenetic Tree Balance}
\label{fig:exploration_j1_index}
\end{subfigure}
\caption{Diversity measures of Picbreeder archives after $2,000$ agent sessions.}
\end{subfigure}
\begin{subfigure}{\linewidth}
\centering
\begin{subfigure}{.31\linewidth}
\includegraphics[width=\linewidth]{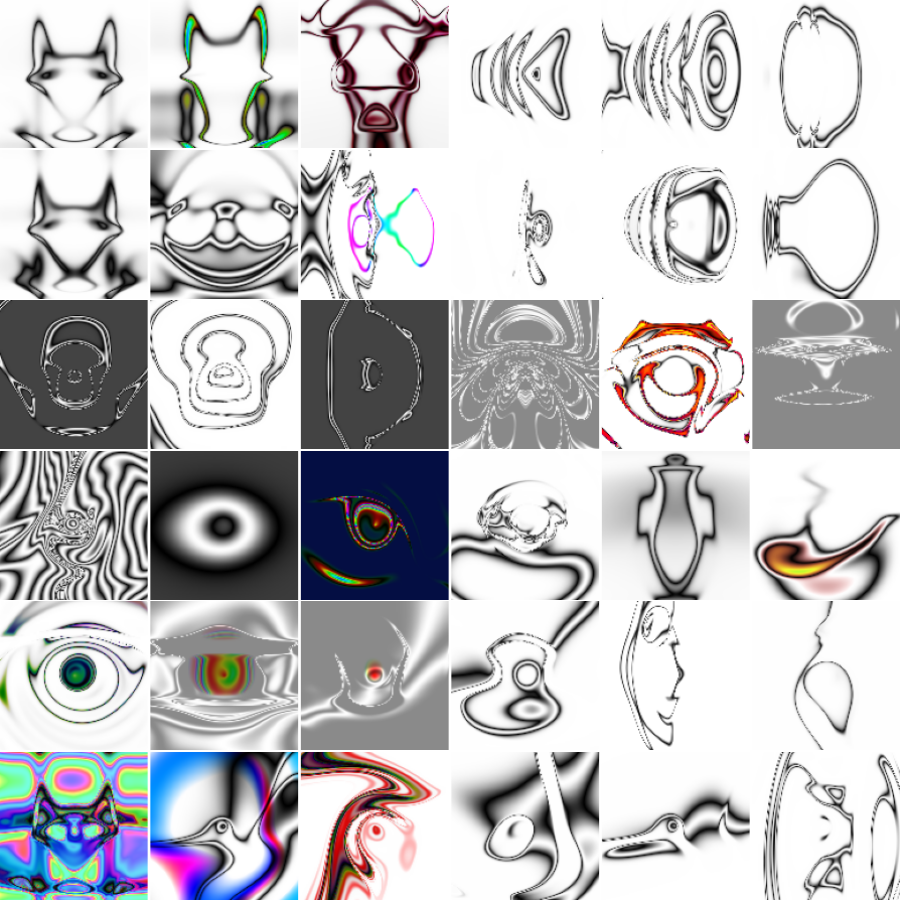}
\caption{$\epsilon$ = 0}
\label{fig:exploration_grids_representative_epsilon_0}
\end{subfigure}
\hfill
\begin{subfigure}{.31\linewidth}
\centering
\includegraphics[width=\linewidth,fbox={1pt 0pt}]{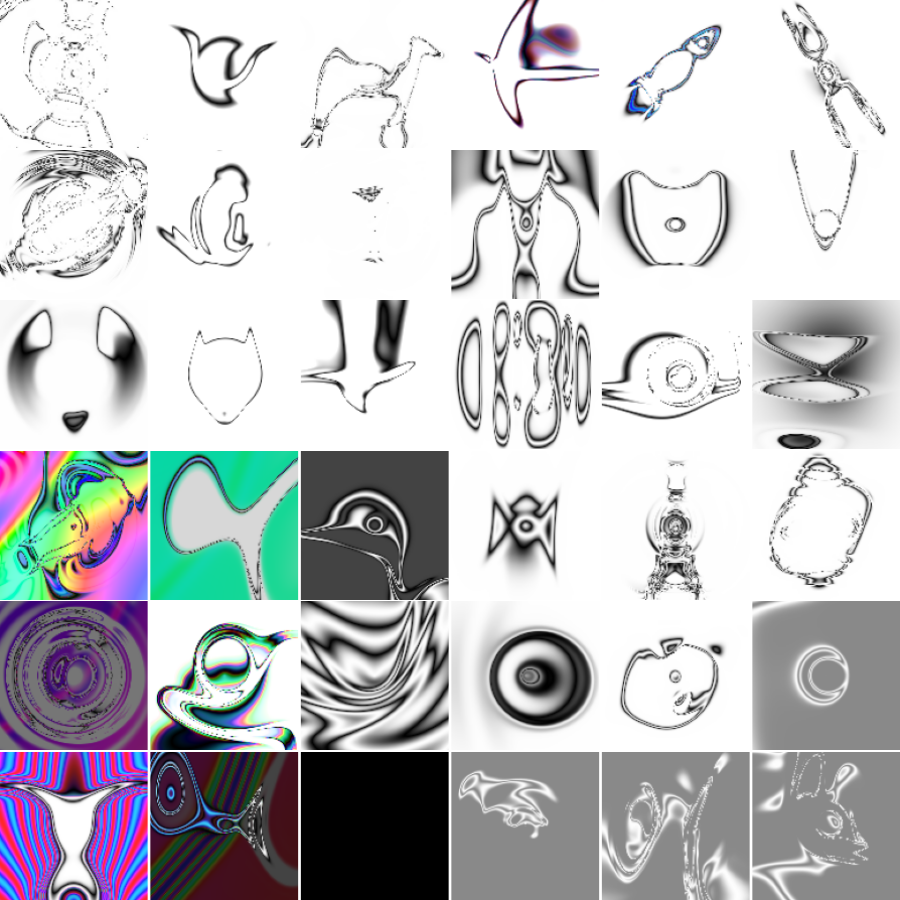}
\caption{$\epsilon$ = 0.25}
\label{fig:exploration_grids_representative_epsilon_0_25}
\end{subfigure}
\hfill
\begin{subfigure}{.31\linewidth}
\centering
\includegraphics[width=\linewidth]{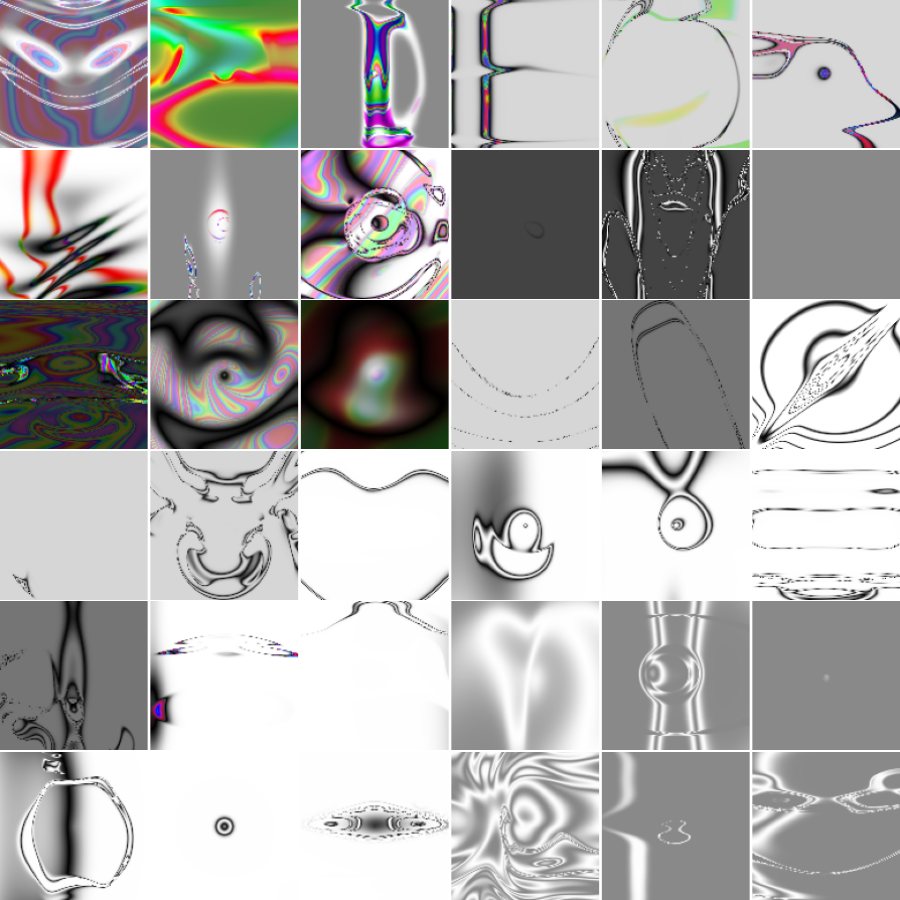}
\caption{$\epsilon = $ 1}
\label{fig:exploration_grids_representative_epsilon_1}
\end{subfigure}
\caption{Visually representative images from the archive, from seeds with the highest Visual Coverage. Archive with highest Visual Coverage (\autoref{fig:exploration_grids_representative_epsilon_0_25}) is outlined.}
\label{fig:exploration_grids_representative}
\end{subfigure}
\caption{Effect of exploration ($\epsilon$-greedy) on the diversity of the Picbreeder archive.
A moderate amount of noise can increase Visual and Semantic Coverage and Tree Balance, but, in excess, reduces the legibility of generated images (cf. \autoref{fig:exploration_grids_representative_epsilon_1}, \autoref{fig:exploration}).
}
\label{fig:exploration_grids}
\end{figure*}

\begin{figure*}
\begin{subfigure}{\linewidth}
\centering
\includegraphics[width=1.0\linewidth]{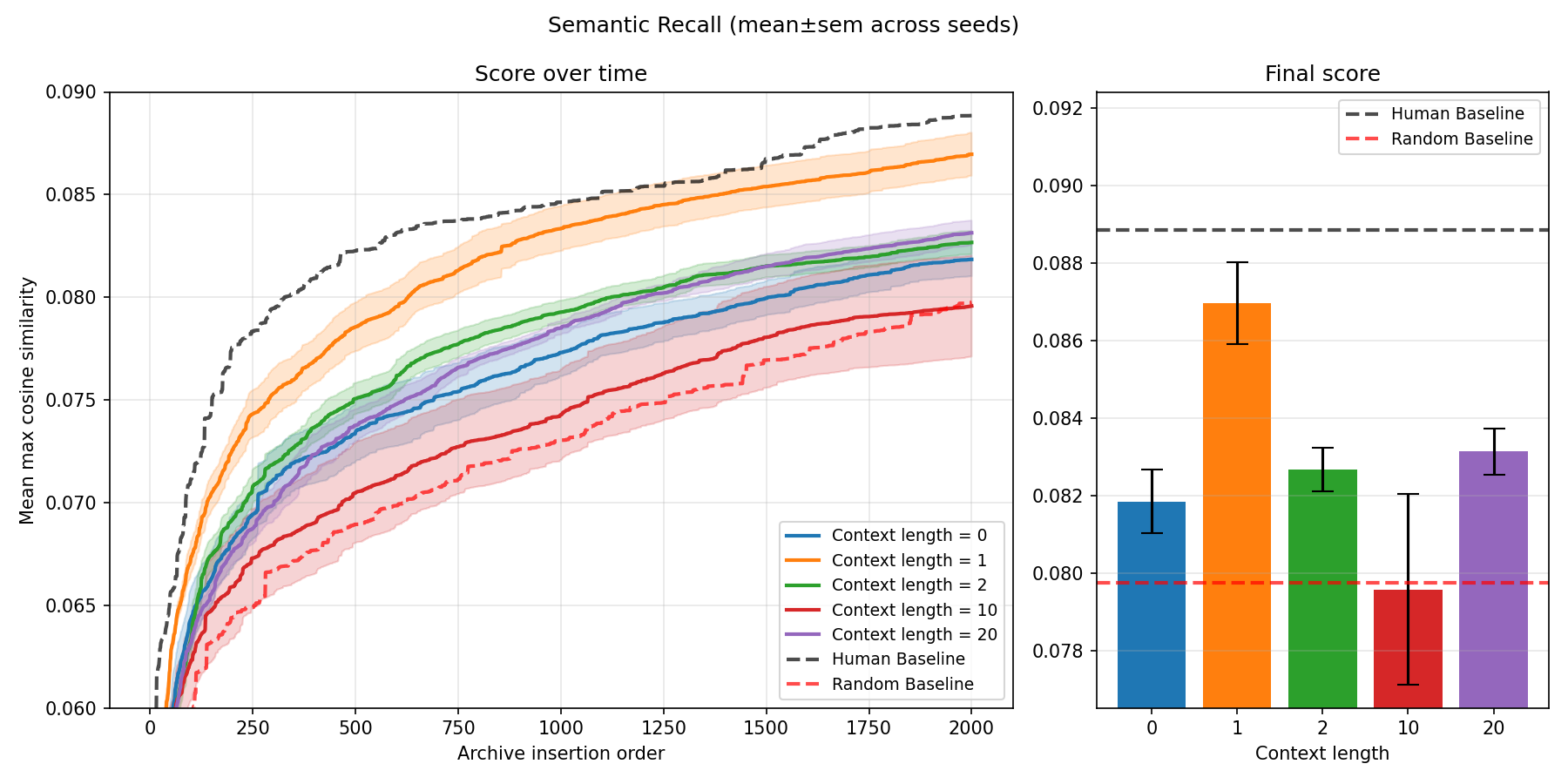}
\caption{Semantic Recall score within the Picbreeder archive over the course of collaborative evolution.}
\label{fig:history_semantic_recall}
\end{subfigure}
\begin{subfigure}{\linewidth}
\centering
\begin{subfigure}{.31\linewidth}
\includegraphics[width=\linewidth]{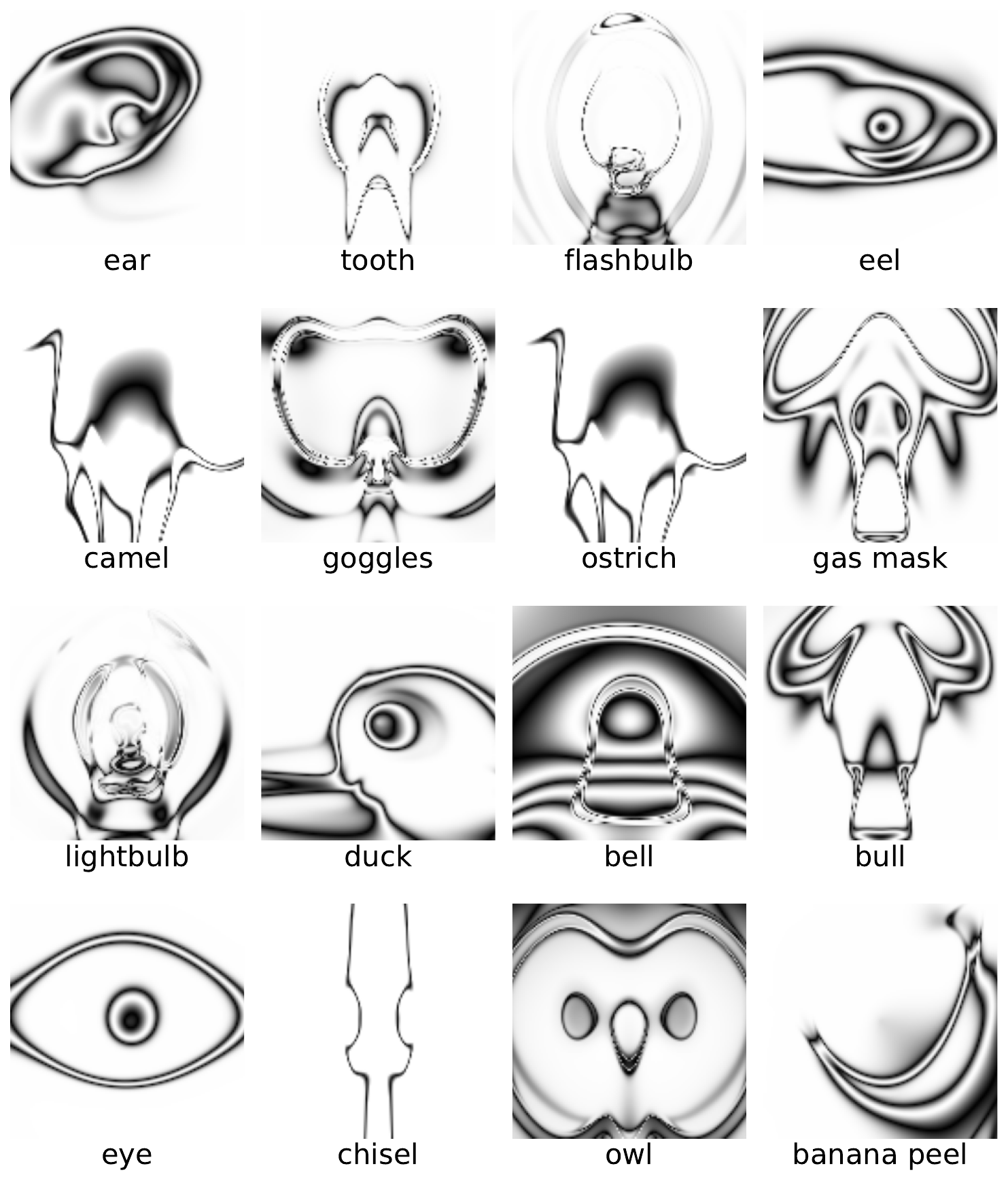}
\caption{Context length = 0}
\label{fig:history_grids_nouniest_0}
\end{subfigure}
\hfill
\begin{subfigure}{.31\linewidth}
\centering
\includegraphics[width=\linewidth]{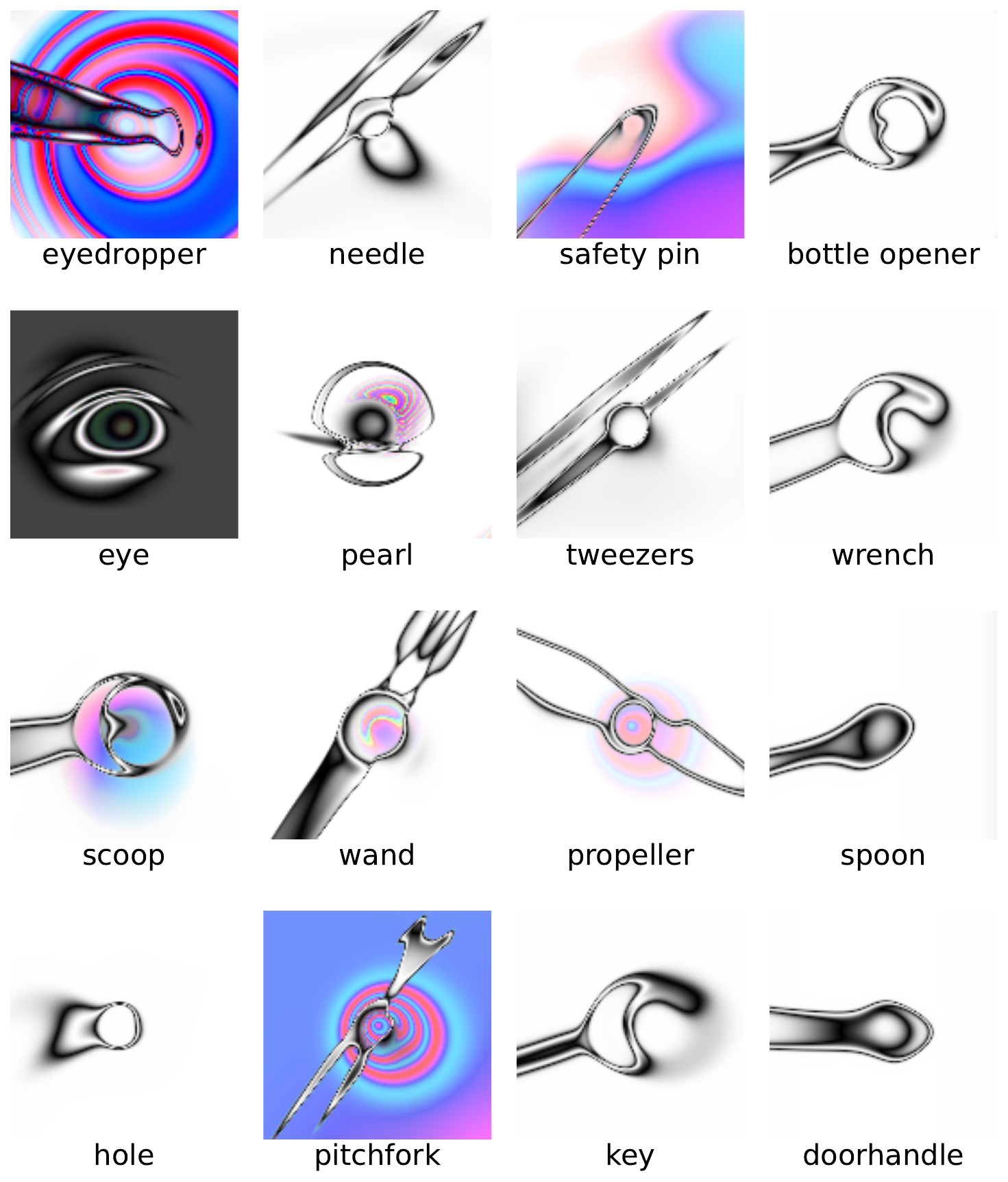}
\caption{Context length = 10}
\label{fig:history_grids_nouniest_10}
\end{subfigure}
\hfill
\begin{subfigure}{.31\linewidth}
\centering
\includegraphics[width=\linewidth,fbox={1pt 0pt}]{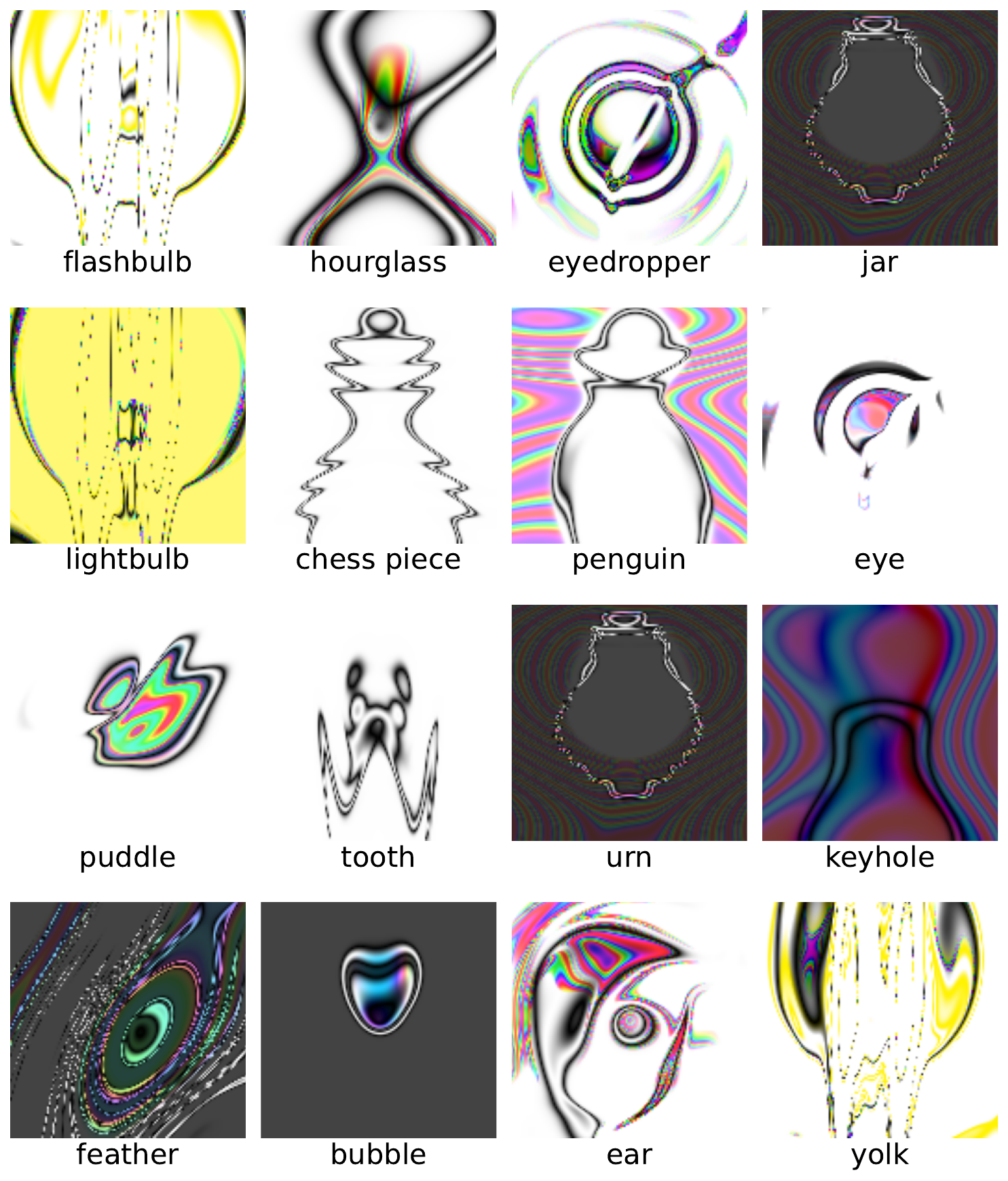}
\caption{Context length = 20 (full)}
\label{fig:history_grids_nouniest_20}
\end{subfigure}
\caption{Most semantically salient images in the archive, from seeds with the highest Semantic Recall. Archive with highest Semantic Recall (\autoref{fig:history_grids_nouniest_20}) is outlined.}
\label{fig:history_grids_nouniest}
\end{subfigure}
\caption{Effect of history---i.e. Context Length ($CL$), the number of previous actions included in an agent's context---on Semantic Recall within the Picbreeder archive. Without any context, mode collapse is common, leading to reduced recall. $CL=1$ proves to be a surprisingly effective sweep spot, with larger $CL$ leading to overly noisy/abstract forms.}
\label{fig:history}
\end{figure*}

\begin{figure*}
\begin{subfigure}{\linewidth}
\begin{subfigure}{0.33\linewidth}
\centering
\includegraphics[width=1.0\linewidth]{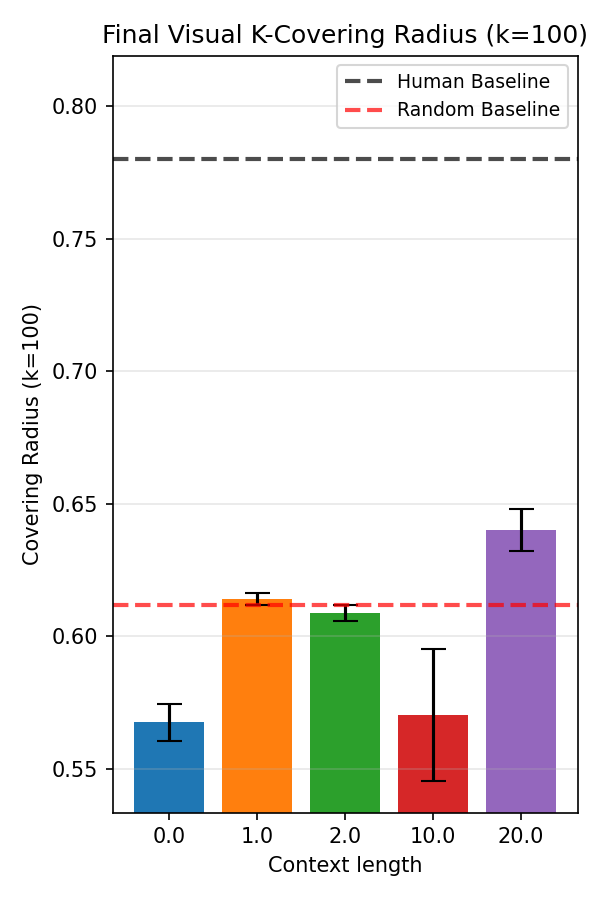}
\caption{Visual Coverage}
\label{fig:history_visual_coverage}
\end{subfigure}
\begin{subfigure}{0.33\linewidth}
\centering
\includegraphics[width=1.0\linewidth]{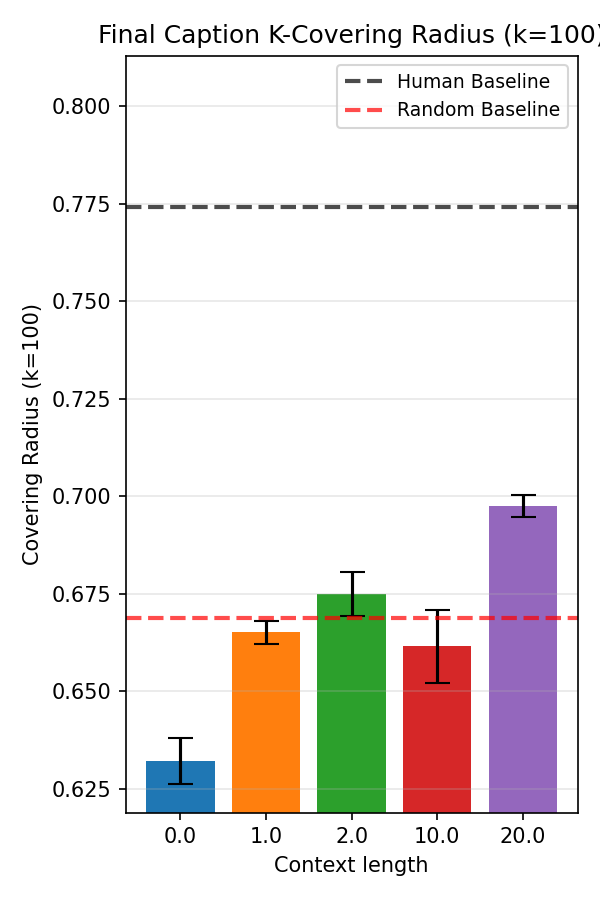}
\caption{Semantic Coverage}
\label{fig:history_semantic_coverage}
\end{subfigure}
\begin{subfigure}{0.33\linewidth}
\centering
\includegraphics[width=1.0\linewidth]{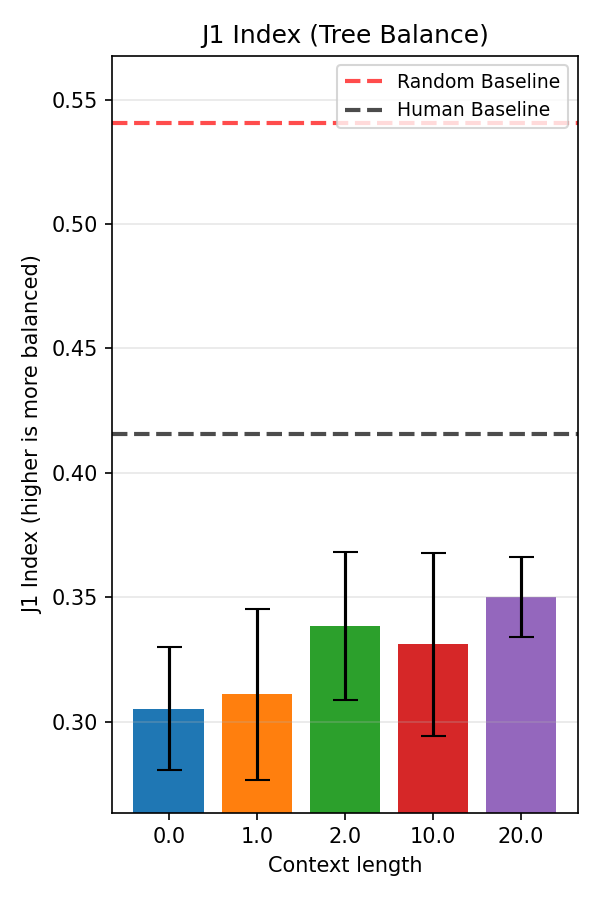}
\caption{Phylogenetic Tree Balance}
\label{fig:history_j1}
\end{subfigure}
\caption{Diversity measures of Picbreeder archives after $2,000$ agent sessions.}
\end{subfigure}
\begin{subfigure}{\linewidth}
\centering
\begin{subfigure}{.31\linewidth}
\includegraphics[width=\linewidth]{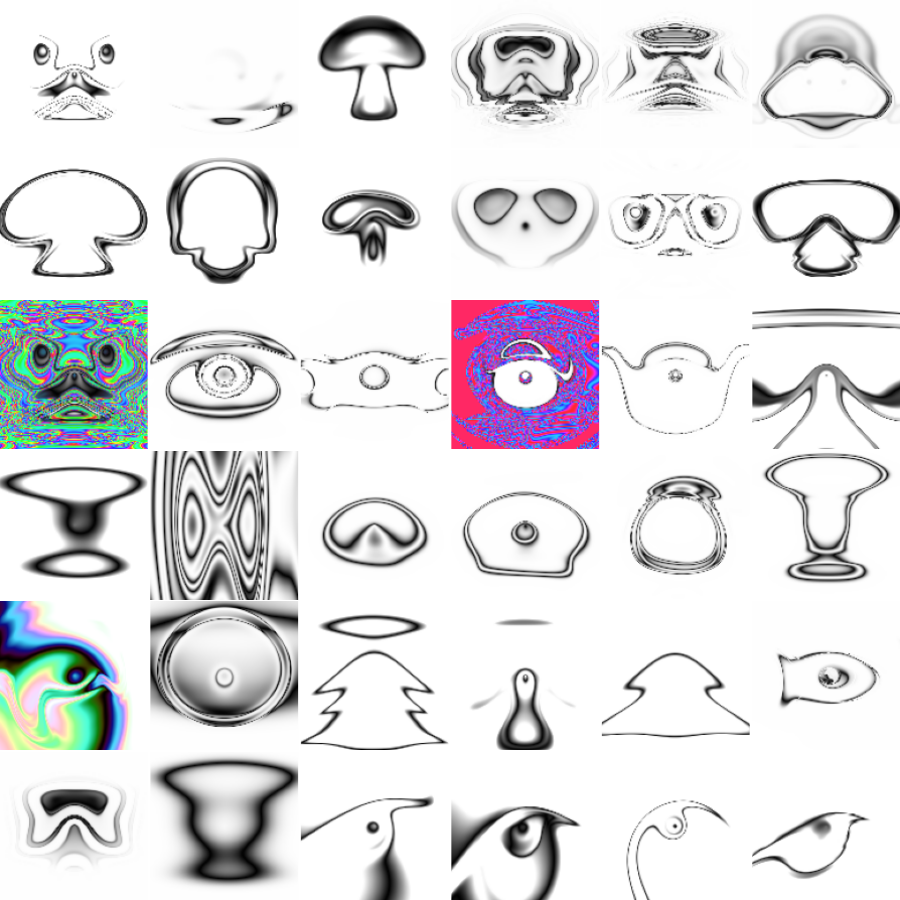}
\caption{Context length $=$ 0}
\label{fig:history_grids_representative_0}
\end{subfigure}
\hfill
\begin{subfigure}{.31\linewidth}
\centering
\includegraphics[width=\linewidth]{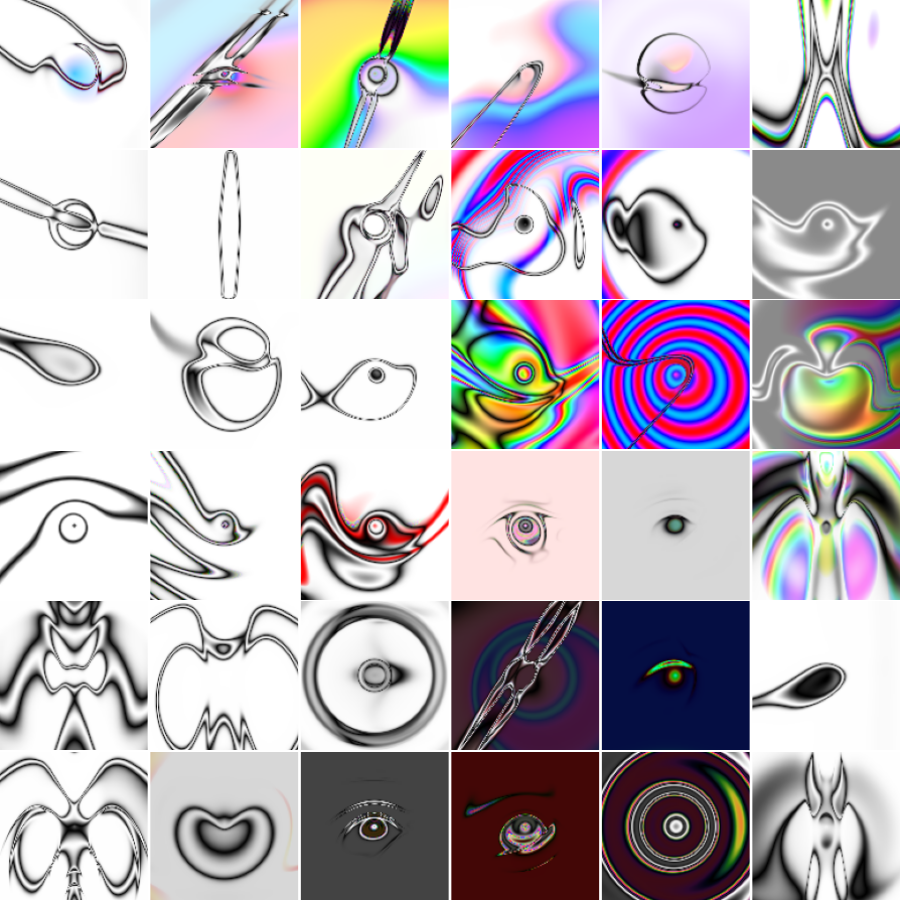}
\caption{Context length $=$ 10}
\label{fig:history_grids_representative_10}
\end{subfigure}
\hfill
\begin{subfigure}{.31\linewidth}
\centering
\includegraphics[width=\linewidth,fbox={1pt 0pt}]{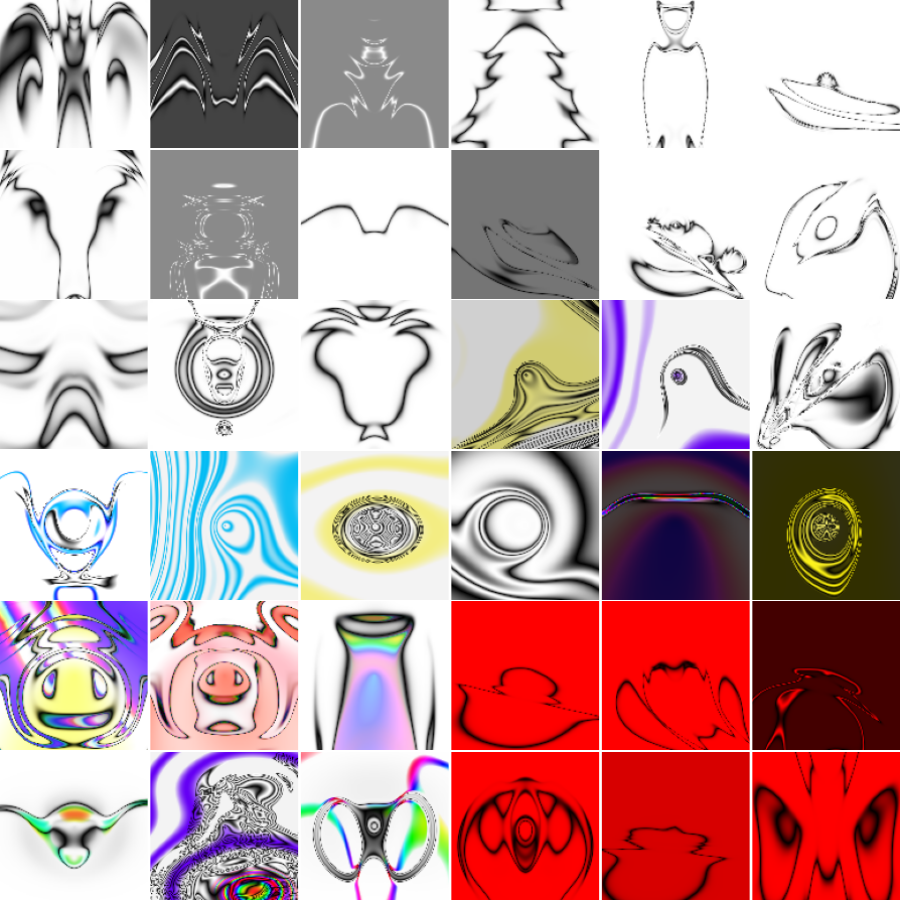}
\caption{Context length $=$ 20 (full)}
\label{fig:history_grids_representative_20}
\end{subfigure}
\caption{Visually representative images from the archive, from seeds with the highest visual coverage. Archive with highest Visual Coverage (\autoref{fig:history_grids_representative_20}) is outlined.}
\label{fig:history_grids_representative}
\end{subfigure}
\caption{Effect of history---i.e. Context Length ($CL$), the number of previous actions included in an agent's context---on diversity within the Picbreeder archive. Increasing $CL$ increases diversity, but with rapidly diminishing returns. $CL=20$ is an exception; here, diversity peaks, likely because in this case the agent is prompted with an additional note encouraging its publication to be novel w.r.t. the still-visible archive sample (\autoref{fig:prompt_addendum}). But the images in these archives are more noisy/abstract, as reflected in their low Semantic Recall scores (\autoref{fig:history_semantic_recall}). This may be due to the context window being overloaded, leading to decreased VLM performance in general; or due to the incentivization of novelty combined with forced publication steps effectively leading to the premature publication of semantically ill-defined works-in-progress.}
\end{figure*}

\begin{figure*}
\begin{subfigure}{\linewidth}
\centering
\includegraphics[width=\linewidth]{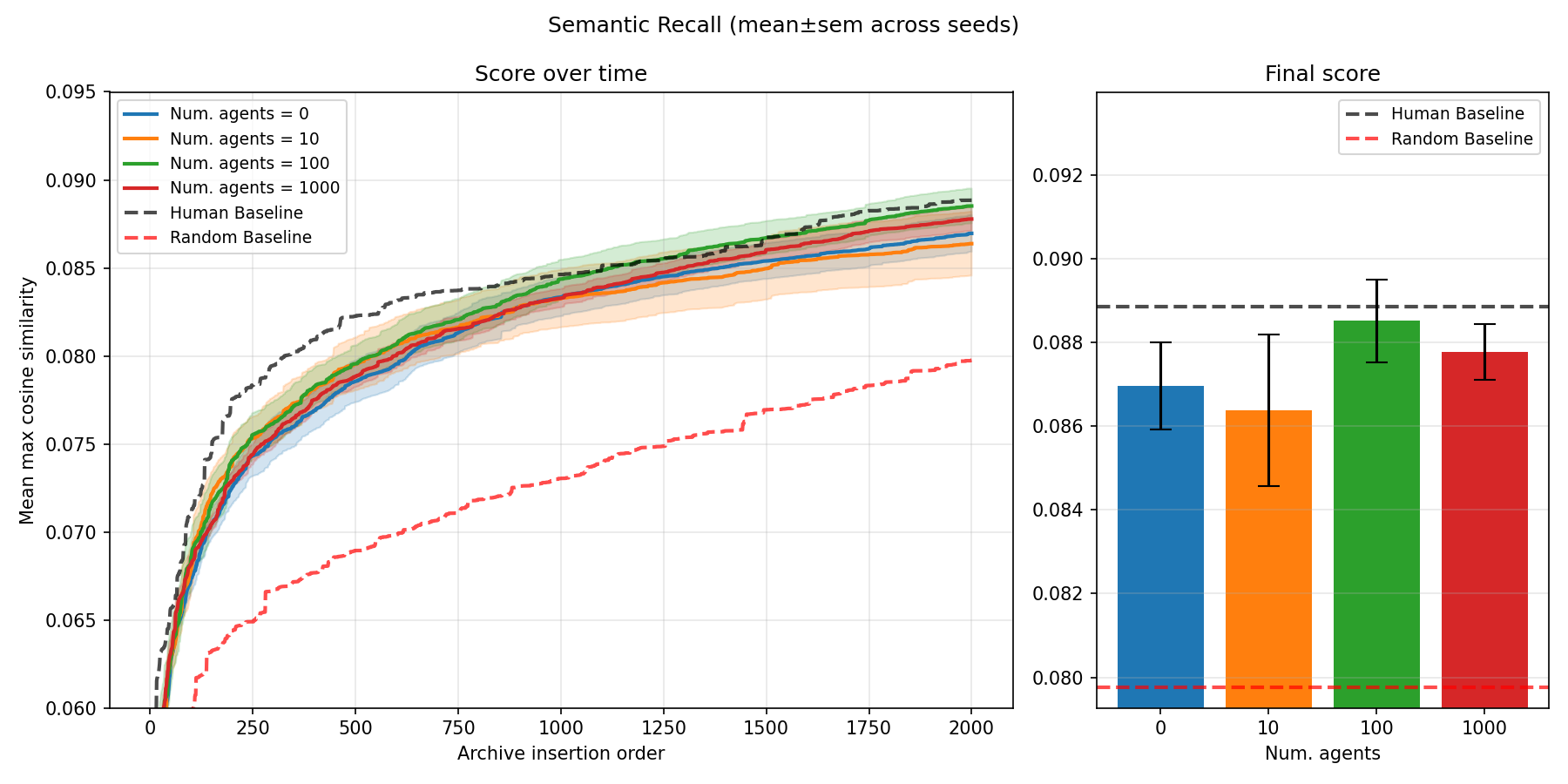}
\caption{Semantic Recall score within the Picbreeder archive over the course of collaborative evolution.}
\label{fig:traits_semantic_recall}
\end{subfigure}
\begin{subfigure}{\linewidth}
\begin{subfigure}{0.33\linewidth}
\centering
\includegraphics[width=1.0\linewidth]{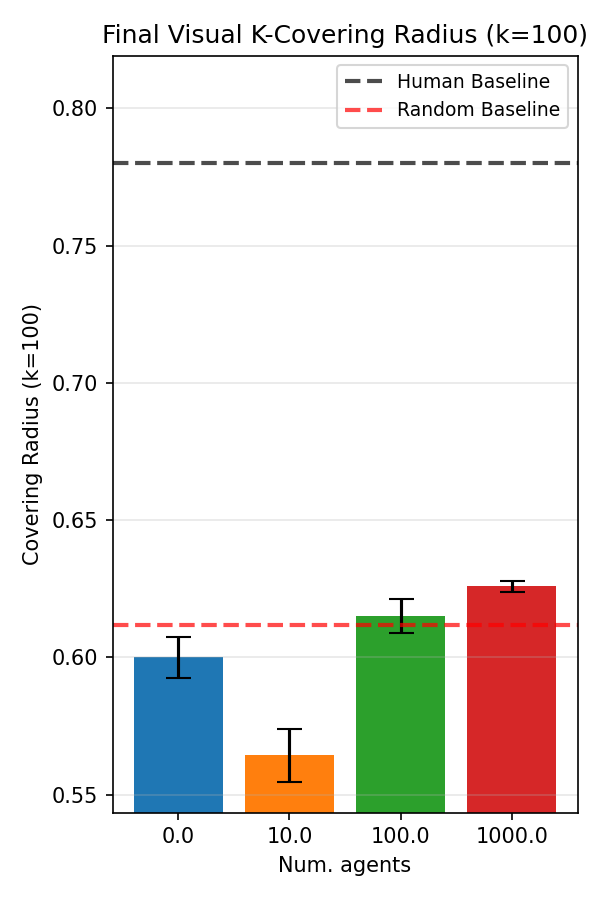}
\caption{Visual Coverage}
\label{fig:traits_visual_coverage}
\end{subfigure}
\begin{subfigure}{0.33\linewidth}
\centering
\includegraphics[width=1.0\linewidth]{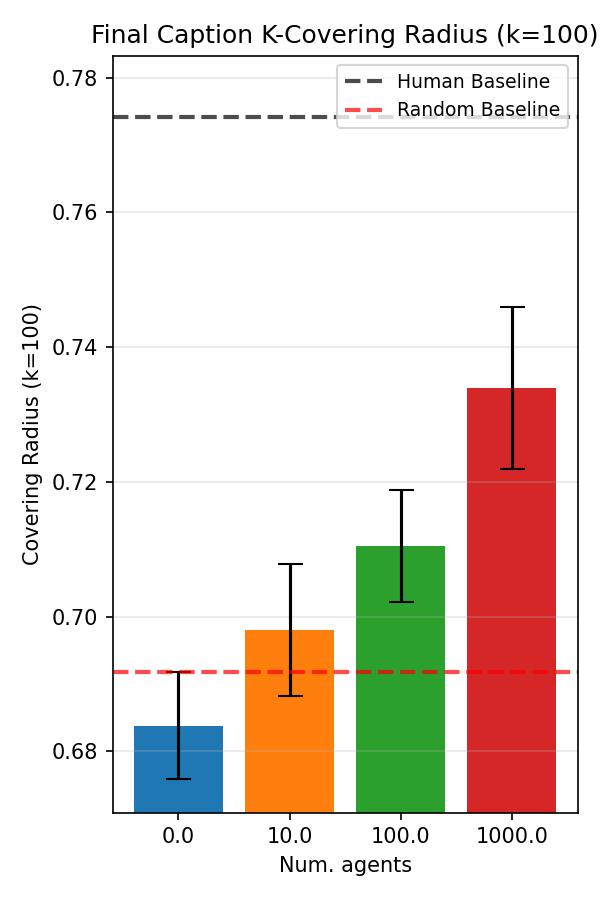}
\caption{Semantic Coverage}
\label{fig:traits_semantic_coverage}
\end{subfigure}
\begin{subfigure}{0.33\linewidth}
\centering
\includegraphics[width=1.0\linewidth]{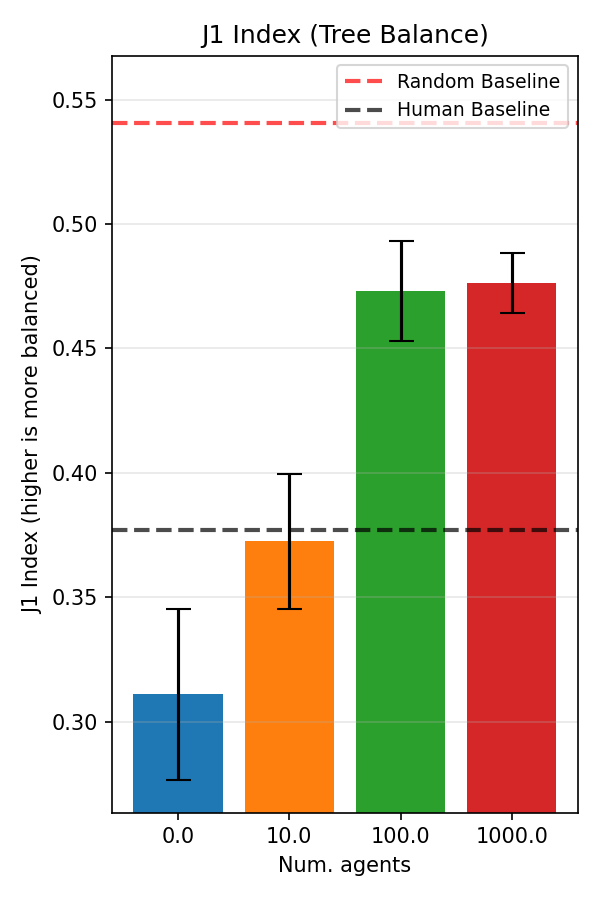}
\caption{Phylogenetic Tree Balance}
\label{fig:traits_j1_index}
\end{subfigure}
\caption{Diversity measures of Picbreeder archives after $2,000$ agent sessions.}
\label{fig:traits_diversity}
\end{subfigure}
\caption{Effect of number of agents $NA$---i.e. the number of distinct personality traits (see \autoref{tab:traits}) distributed among Picbreeder sessions---on Semantic Recall score and diversity metrics. Increasing $NA$ improves various metrics of diversity without harming Semantic Recall.}
\label{fig:traits}
\end{figure*}

\begin{figure*}
\centering
\begin{subfigure}{.49\linewidth}
\includegraphics[width=\linewidth]{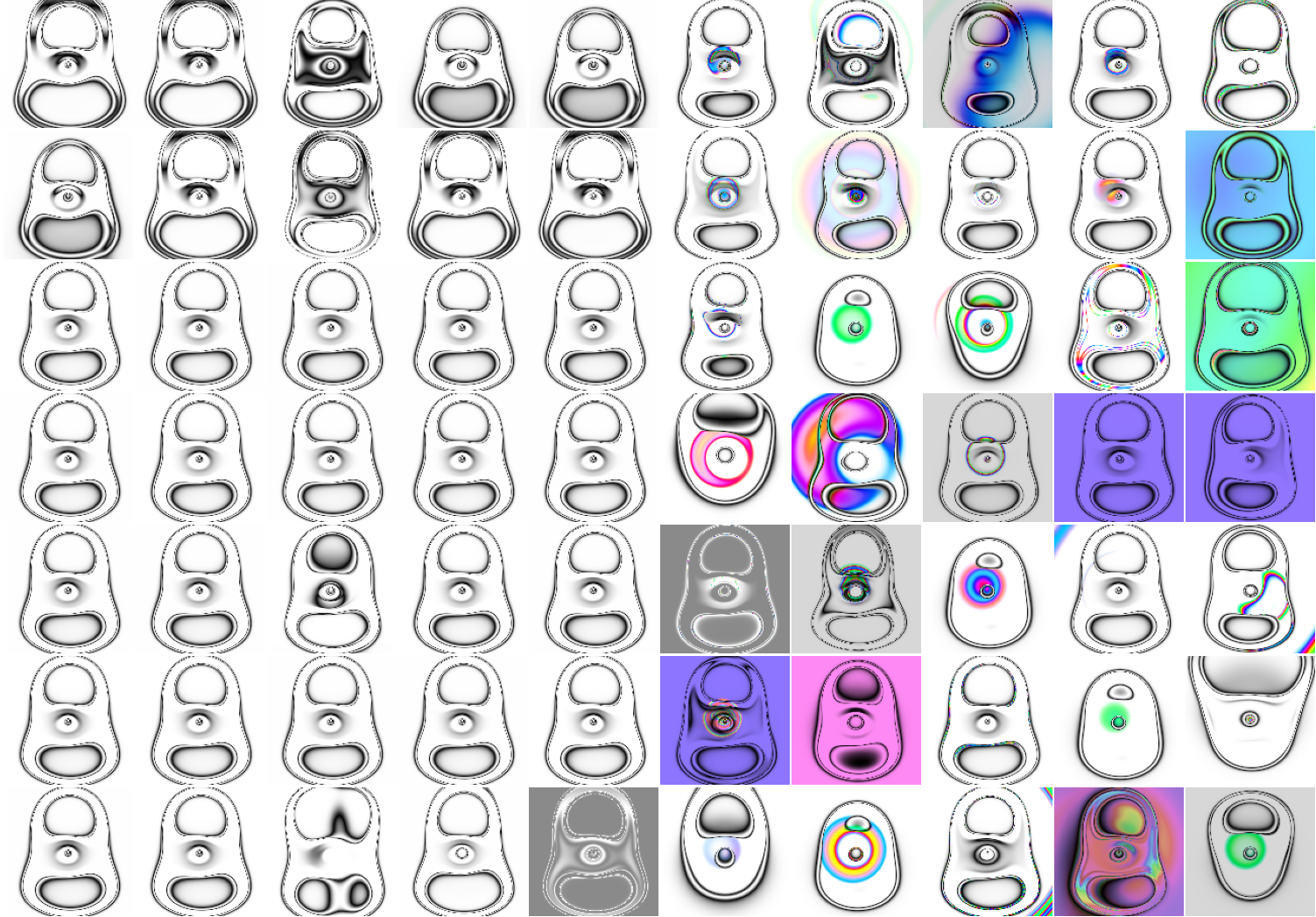}
\caption{Soda can pull tabs}
\label{fig:soda}
\end{subfigure}
\hfill
\begin{subfigure}{.49\linewidth}
\includegraphics[width=\linewidth]{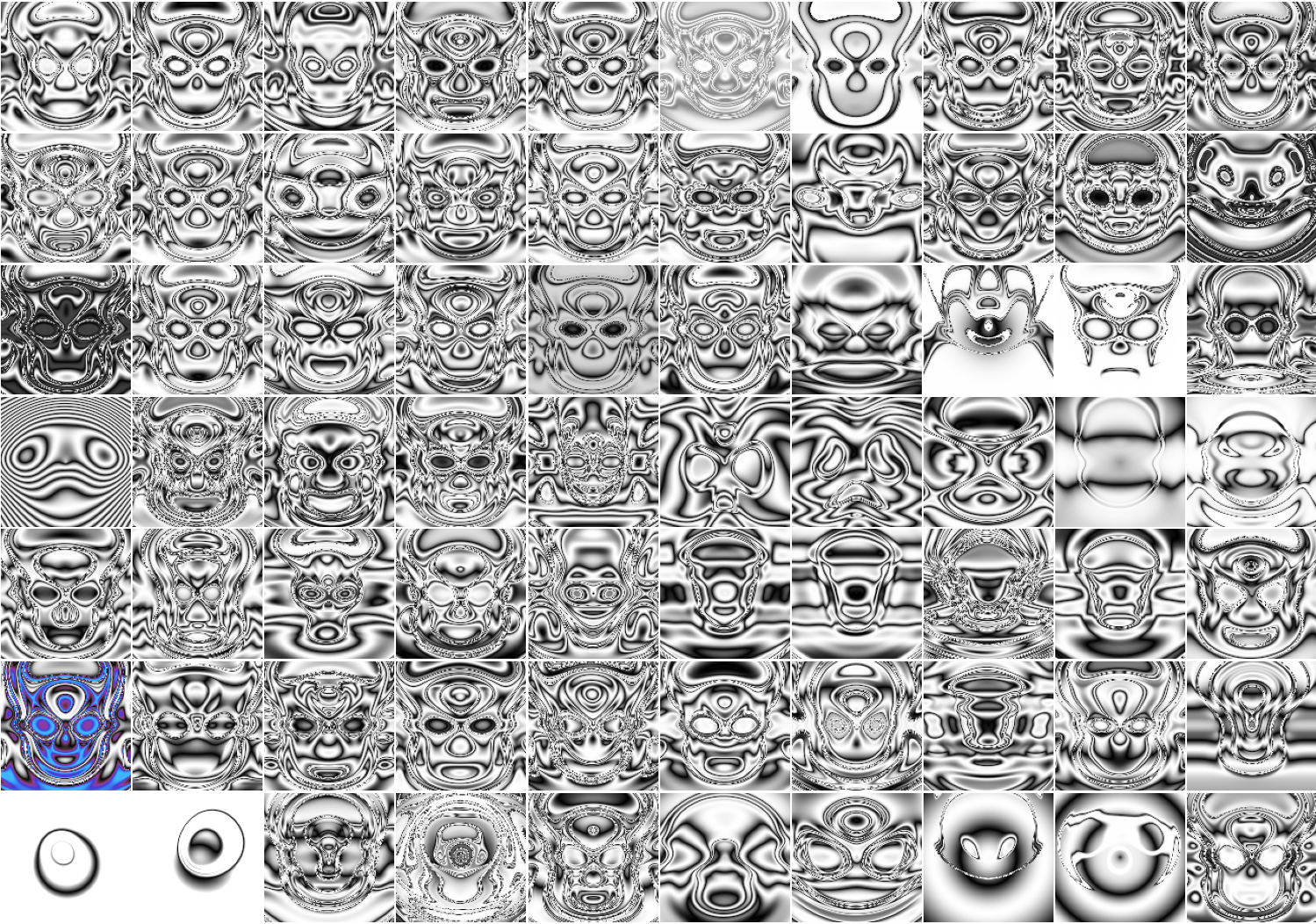}
\caption{Masks}
\label{fig:masks}
\end{subfigure}
\begin{subfigure}{.49\linewidth}
\includegraphics[width=\linewidth]{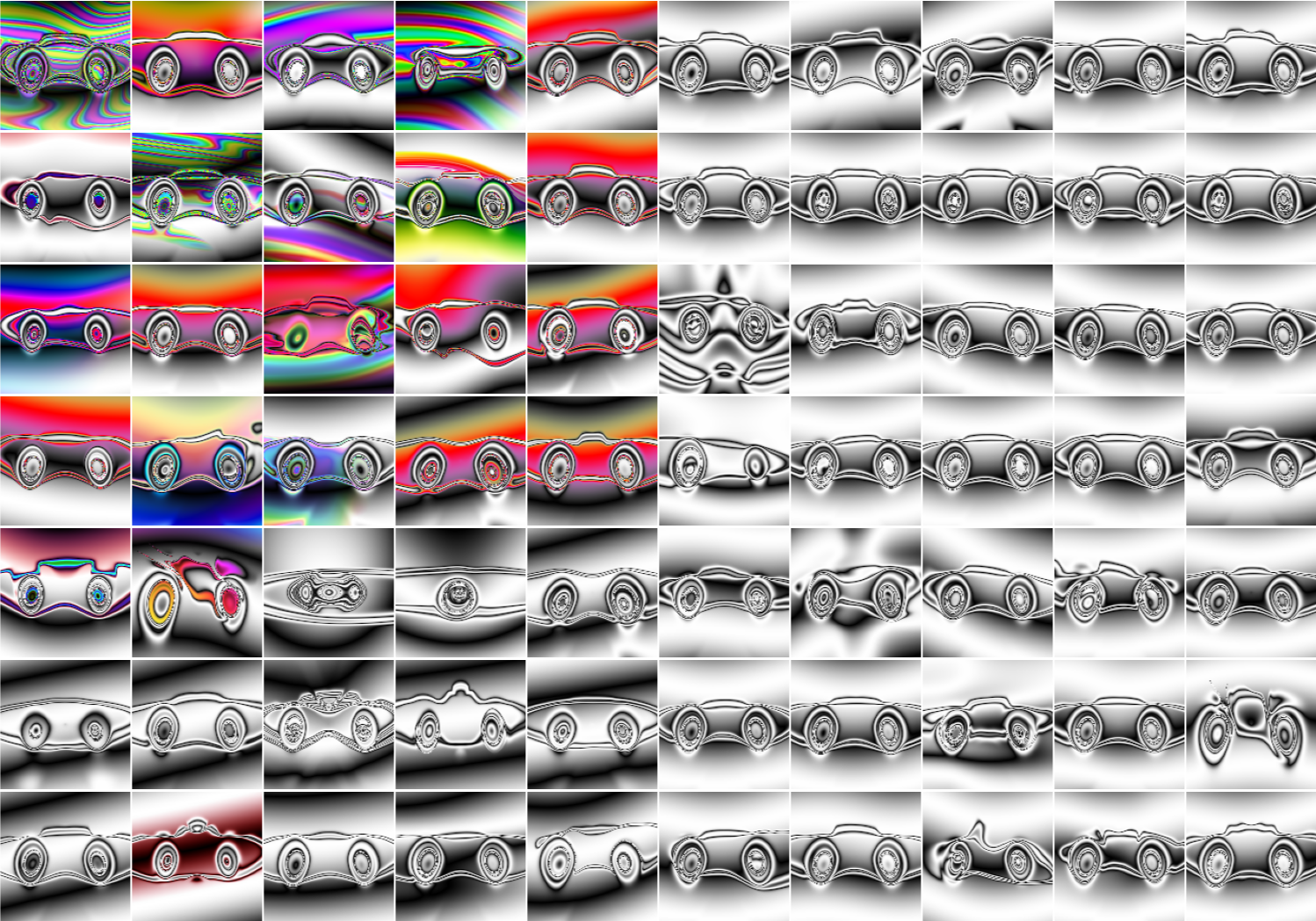}
\caption{Cars}
\label{fig:car_attractor}
\end{subfigure}
\hfill
\begin{subfigure}{.49\linewidth}
\includegraphics[width=\linewidth]{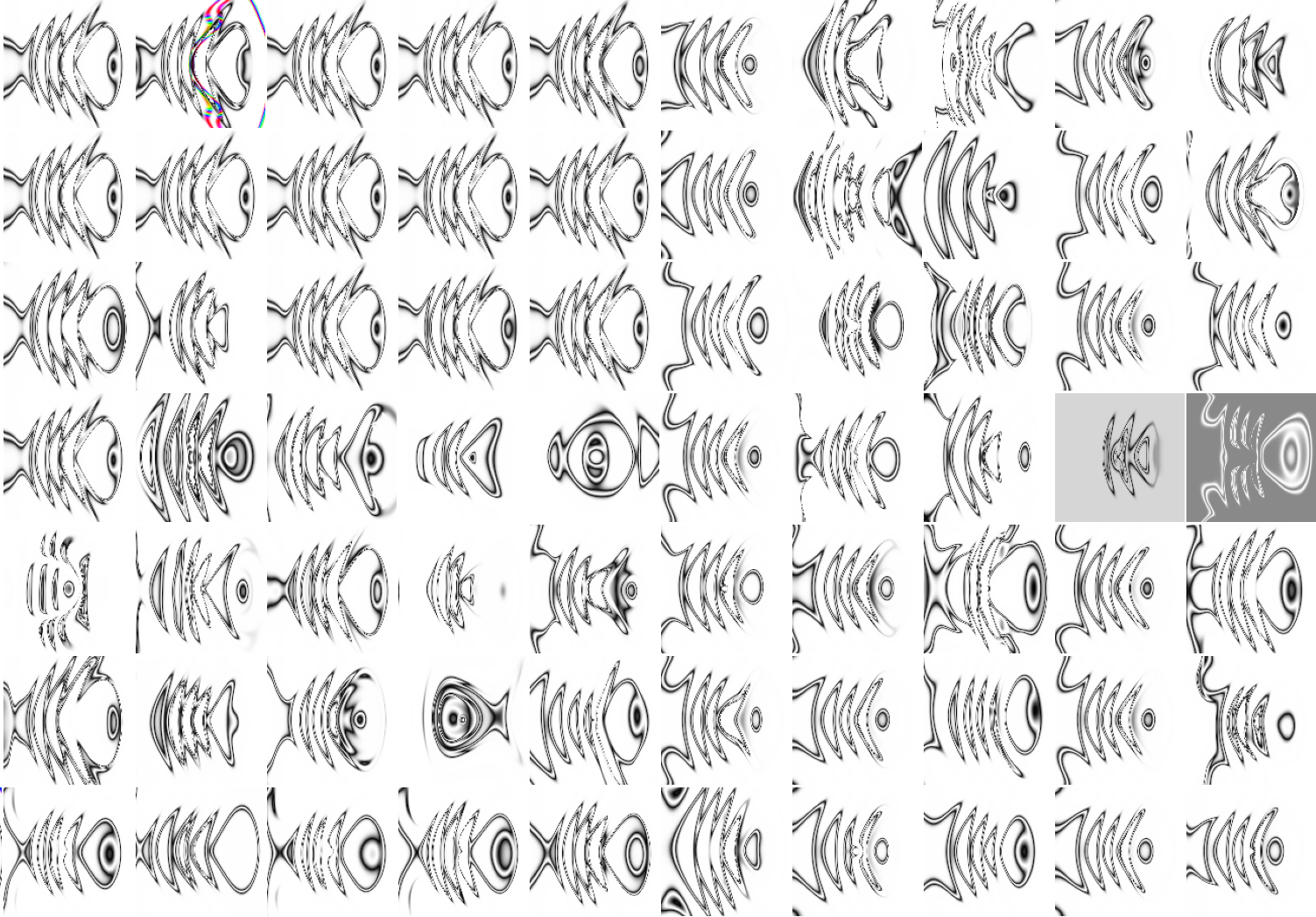}
\caption{Fish bones}
\label{fig:fish_attractor}
\end{subfigure}
\begin{subfigure}{.49\linewidth}
\includegraphics[width=\linewidth]{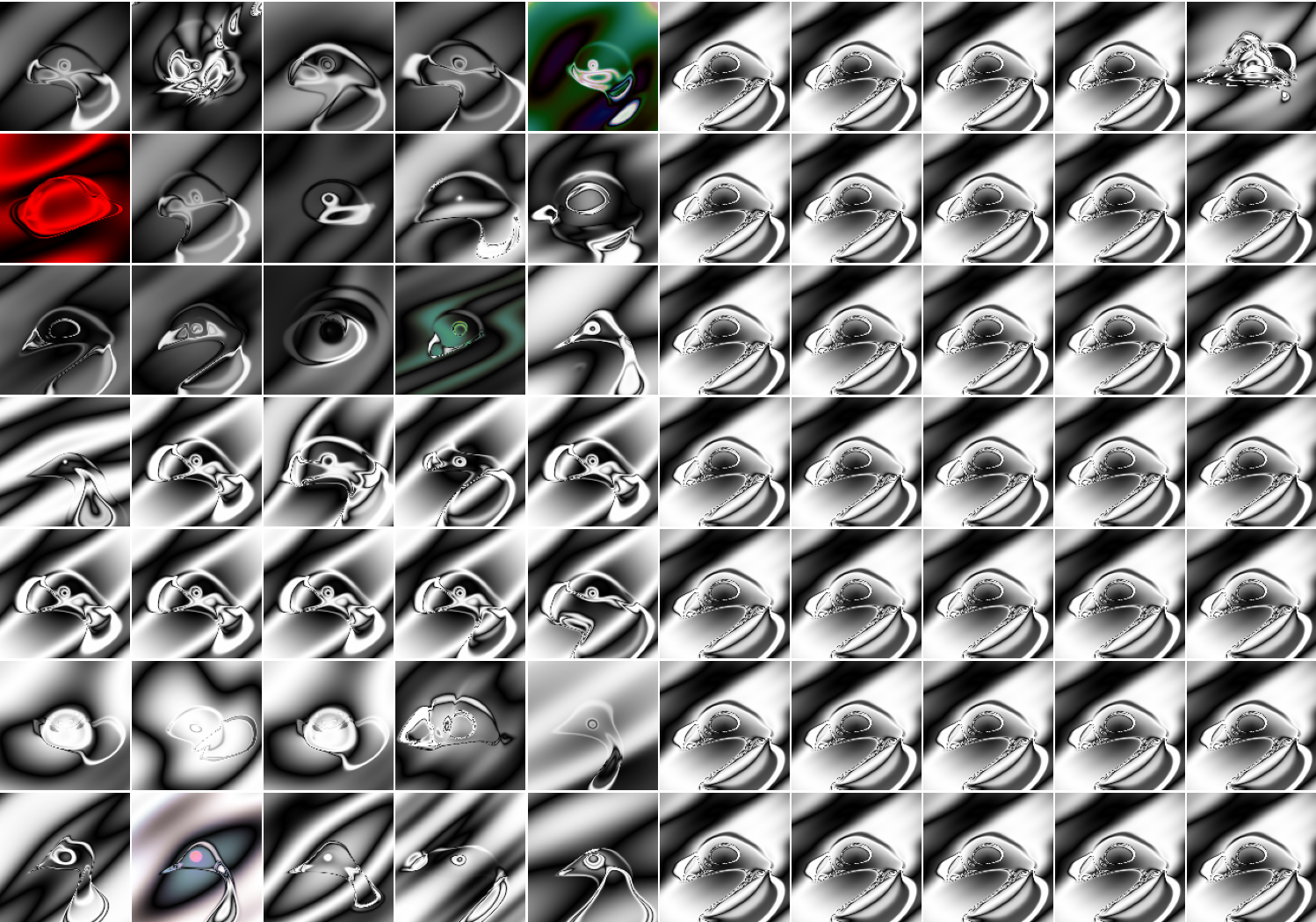}
\caption{Geese}
\label{fig:goose_attractor}
\end{subfigure}
\hfill
\begin{subfigure}{.49\linewidth}
\includegraphics[width=\linewidth]{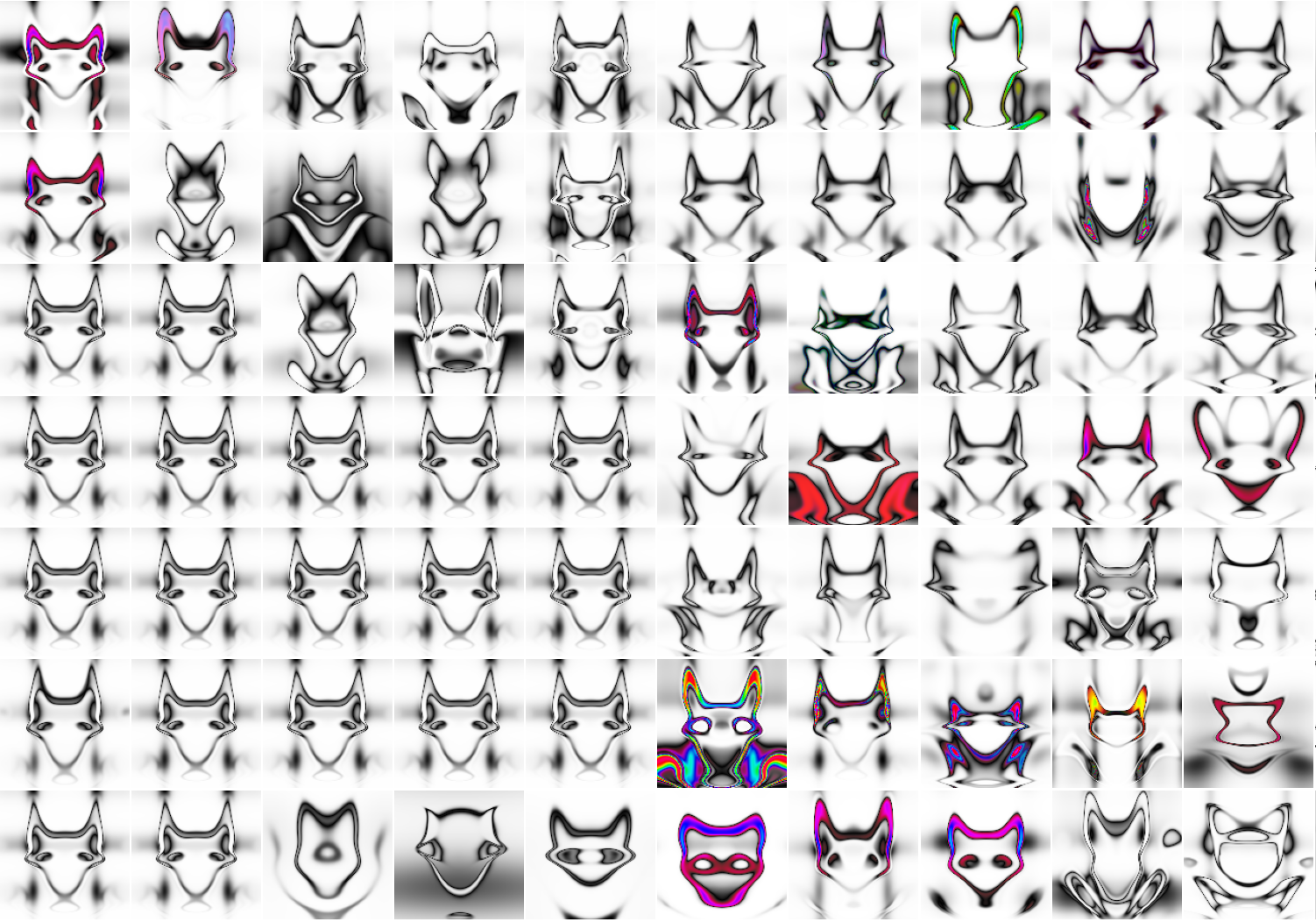}
\caption{Foxes}
\label{fig:fox_attractor}
\end{subfigure}
\caption{Semantic attractors. A common failure case of VLMs when playing Picbreeder is their tendency to fall into apparent attractors (mode collapse) in CPPN-image space. We show snapshots of various archives, with images arranged according to visual embedding distance, selecting subregions that showcase such attractors.}
\label{fig:attractor_grids}
\end{figure*}

\begin{figure*}[t]
\centering
\begin{subfigure}[t]{\textwidth}
  \centering
  \begin{tcolorbox}[
    promptlisting,
    title={Picbreeder VLM System Prompt}
  ]
  \input{prompts/system_prompt.txt}
  \end{tcolorbox}
  \vspace{-0.3cm}
  \caption{VLM system prompt.}
  \label{fig:prompt_system}
\end{subfigure}\hfill
\vspace{0.3cm}
\begin{subfigure}[t]{\textwidth}
  \centering
  \begin{tcolorbox}[
    promptlisting,
    title={Picbreeder VLM Novelty Prompt}
  ]
  \input{prompts/novelty_addendum.txt}
  \end{tcolorbox}
  \caption{VLM novelty prompt. Appended to the system prompt when Context Length $CL=20$, i.e. when VLM agent always receives the full history of the current Picbreeder session as well as the initial archive sample with which it was initially presented for branching.}
  \label{fig:prompt_addendum}
\end{subfigure}
\caption{Prompt components used by the Picbreeder VLM agent.}
\label{fig:prompt}
\end{figure*}

\begin{figure*}[t]
\centering
\begin{tcolorbox}[
promptlisting,
title={Personality Generation Prompt}
]
\input{prompts/traits_prompt.txt}
\end{tcolorbox}
\vspace{-0.3cm}
\caption{Prompt for generating personality traits used when $NA > 0$ (see \autoref{tab:traits} for sample output from \texttt{gemini-3-pro-preview}). The VLM Picbreeder system prompt (\autoref{fig:prompt_system}) is injected where indicated.}
\label{fig:traits_prompt}
\end{figure*}

\begin{figure*}
\includegraphics[width=\linewidth]{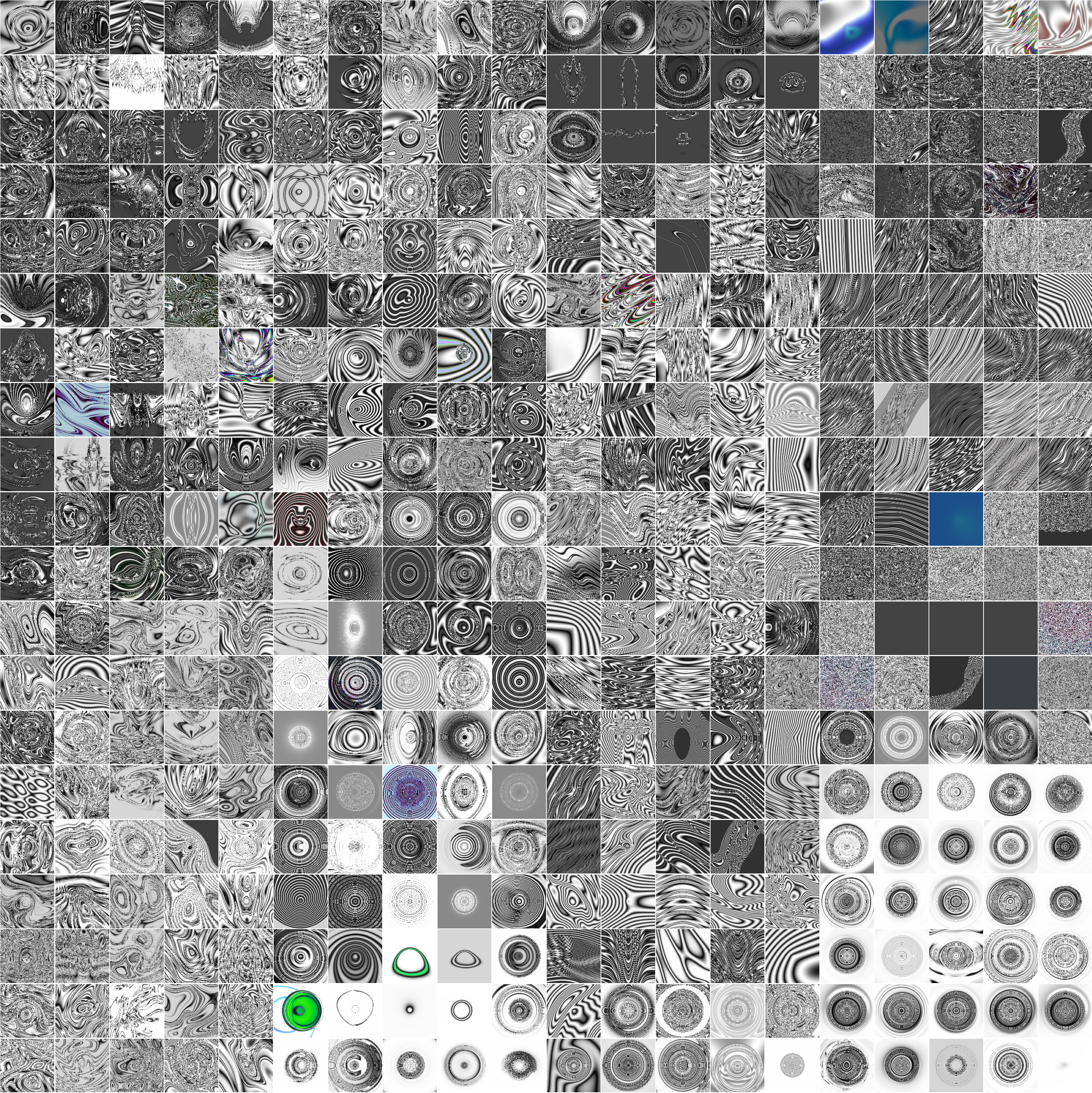}
\caption{Sample of an archive with Number of Agents $NA=1,000$, with images arranged according to visual embedding distance. We select a region of the archive that showcases the noisy, potentially adversarial images that seem to emerge at large $NA$. These might be owing to a large number of traits imbuing agents with an eye for relatively abstract properties of an image, where VLMs focused on roleplaying may be keen to project the satisfaction of such abstract inclinations onto otherwise meaningless forms. See \autoref{tab:traits}; traits like 32 ``You are drawn to the aesthetic of bad analog TV reception'' even explicitly incentivize noise, despite our prohibiting such explicit search objectives (\autoref{fig:traits_prompt}).}
\label{fig:traits_1k_mush}
\end{figure*}

\end{document}